\documentclass[10pt,journal,compsoc]{IEEEtran}
\usepackage{amsmath,amsfonts}
\usepackage{algorithmic}
\usepackage{algorithm}
\usepackage{array}
\usepackage{textcomp}
\usepackage{stfloats}
\usepackage{url}
\usepackage{verbatim}
\usepackage{graphicx}
\usepackage{cite}
\usepackage{scalerel}
\usepackage{color}
\usepackage[english]{babel}
\usepackage{hyperref}
\usepackage{mathtools}
\usepackage{multirow}
\usepackage{makecell}
\usepackage{caption}
\usepackage{subcaption}
\usepackage{rotating}
\usepackage{diagbox}

\title{Optimal Image Transport on Sparse Dictionaries}

\author{Junqing Huang~\IEEEmembership{Student Member,~IEEE,}
	Haihui Wang~\IEEEmembership{Member,~IEEE,}
	Andreas Weiermann,
	Michael Ruzhansky
	\thanks{Manuscript received XX, XXXX, 2023; revised XX, XXXX, XX and accepted XX, XXXX, XX. Date of publication XX, XXXX, XX; date of current version XX, XXXX, XX. This work was supported in part by the Research Foundation – Flanders (FWO) Odysseus 1 under Grant G.0H94.18N; Methusalem Programme of the Ghent University Special Research Fund (BOF) under Grant 01M01021; and in part by the National Science and Technology Major Project, China, under Grant J2019-I-0001-0001 and Grant J2019-I-0019-0018. Michael Ruzhansky was also supported by Engineering and Physical Sciences Research Council (EPSRC) under Grant EP/R003025/2. (Corresponding author: Michael Ruzhansky.)}
	\thanks{Junqing Huang, Andreas Weiermann, Michael Ruzhansky are with the Department of Mathematics: Analysis, Logic and Discrete Mathematics, Ghent University, 9000 Ghent, Belgium; Michael Ruzhansky is also with the School of Mathematical Sciences, Queen Mary University of London, E1 4NS London, UK (e-mail: \{Junqing.Huang, Michael.Ruzhansky\} @UGent.be).}
	\thanks{Haihui Wang is with the School of Mathematical Sciences, Beihang University (BUAA), China (e-mail: whhmath@buaa.edu.cn).}
	\thanks{Junqing Huang and Haihui Wang contributed equally to this work.}
	\thanks{Digital Object Identifier no. XX.XXXX/TIP.XXXX.XXXXXXX.}}
\begin{document}

\IEEEtitleabstractindextext{
	
    \begin{abstract}
    In this paper, we derive a novel optimal image transport algorithm over sparse dictionaries by taking advantage of Sparse Representation (SR) and Optimal Transport (OT). Concisely, we design a unified optimization framework in which the individual image features (color, textures, styles, etc.) are encoded using sparse representation compactly, and an optimal transport plan is then inferred between two learned dictionaries in accordance with the encoding process. This paradigm gives rise to a simple but effective way for simultaneous image representation and transformation, which is also empirically solvable because of the moderate size of sparse coding and optimal transport sub-problems. We demonstrate its versatility and many benefits to different image-to-image translation tasks, in particular image color transform and artistic style transfer, and show the plausible results for photo-realistic transferred effects. 
    \end{abstract}

}
\maketitle

\IEEEraisesectionheading{\section{Introduction}
	\label{sec:introduction}}

\IEEEPARstart{I}{mage}-to-image translation is an interesting but rather challenging image synthesis problem in image processing and computer vision fields. Recent studies have shown that many image-to-image translation tasks can be identically posed as a special image transportation problem within the context of optimal transport framework. For example, image color matching~\cite{reinhard2001color, morovic2003accurate} and transfer~\cite{pitie2007linear, ferradans2014regularized, rabin2014adaptive, frigo2014optimal, hristova2015style} are easily formulated into an optimal transport problem by inferring a mapping between image color distributions. Image super-resolution~\cite{yang2010image, wang2012semi} can be viewed as to find an optimal mapping between images with different scales or resolutions. Similarly, the more complex image texture synthesis~\cite{efros2001image, gatys2015texture}, non-photorealistic rendering~\cite{gooch2001non, kang2008flow} and artistic stylization~\cite{kyprianidis2012state, winnemoller2012xdog, gatys2016image, elad2017style, gatys2015texture, li2017universal} are essentially aim to infer transportation maps between the abstract semantic features (saturation, textures, styles, etc.) when considered them in optimal transport context. Despite the varying backgrounds, forms and generalizations, they in nature share a very similar goal --- that is, automatically converting an image from one domain to another while preserving the semantic styles or contents for either better interpretation or visual-pleasant purposes.

In a general sense, image translation problem can be formed as an admissible map in latent feature spaces while preserving the interest of contents or information (color, texture or styles, etc.). In nature, it is necessary to solve two fundamental sub-problems, that is, image encoding and feature transformation. The former is to seek a useful image representation tool to extract the discriminative or individual image styles, while the latter aims to infer an appropriate mapping for the encoded image styles while maintaining abstract information such as image structures, textures and high-level semantic characteristics. As illustrated hereafter, many vision-based tasks can be understood from the two sub-problems. The difference mainly underpins the process of image encoding and feature transformation. Image color matching~\cite{reinhard2001color, ferradans2014regularized, rabin2014adaptive}, for example, tends to take image intensities (or, hue and saturation) as a meta-representation and solves a mapping problem between color palettes to determine the transferred results. In contrast, it is crucial to have both concisely-designed image encoding and feature transformation for artistic image stylization  in view of the complexity of abstract styles. In many scenarios, it is also important for the encoding process to have a compact form for the sake of reducing computational cost, while the translation problem involves a special transport map between two distributions of individual image features. It has also witnessed recent efforts to address the two sub-problems for different image style transfer applications, including the ever-increasing deep learning-based methods ~\cite{dong2014learning, gatys2015texture, gatys2016image, huang2017arbitrary}. Despite the great success, there is still considerable interest to exploit more easy-configured and powerful tools to achieve more visual-appealing results. 

\begin{figure*}[!t]
\begin{center}
    \centering
    \captionsetup{type=figure}
    \begin{subfigure}{0.18\linewidth}
        \includegraphics[width = \textwidth, height=0.16\textheight]{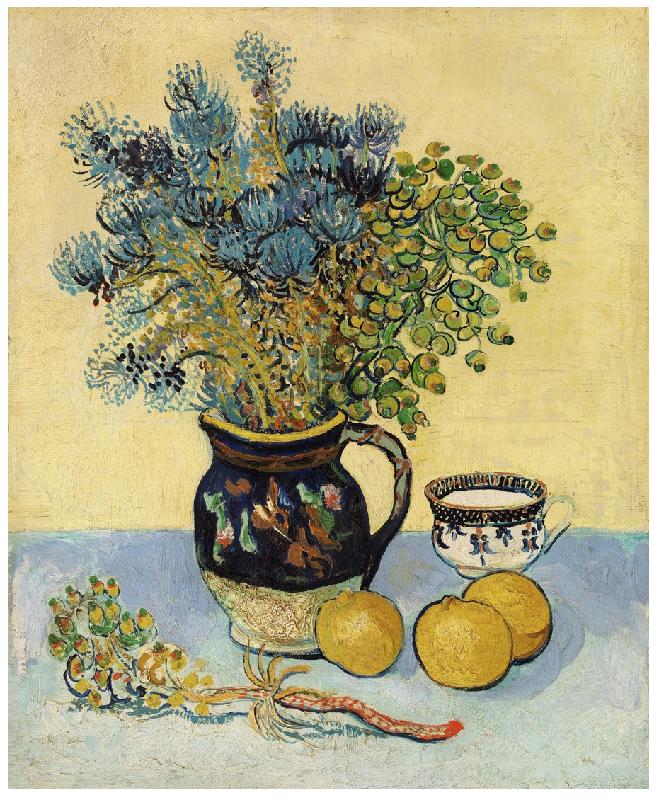}
        \caption{}
        \label{fig1-a}
    \end{subfigure}
    \hfill
    \begin{subfigure}{0.18\linewidth}
        \includegraphics[width = \textwidth, height=0.16\textheight]{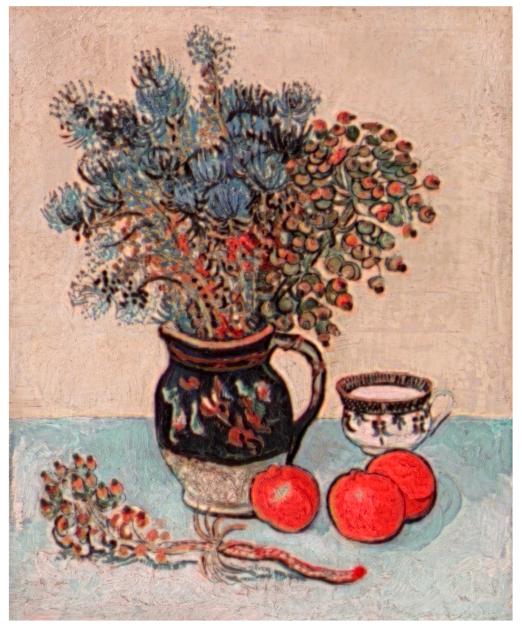}
        \caption{}
        \label{fig1-b}
    \end{subfigure}
    \hfill
    \begin{subfigure}{0.12\linewidth}
        \includegraphics[width = \textwidth, height=0.16\textheight]{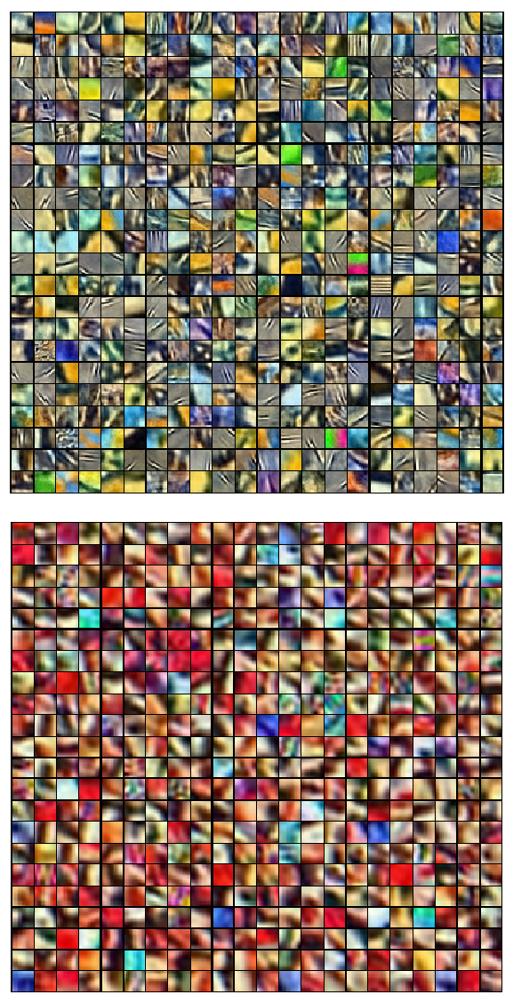}
        \caption{}
        \label{fig1-c}
    \end{subfigure}
    \hfill
    \begin{subfigure}{0.24\linewidth}
        \includegraphics[width = \textwidth, height=0.16\textheight]{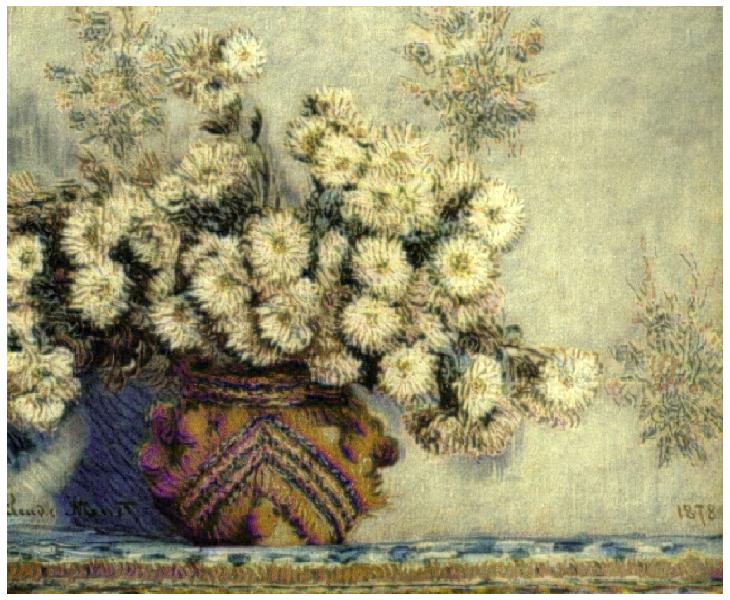}
        \caption{}
        \label{fig1-d}
    \end{subfigure}
    \hfill
    \begin{subfigure}{0.24\linewidth}
        \includegraphics[width = \textwidth, height=0.16\textheight]{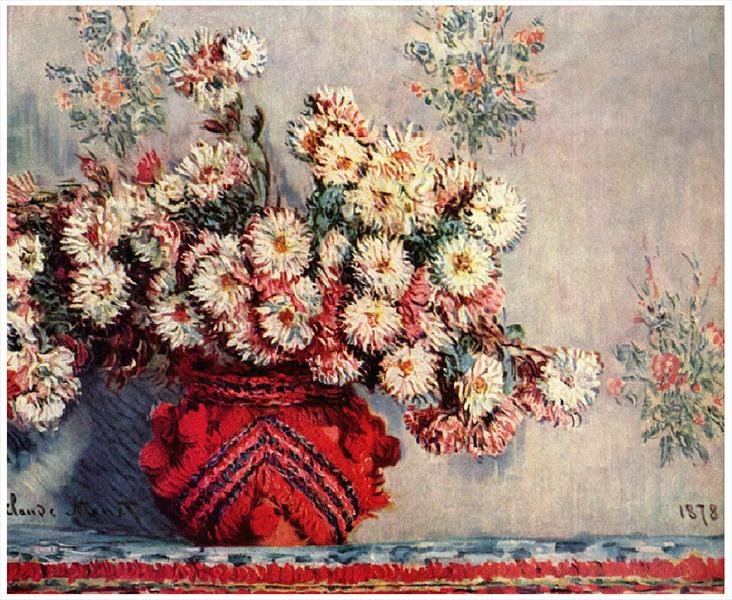}
        \caption{}
        \label{fig1-e}
    \end{subfigure}
	\caption{An illustration of optimal image transfer on sparse dictionaries. Given a content image Fig. \ref{fig1-a} and reference image Fig. \ref{fig1-e}, the proposed method learns the individual dictionaries Fig. \ref{fig1-c} and then derives an optimal transport plan over the learned dictionaries, giving the transferred results Fig. \ref{fig1-b} and Fig. \ref{fig1-d}, respectively.}
	\label{fig1}
\end{center}%
\end{figure*}

In this paper, we propose a novel approach for image-to-image translation, in which an optimal transport map is directly posed on sparse dictionaries learned from sparse image coding. Specifically, sparse representation is applied as a feature extractor to encode the latent features of images. Optimal transport is subsequently inferred over the learned dictionaries to provide an optimal styles-swapping plan in accordance with the style encoding process. This new model, as shown in \ref{fig1}, inherits two-fold benefits of sparse representation and optimal transport. On the one hand, sparse representation provides us a maneuverable and easy-understood image editing tool to encode the semantic features of images, as it has been demonstrated in a variety of successful applications such as image denoising~\cite{aharon2006k} and super-resolution ~\cite{yang2010image}. On the other hand, optimal transport allows us to swap the encoded features or styles based on a linear mapping. Moreover, the size of learned dictionaries is moderate in practice, which also helps to alleviate the high computational cost of the native optimal transport. We will illustrate that this new paradigm, with a slight relaxation, is empirically solvable and gives rise to a closed-form solution to many image translation problems. We also demonstrate its versatility and many benefits to image-to-image translation with two typical tasks: color transform and artistic style transfer, and show their high-quality transferred results.

Our main contributions are summarized as follows:

\begin{itemize}
	\setlength{\itemsep}{0pt}
	\setlength{\parsep}{0pt}
	\setlength{\parskip}{0pt}
	
	\item  We recall a typical of image-to-image translation tasks and interpret them that can be cast into an optimal transport context by infer an transportation map between some abstract semantic features (saturation, color, textures, styles, etc.) 
	
	\item A generalized optimization framework is concisely designed for image-to-image translation by taking advantage of both sparse representation and optimal transport, which provides a simultaneous image representation and transformation tool for a wide range of vision-based tasks.  
	
	\item  We present an alternative solution by decomposing the proposed optimization problem into three sub-problems: sparse coding, learning style dictionaries and optimal transport with a series of relaxation. Since each sub-problem can be efficiently solved with standard algorithms, which provides a practical solution for the proposed optimization problem. 
	
	\item  We demonstrate the versatility and many benefits of the proposed method to image-to-image translation with two typical tasks: color transform and artistic style transfer, and show their high-quality transferred results. 
	
\end{itemize}

We further conclude the merit of simultaneous image representation and transformation beneficial from sparse representation and optimal transport. On the one hand, sparse representation provides a relative simple but effective encoding tool to represent image low-level or semantic image features such as image color, saturation, textures, styles, etc. On the other hand, the transportation mapping over sparse dictionaries significantly reduces the computational cost due to the small size of learned dictionaries. Due to the two-folds of benefits, the proposed method give arise to a practical tool for a wide range of image-to-image tasks, such as image color matching and transfer, super-resolution, texture synthesis, artistic stylization, and so on.

\section{Related work}

Image-to-image translation, as aforementioned, covers a wide range of vision-based tasks despite their different backgrounds and generalizations. We briefly review some existing color transform and artistic style transfer methods, in particular the ones for photo-realistic results because of the close connections to the proposed optimal style transfer on sparse dictionaries.

Color matching or transfer is a typical image-to-image translation application keen on photo-realistic results. The purpose is straightforward, that is, to alter color appearance of an image based on a reference image~\cite{hristova2015style, reinhard2001color, xiao2006color}.  In the early stage,  color transfer is usually posed as a one-dimensional histogram matching problem between two color distributions --- for example, histogram equalization or specification. As suggested in the pioneering work~\cite{reinhard2001color}, color transfer is implemented by matching their global statistical mean and covariance of two images. Such a strategy is then extended to other color space~\cite{morovic2003accurate, xiao2006color} or combined with the lightness and brightness information~\cite{hristova2015style}. They, however, may produce non-harmonic results because of the non-consistent color distributions of natural images. Other methods~\cite{faridul2014survey, hristova2015style, tai2005local} also resort to some color segmentation techniques for local color matching, while such a strategy is highly dependent on the semantic constraints for color segmentation in practice.

To reduce the notorious non-harmonic artifacts, recent advances based on optimal transport~\cite{ferradans2014regularized, pitie2007linear} have gained great attention in color transfer applications. The work can be dated back to the study of histogram-matching problems and the relation to optimal transport for gray images~\cite{delon2004midway, morovic2003accurate}. Such a strategy is then extended to color images and videos. The assigned color could more or less avoid the undesired visual artifacts. It is worth noting that the transport map is mostly deduced from the discrete samplings, the solution may be not reachable for a very large-scale problem, for example, in the cases of using optimal transport, where a naive optimal transport has a $\mathcal{O}{(n^3)}$ complexity for $n$ pair samples. More recently, the relaxed and regularized OT methods are also explored for color transfer to tackle the high computational cost. However, as pointed out in ~\cite{ferradans2014regularized, rabin2014adaptive}, an exact transform of color distributions is not enough in practical applications because color densities may have very different shapes and outliers. As a consequence, the transfer performance may be limited by the sampling and interpolation processes.

Simultaneously, image style transfer --- which is mostly dedicated to non-photorealistic image rendering for artistic effects, has been extensively studied for the long-standing dream of generating attractive artworks automatically. Most traditional methods are either based on line-drawing and stroke-based rendering techniques to produce the prescribed effects, including image stippling~\cite{kim2009stippling},  pencil sketching~\cite{kang2008flow, winnemoller2012xdog}, watercolor~\cite{bousseau2006interactive}, oil painting~\cite{gooch2002artistic, hertzmann1998painterly}, and so on. As shown in difference-of-Gaussians (DoG) operator~\cite{winnemoller2012xdog} and flow-based filtering~\cite{kang2008flow}, they boost the salient line features or main structures of images, and help to yield aesthetically pleasing lines when synthesizing line drawings and cartoon-like art effects. The stroke-based rendering technique is another prevalent strategy for artistic image stylization~\cite{kang2008flow, winnemoller2012xdog}, in which the brush strokes are iteratively aligned according to the variants of local color, size, and orientation information. With careful design, it is possible to generate high-quality results for some prescribed styles but may be limited in style diversity. The reader is referred to the surveys~\cite{gooch2002artistic, winnemoller2012xdog, kyprianidis2012state} for more details.

More recently, it has also witnessed the great success of neural style transfer with the renaissance of deep learning methods. In the pioneering work~\cite{gatys2016image}, for example, a novel iterative optimization scheme over a conventional neural network is proposed to match the learned features within a pre-trained classification network. The idea is subsequently developed by many deep learning methods for more efficient stylization~\cite{huang2017arbitrary, xia2021real}, stroke-based paintings~\cite{liu2021paint, tong2022im2oil, zou2021stylized} and universal style transfer~\cite{deng2022stytr2, gu2018arbitrary, hong2021domain, huang2023quantart, isola2017image, johnson2016perceptual, kotovenko2019content, li2017universal, liu2017unsupervised}. In the last few years, it has also witnessed many efforts for photo-realistic style transfer~\cite{an2020ultrafast, ho2021deep, ke2023neural, luan2017deep, li2018closed, lu2019closed}. Despite the impressive artistic effects, they may suffer from some unpredictable effects with spatial distortions and artifacts which are not consistent with semantic interpretations or should not happen in real photographs, since it is still not comprehensively understood the mechanism of the deep encoding process. Recent studies have shown that the matching of features can be formulated as an optimal transport problem between the learned features for more favorable results~\cite{kolkin2019style, mroueh2020wasserstein}. The achievement of deep learning methods is largely due to the two-fold benefits of many deep learning architectures --- that is, the encoding ability of neural networks offers a powerful and ubiquitous tool for high-level visual features, and the transformation map between deep features is also learnable during the training process. Moreover, many deep learning methods are usually limited by the availability of very large-scale training datasets. The reader is also referred to the work~\cite{isola2017image, jing2019neural, zhu2017unpaired} for more details of deep learning-based techniques.

\section{Preliminary} 

In this section, we briefly introduce sparse representation and optimal transport, as they form the key ingredients of the proposed model for image style (feature) encoding and transformation.

\begin{figure}[t]
    \begin{center}
        \includegraphics[width=0.5\textwidth]{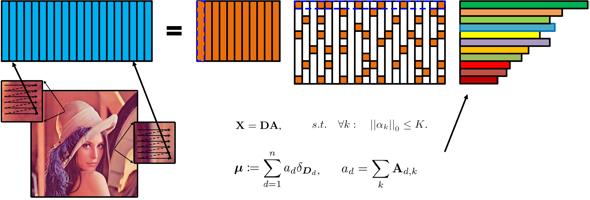}
    \end{center}
    \caption[Sparse representation]{Sparse representation of an image with the distribution of dictionary}
    \label{fig2} 
\end{figure}

\subsection{Sparse Representation}

Sparse and redundant representation~\cite{aharon2006k, rubinstein2010dictionaries} has been used as a simple and important method for signal/image analysis and processing, in which the signal/image is assumed to be compactly approximated by a linear combination of a few fundamental elements --- known as a basis set or a dictionary. One of the overwhelming benefits of sparse representation is to reduce the size of large-scale problems in signal/image processing fields since the majority of information is encoded by a small set of basis functions weighted by sparse coefficients. In image processing and computer vision fields, sparse representation provides an effective image editing tool to encode the latent middle-level vision features of images. The basis or dictionary of sparse representation can be either selected from a group of pre-defined functions such as discrete cosine transform (DCT) and wavelet or learned from training data. We here consider the learning-based dictionaries for better performance. 

Mathematically, let $\mathbf{X}=\{\boldsymbol{x}_i\}_{i=1}^N$,$ \boldsymbol{x}_i \in \mathbb{R}^{d}$ be a set of data samplings, for example, the vectorized image patches, sparse representation then aims to discover a group of dictionary vectors $\{\boldsymbol{d}_j\}_{j=1}^n, \boldsymbol{d}_i \in \mathbb{R}^{d},n \ll N$ (or, denoted as $\mathbf{D}= [\boldsymbol{d}_1, \cdots, \boldsymbol{d}_n]$) associated with the efficient matrix $\mathbf{A}\in \mathbb{R}^{n \times N}$, which can be written as,
\begin{equation}
\begin{aligned}
\mathbf{X} = \mathbf{D}\mathbf{A} \quad \text{s.t.} \left\|\boldsymbol{\alpha}_i\right\|_{0} \leq K,
\end{aligned}
\label{eq1}
\end{equation}
where $\boldsymbol{\alpha}_i \in \mathbb{R}^{n}$ denotes the represented coefficients of sampling $\boldsymbol{x}_i$, corresponding to the $i$-th column vector of the coefficient matrix $\mathbf{A}$, $\left\|\cdot\right\|_{0}$ is known as pseudo $L_0$-norm counting the non-zero elements of a vector. The constraint $\left\|\boldsymbol{\alpha}_i\right\|_{0} \leq K$ suggests that the number of non-zero entries in $\boldsymbol{\alpha}_i$ is no more than $K$. In other words, the coefficient $\mathbf{A}$ has sparse characteristics, the assumption of which forms the nature of sparse representation.

Despite the simple form, it is generally a challenging problem to give a direct solution for sparse representation because both dictionary $\mathbf{D}$ and sparse coefficient matrix $\mathbf{A}$ in Eq. \ref{eq1} are unknown in advance. Moreover, the solution is always not unique when solving one by fixing another. The problem is also known as an NP-hard in view of the non-convex $L_0$-norm constraint. In practice, it always resorts to some approximate algorithms such as the method of optimal directions (MOD)~\cite{engan1999method}, generalized PCA~\cite{vidal2005generalized} or K-SVD algorithm~\cite{aharon2006k} to give a solution. Other methods for approximate solutions may be based on relaxation techniques, for example, replacing the non-convex $L_0$ constraint with its convex $L_1$ approximation.

The choice of an appropriate dictionary is also crucial to sparse representation for many practical applications. We here consider an example in Fig. \ref{fig2} for interpretation.  The process starts with the cropped image patches that are randomly sampled from an image or a dataset. The data matrix $\mathbf{X}$ is formed by concatenating these vectorized patches, and the dictionary $\mathbf{D}$ and coefficient $\mathbf{A}$ can be achieved by solving Eq. \ref{eq1} accordingly. Before diving deeper, we point here out that the row sum of the coefficient matrix is important to the proposed method, as it measures the frequency of each dictionary atom occurring in an image. Mathematically, it provides a probability distribution of image dictionary atoms. We will illustrate how to learn a pair of coupled dictionaries to represent the abstract styles of images in accordance with the optimal transport between learned dictionaries.

\subsection{Optimal Transport}
\label{subsec3:optimal_transport}

Optimal transport is also a well-developed mathematical theory~\cite{villani2021topics}, which can be traced back to Monge’s problem and then discovered under different backgrounds~\cite{peyre2019computational, villani2021topics}. We here review the Monge problem and its Kantorovitch relaxation for the sake of complementary.

\textbf{The Monge's Problem}: Let $\mu, \nu$ be two probability measures on two metric spaces $\mathcal{X} \in \mathbb{R}^{n},\mathcal{Y} \in \mathbb{R}^{m}$, and given a cost function $c(x, y): \mathcal{X} \times \mathcal{Y} \rightarrow [0, \infty]$, which represents the effort of transporting the mass from $x \in \mathcal{X}$ to $y \in \mathcal{Y}$, the Monge's formulation aims to find a transport map $T: \mathcal{X} \rightarrow \mathcal{Y}$, realizing the infimum of the function:
\begin{equation}
\begin{aligned}
\inf_{{T_{\sharp}\mu=\nu} }{\int_{\mathcal{X}}{c(\mu,T(\mu)) d \nu(x)}},
\end{aligned}
\label{eq2}
\end{equation}
where ${T_{\sharp}\mu\stackrel{\text{def}}{=} \nu}$ is known as the \textit{push-forward} operator that pushes forward the mass of $\mu$ to $\nu$~\cite{oberman2015efficient, peyre2019computational}. The transport map $T$ attains when reaching the infimum, the existence of which in practice, however, is not always guaranteed, for example, when only one of $\mu$ and $\nu$ is a Dirac function. 

\textbf{The Kantorovitch Relaxation}: Alternatively, a simple relaxation of Monge's problem initiated by Kantorovich is guaranteed to have a solution. The key idea is that the mass at any point of
$x$ can be potentially dispatched across several locations of $y$. The equivalent Kantorovitch formulation of the optimal transport seeks for a probabilistic coupling $\pi \in \mathcal{P}\left(\mathcal{X} \times \mathcal{Y} \right)$ between $\mathcal{X}$ and $\mathcal{Y}$~\cite{peyre2019computational}:
\begin{equation}
\begin{aligned}
\inf_{\pi \in \Pi} \int_{\mathcal{X} \times \mathcal{Y}} c(x, y) \mathrm{d} \pi(x, y),
\end{aligned}
\label{eq3}
\end{equation}
where $\Pi\stackrel{\text{def}}{=}\left\{\pi \in\left(\mathbb{R}^{\!+\!}\right)^{\mathcal{X} \times \mathcal{Y}} \mid \pi_{\mathcal{X} }=\mu, \pi_{\mathcal{Y} }=\nu\right\}$ is the set of transportation plans with the joint distribution of marginals $\mu$ and $\nu$. In this formulation, $\pi$ can be understood as a joint probability measure with marginals $\mu$ and $\nu$. The cost function $c(x,y)$ can be chosen, for instance, as Euclidean distance between two locations $x$ and $y$, while other types of metrics could be considered, such as Riemann distances over a manifold.

\textbf{Discrete Case}: In practice, the distributions $ \mu$ and $\nu$ are always accessible through discrete samples, which leads to a discrete optimal transport problem~\cite{merigot2016discrete, peyre2019computational}. Considering two discrete probability measures $\mu \!\coloneqq \!\sum^m_{i=1} \boldsymbol{a}_i {\delta}_{\boldsymbol{x}_i}$ and $\nu \!\coloneqq \!\sum^n_{j=1} \boldsymbol{b}_j {\delta}_{\boldsymbol{y}_j}$ sampled from the source and target samples $\left\{\boldsymbol{x}_{i}\right\}_{i=1}^{m}$, $\left\{\boldsymbol{y}_{j}\right\}_{j=1}^{n}$ with $\boldsymbol{x}_{i}, \boldsymbol{y}_{j} \in \mathbb{R}^d$, it is straightforward to rewrite the discrete Kantorovitch optimal transport as,
\begin{equation}
\begin{aligned}
\min_{\mathbf{T} \in \Pi(\boldsymbol{a},\boldsymbol{b})} \langle \mathbf{C}, \mathbf{T} \rangle \stackrel{\text{def}}{=} \min_{\mathbf{T} \in \Pi(\boldsymbol{a},\boldsymbol{b})}{\sum_{i,j} \mathbf{C}_{i,j} \mathbf{T}_{i,j}}
\end{aligned}
\label{eq4}
\end{equation}
where $\mathbf{T}$ is a coupling matrix with entries $\mathbf{T}_{i,j}$ describing the amount of mass flowing from $\boldsymbol{a}_i$ to $\boldsymbol{b}_j$ and $\Pi(\boldsymbol{a},\boldsymbol{b}) \stackrel{\text{def}}{=} \{ \mathbf{T} \in \mathbb{R}_+^{m\times n} \vert \mathbf{T}\boldsymbol{1}_n =\boldsymbol{a},   \mathbf{T}^\top\boldsymbol{1}_m = \boldsymbol{b} \}$. The cost matrix  $\mathbf{C}\in \mathbb{R}^{m \times n}$ has the entries $\mathbf{C}_{\boldsymbol{i,j}} = c(\boldsymbol{x}_i,\boldsymbol{y}_j)$ specifying the transport effort between the location pair $(\boldsymbol{x}_i, \boldsymbol{y}_j)$.  

It is well-known that Eq. \ref{eq4} can be rewritten into a special linear program problem and solved using linear solvers such as network flow solver or transportation simplex~\cite{dantzig2016linear}. Despite the simple form, the linear solvers are also computationally expensive, especially for a large-scale case --- for example, having $\mathcal{O}{(n^3)}$ complexity for $n$ pair samples with network flow solvers. In many practical tasks~\cite{perrot2016mapping, peyre2019computational}, the Sinkhorn's algorithm~\cite{dvurechensky2018computational} is always chosen as a faster alternative method to solve such a discrete optimal transport approximately.

\begin{figure*}[!t]
	\centering
	\begin{subfigure}{0.17\textwidth}
		\includegraphics[width=\textwidth]{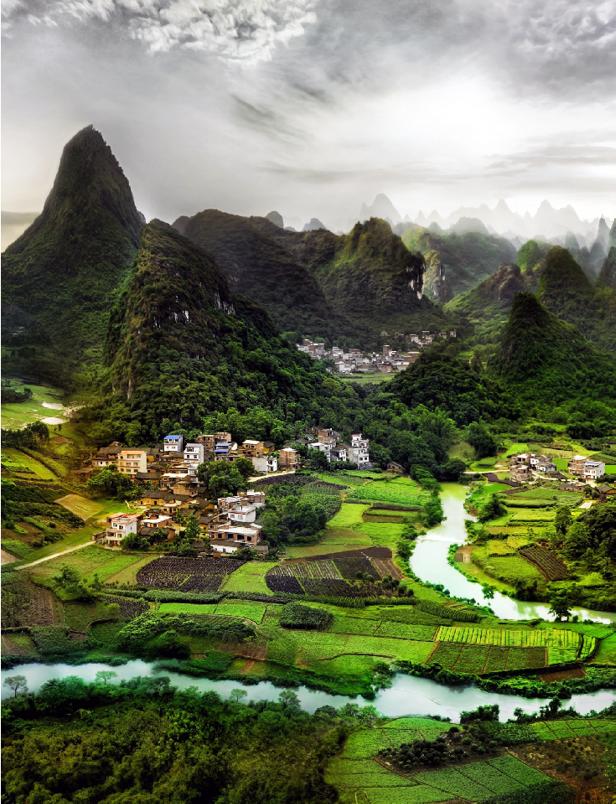}
		\caption{\makecell{Content image}}
		\label{fig3-a}
	\end{subfigure}
	\hfill
	\begin{subfigure}{0.17\textwidth}
		\includegraphics[width=\textwidth]{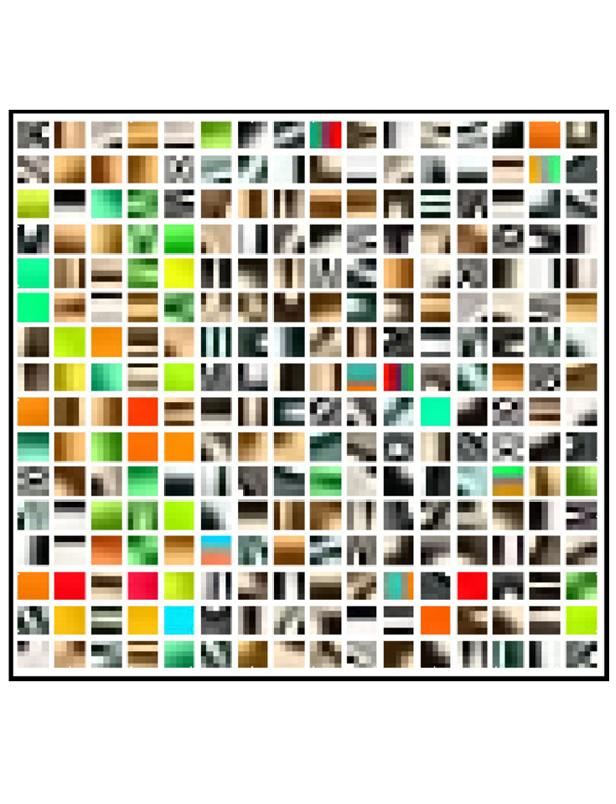}
		\caption{\small{Content dictionary}}
		\label{fig3-b}
	\end{subfigure}
	\hfill
	\begin{subfigure}{0.17\textwidth}
		\includegraphics[width=\textwidth]{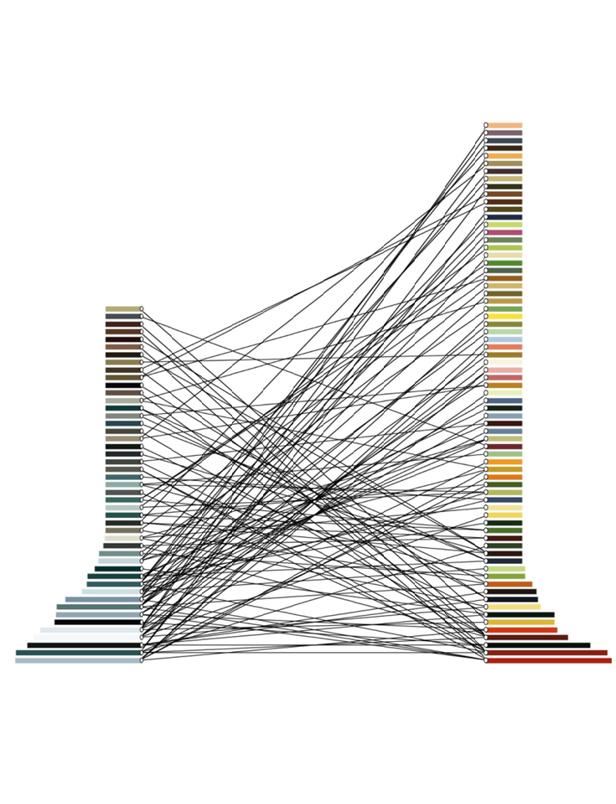}
		\caption{\makecell{Transport map}}
		\label{fig3-c}
	\end{subfigure}
	\hfill 
	\begin{subfigure}{0.17\textwidth}
		\includegraphics[width=\textwidth]{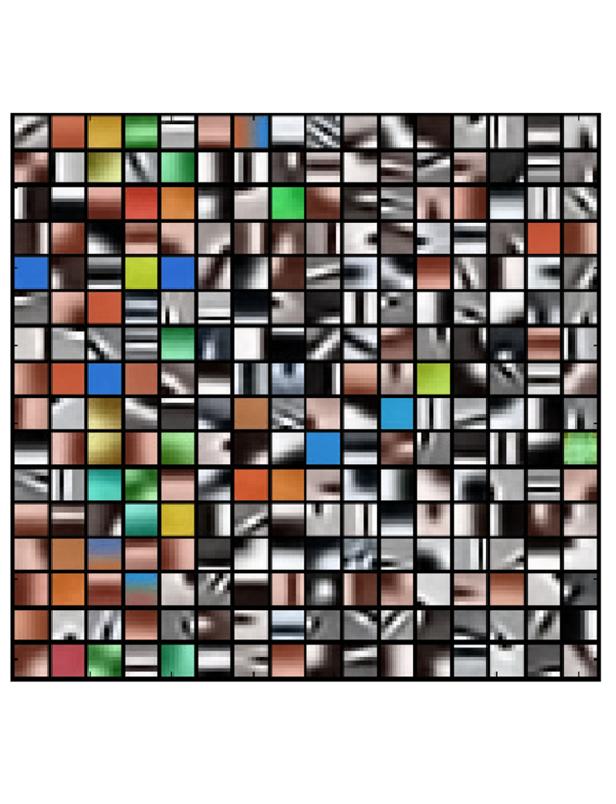}
		\caption{\makecell{Style dictionary}}
		\label{fig3-d}
	\end{subfigure}
	\hfill
	\begin{subfigure}{0.17\textwidth}
		\includegraphics[width=\textwidth]{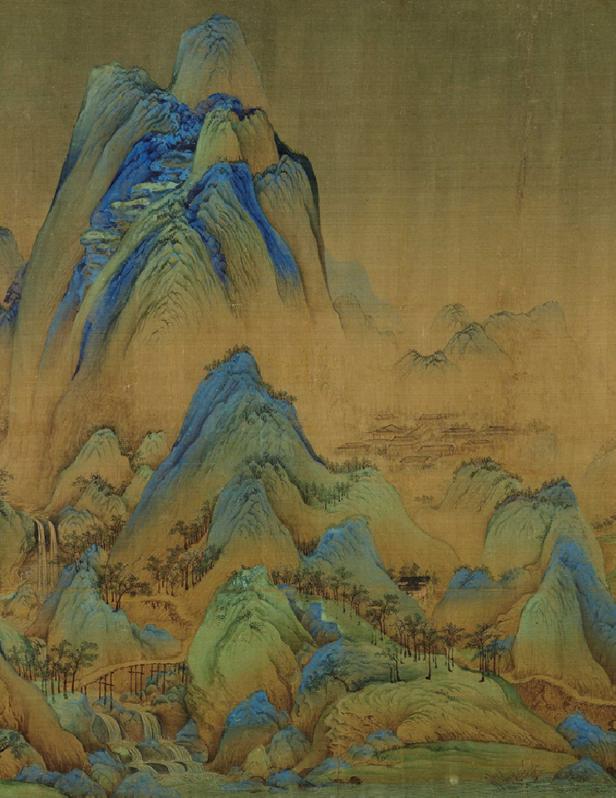}
		\caption{\makecell{Reference image}}
		\label{fig3-e}
	\end{subfigure}
	\caption{An illustration of optimal transport over two dictionaries. Given the content and reference images, the proposed method firstly learns the individual dictionaries and then derives an optimal transport map between the learned dictionaries.}
	\label{fig3}
\end{figure*}

\section{Optimal Image Transfer}
\label{sec4:optimal_style_transfer}

We suggest that image-to-image translation problems can be implemented by means of sparse representation and optimal transport. Image transfer, as pointed out, aims to solve the feature encoding and  transformation problems. We illustrate that sparse representation provides an effective tool to encode the individual and discriminate features of different images since it has been used as a feature extractor in many image processing tasks. Sparse coefficients can be viewed as a counting process of dictionary elements, as shown in Fig. \ref{fig2}, which gives a probability measure to weigh the importance of each style element. Consequently, a transport plan between the encoded features is attainable and efficiently computed based on optimal transport under the small size of learned dictionaries (See Fig. \ref{fig3}).

\subsection{Problem Formulation} 

For simplicity, we take into account an image transfer problem between two images $x$ and $y$ with the features or styles $s_x$, $s_y$, respectively. Without loss of generality, let $\mathbf{X}=\{\boldsymbol{x}_i\}_{i=1}^M, \mathbf{Y}=\{\boldsymbol{y}_j\}_{j=1}^N, \boldsymbol{x}_i,\boldsymbol {y}_j \in \mathbb{R}^{d}$ be the vectorized patches sampled from image $x$ and $y$, the latent styles $s_x$ and $s_y$ can be expressed by sparse dictionaries given by, 
\begin{equation}
\begin{aligned}
\begin{cases}
&\mathbf{X} = \mathbf{D}^x\mathbf{A}, \qquad   s.t. \quad   {\lvert\lvert \boldsymbol{\alpha}_i \rvert\rvert}_0 \leq K_1,\\
&\mathbf{Y} = \mathbf{D}^y\mathbf{B}, \qquad   s.t. \quad   {\lvert\lvert \boldsymbol{\beta}_j \rvert\rvert}_0 \leq K_2,  
\end{cases}
\end{aligned}
\label{eq5}
\end{equation}
where $\mathbf{D}^x \!=\! [\boldsymbol{d}_1^x, \cdots, \boldsymbol{d}_m^x]$ and $\mathbf{D}^y\!=\! [\boldsymbol{d}_1^y, \cdots, \boldsymbol{d}_n^y]$ are the style dictionaries. We assume the entries $\boldsymbol{d}_i^x,\boldsymbol{d}_j^y \in \mathbb{R}^{d}$ are in the same space. $\mathbf{A} \in \mathbb{R}^{d\times M}, \mathbf{B} \in \mathbb{R}^{d\times N}$ contain the coefficient vectors $\boldsymbol{\alpha}_i$ and $\boldsymbol{\beta}_j$ of the $i$-th and $j$-th samplings $\boldsymbol{x}_i$ and $\boldsymbol{y}_j$. The constraints suggest the weights $\boldsymbol{\alpha}_i$ and $\boldsymbol{\beta}_j$ tend to be sparse --- that is, the number of non-zero entries is less than the positive $K_1$ (or, $K_2$).

The dictionaries $\mathbf{D}^x$ and $\mathbf{D}^y$, in general cases, may not exactly encode the styles $s_x$ and $s_y$, while, as demonstrated hereafter, it is also enough to provide favorable transferring results for image transfer tasks. The use of sparse/redundant representation, on the one hand, is to find a compact and effective image editing tool for the latent styles. Notice that the size of dictionaries is always much less than samplings in practice, thus optimal transport between two dictionaries $\mathbf{D}^x$ and $\mathbf{D}^y$ is affordable even using linear program solver~\cite{dantzig2016linear}, which is significantly reduced the computational cost compared with the naive case over the samplings $\mathbf{X}$ and $\mathbf{Y}$. On the other hand, it is reasonable to assume that the weights of each element in learned dictionaries indicate the contribution of the latent style to an image. Notice also that the row sum of each row of coefficient matrices counts the total contribution of each style element. Let $\boldsymbol{a}, \boldsymbol{b}$ be the row sum of coefficient matrices, we have $\boldsymbol{a} = \boldsymbol{A}\boldsymbol{1}_M, \boldsymbol{b}  =\boldsymbol{B}\boldsymbol{1}_N$, where $\boldsymbol{1}_M(N)$ is the vector with all $M(N)$ entries being value 1; and  two discrete probability distributions $\mu \coloneqq \sum^n_{i=1} \boldsymbol{a}_i {\delta}_{\boldsymbol{d}^x_i}$ and $\nu \coloneqq \sum^m_{j=1}\boldsymbol{b}_j {\delta}_{\boldsymbol{d}^y_j}$ of the learned dictionaries, where $\boldsymbol{a}_k$ and $\boldsymbol{b}_k$ are the $k$-th element of $\boldsymbol{a}$ and $\boldsymbol{b}$, and $\delta{(\cdot)}$ is the Dirac function. Recalling the Kantorovich relaxation of the transport problem in Sec. \ref{subsec3:optimal_transport}, we have an optimal transport on the learned dictionaries,
\begin{equation}
\begin{aligned}
\min_{\mathbf{T} \in \Pi(\boldsymbol{a},\boldsymbol{b})} \langle \mathbf{C}, \mathbf{T} \rangle \stackrel{\text{def}}{=} {\sum_{i,j} \mathbf{C}_{i,j} \mathbf{T}_{i,j}},
\end{aligned}
\label{eq6}
\end{equation}
where $\mathbf{C}_{i,j} \stackrel{\text{def}}{=} c(\boldsymbol{d}^x_i, \boldsymbol{d}^y_j)$ is the ground cost function to move the dictionary element $\boldsymbol{d}^x_i$ to $\boldsymbol{d}^y_j$, and the transport mapping function $\mathbf{T}$ satisfies, 
\begin{equation}
\begin{aligned}
\Pi(\boldsymbol{a},\boldsymbol{b}) \stackrel{\text{def}}{=} \{ \mathbf{T} \in R_+^{m\times n} \vert  \mathbf{T}\boldsymbol{1}_n = \boldsymbol{a},   \mathbf{T}^\top\boldsymbol{1}_m = \boldsymbol{b}\}.
\end{aligned}
\label{eq7}
\end{equation}

In view of the above notations, we now interpret that the optimal style transfer over the learned style dictionaries, parameterized by $\mathbf{T} \in R_+^{n\times m}$,  is generalized into,
\begin{equation}
\begin{tabular}{c r l l}
		\multicolumn{4}{c}{${\displaystyle \min_{\mathbf{T}} \langle \mathbf{C}, \mathbf{T} \rangle  \stackrel{\text{def}}{=}\displaystyle \min_{\mathbf{T}} {\sum_{i,j} \mathbf{C}_{i,j} \mathbf{T}_{i,j}}}$,} \\
$s.t.$ 	& $\mathbf{D}^x\mathbf{A} =\mathbf{X} $,           & $ {\lvert\lvert \boldsymbol{\alpha}_i \rvert\rvert}_0 \leq K_1$, 		& 	 \\
		& $\mathbf{D}^y\mathbf{B} = \mathbf{Y} $,          & ${\lvert\lvert \boldsymbol{\beta}_i \rvert\rvert}_0 \leq K_2$, 			& 	 \\
		& $\mathbf{A}\boldsymbol{1}_M=\boldsymbol{a}$,     & $\mathbf{B}\boldsymbol{1}_N =\boldsymbol{b} \: $,		                &    \\
		& $\mathbf{T}\boldsymbol{1}_n = \boldsymbol{a}$,   & $\mathbf{T}^\top\boldsymbol{1}_m = \boldsymbol{b}$. 	                & 	
\end{tabular}
\label{eq8}
\end{equation}

It is clear from Eq. \ref{eq8} that the objective function aims to infer an optimal transport plan between the learned style dictionaries, in which the first and second constraints are sparse representations for images, and the last two constraints specify the property distributions of dictionaries and the necessary conditions of transport plan. Despite the simple form of Eq. \ref{eq8}, it is not easy to solve due to the $L_0$-norm constraints. In what follows, a relaxed model of Eq. \ref{eq8} is further discussed with an approximate solution under some mild assumptions.

\textbf{Relaxed Model:} As illustrated, image style transfer is formulated into an optimal transport over sparse dictionaries with both  sparse representation and optimal transport constraints. However, a direct solution to Eq. \ref{eq8} is not available due to two-fold facts: (1) the sparse coefficient constrains $ {\lvert\lvert \boldsymbol{\alpha}_i\rvert\rvert}_0 \leq K_1$ and ${\lvert\lvert \boldsymbol{\beta}_i\rvert\rvert}_0 \leq K_2$ are non-convex and difficult to solve in practice; and (2) the sub-problem with respect to the variable $\mathbf{A}$ is a typical Sylvester equation\cite{bartels1972solution} constrained by $\mathbf{D}^x\mathbf{A} =\mathbf{X}$ and $\mathbf{A}\boldsymbol{1}_M = \boldsymbol{a}$\footnote{The Sylvester equation has the form $AX+XB=C$, whose solution is computational expensive in case of a large-scale problem\cite{bartels1972solution}.}. As a result, it is necessary to reduce the problem for a more efficient solution.

Paying attention to $\mathbf{A}\boldsymbol{1}_M\!=\!\boldsymbol{a}$, $\mathbf{B}\boldsymbol{1}_N \!=\! \boldsymbol{b}$, we have  $\mathbf{X}\boldsymbol{1}_M \!=\! \mathbf{D}^x \boldsymbol{a}, \mathbf{Y}\boldsymbol{1}_N \!=\! \mathbf{D}^y \boldsymbol{b}$ by multiplying $\mathbf{D}^x $ and $\mathbf{D}^y$ in both sides of the third-line constraints in Eq. \ref{eq8}. As a result, a relaxed minimization problem can be written in the form,
\begin{equation}
    \begin{tabular}{c l l l}
    \multicolumn{4}{c}{${\displaystyle \min_{\mathbf{T}} \langle \mathbf{C}, \mathbf{T} \rangle  \stackrel{\text{def}}{=}\displaystyle \min_{\mathbf{T}} {\sum_{i,j} \mathbf{C}_{i,j} \mathbf{T}_{i,j}}}$,} \\
    $s.t.$ 	& $\mathbf{D}^x\mathbf{A} =\mathbf{X} $,           & $ {\lvert\lvert \boldsymbol{\alpha}_i \rvert\rvert}_0 \leq K_1$, 		& 	 \\
    & $\mathbf{D}^y\mathbf{B} = \mathbf{Y} $,          & ${\lvert\lvert \boldsymbol{\beta}_i \rvert\rvert}_0 \leq K_2$, 			& 	 \\
    & $\mathbf{D}^x\boldsymbol{a}=\mathbf{X}\boldsymbol{1}_M$,     & $\mathbf{D}^y\boldsymbol{b}=\mathbf{Y}\boldsymbol{1}_N$,		                &    \\
    & $\mathbf{T}\boldsymbol{1}_n = \boldsymbol{a}$,   & $\mathbf{T}^\top\boldsymbol{1}_m = \boldsymbol{b}$. 	                & 	
    \end{tabular}
    \label{eq9}
\end{equation}

This relaxation adores the constraints on sparse dictionaries instead of coefficients, leading to two-fold benefits. On the one hand, the constraints $\mathbf{D}^x\boldsymbol{a}=\mathbf{X}\boldsymbol{1}_M$,  and $\mathbf{D}^y\boldsymbol{b}=\mathbf{Y}\boldsymbol{1}_N$ in Eq. \ref{eq9} change the sparse coefficient constraints into style dictionaries constraints, which helps to learn more favorable style dictionaries. On the other hand, it reduces the sub-problem with respect to $\mathbf{A} $ (or, $\mathbf{B}$) into a standard sparse coding form, thereby avoiding the complex Sylvester equation. As interpreted hereafter, the relaxed problem can be approximately optimized using an alternative variable splitting method under the assumption of $p$-Wasserstein cost function.

\subsection{The Solution of $p$-Wasserstein Case} 

The cost function $\mathbf{C}_{\boldsymbol{i,j}} \stackrel{\text{def}}{=}  c(\boldsymbol{d}^x_i, \boldsymbol{d}^y_j)$ is important to the optimal transport plan~\cite{peyre2019computational}. It turns out that the transport plan always exists when taking into account $p (p \!\geq\! 1)$-Wasserstein distance $\boldsymbol{W}_p^p (\boldsymbol{\mu},\boldsymbol{\nu})$. For simplicity, we consider $p\!=\!2$ Wasserstein distance, where the cost function is defined as $\mathbf{C}_{{i,j}} {\Vert\boldsymbol{d}^x_i\!-\!\boldsymbol{d}^y_j \Vert}^2_2$, which measures the distance of a pair of dictionary elements $(\boldsymbol{d}^x_i, \boldsymbol{d}^y_j)$. Clearly, we have $\mathbf{C}_{\boldsymbol{i,j}}=0$ if $\boldsymbol{d}^x_i = \boldsymbol{d}^y_j$.  

Recalling the relaxed model in Eq. \ref{eq9}, we first rewrite the constrained optimization problem into an unconstrained one based on regularization techniques and then apply the well-known alternative variable splitting algorithm to solve it approximately. By introducing the Lagrangian multiplier technique, the above problem can be reformulated into an unconstrained optimization problem and solved via an alternative minimization scheme as follows:
\begin{equation}
\small
    \begin{aligned}
        \begin{split}
        \operatorname*{argmin}_{\{\mathbf{D}^x,\mathbf{D}^y, \mathbf{A},\mathbf{B}, \mathbf{T}\}} & \gamma \sum_{i,j}  \mathbf{T}_{i,j} {\Vert\boldsymbol{d}^x_i\!-\!\boldsymbol{d}^y_j \Vert}^2_2  + {\big\Vert \mathbf{X}\!-\! \mathbf{D}^x \! \mathbf{A} \big\Vert}_{F}^2\!\!+\! {\big\Vert \mathbf{Y}\!-\! \mathbf{D}^y \mathbf{B} \big\Vert}_{F}^2 \\ + & \lambda_x {\big\Vert \mathbf{X}\boldsymbol{1}_M\!-\! \mathbf{D}^x \boldsymbol{a} \big\Vert}_{F}^2 
        \!\!+ \lambda_y {\big\Vert \mathbf{Y}\boldsymbol{1}_M\!-\! \mathbf{D}^y\boldsymbol{a} \big\Vert}_{F}^2 \\ + & \tau_x {\big\Vert\mathbf{T}\boldsymbol{1}_n \!-\! \boldsymbol{a} \big\Vert}_{2}^2  \!+\! \tau_y {\big\Vert\mathbf{T}^\top\boldsymbol{1}_m \!-\! \boldsymbol{b} \big\Vert}_{2}^2 	
        \end{split}	
    \end{aligned}
    \label{eq10}
\end{equation}
Where $\gamma, \lambda_{x(y)}, \tau_{x(y)}$ and $\kappa_{x(y)}$ are positive Lagrangian multipliers. Accordingly, the solution of Eq. \ref{eq10} convergent to that of Eq. \ref{eq9} when the Lagrangian multipliers go to infinity. The main idea of the alternating method is to solve the problem sequentially by fixing one variable from another. It is easy to see that Eq. \ref{eq9} can be divided into three sub-problems: sparse coding, style dictionaries learning and transport map inferring, respectively. For brevity, we describe the solution for the variables $\mathbf{T}, \mathbf{D}^x, \mathbf{A}$, and $\boldsymbol{a}$, and $\mathbf{D}^y, \mathbf{B}$, and $\boldsymbol{b}$ can be processed analogically. 

\textbf{Sparse coding:}  By fixing $\mathbf{T}, \mathbf{D}^x, \mathbf{D}^y$ and $\boldsymbol{a}$,  $\boldsymbol{b}$ in Eq. \ref{eq10}, the optimization problem w.r.t. the variables  $\mathbf{A}$ and $\mathbf{B}$ is then reduced into a standard sparse encoding problem. Considering $\mathbf{X}\boldsymbol{1}_M= \mathbf{D}^x \boldsymbol{a}$,$ \mathbf{Y}\boldsymbol{1}_N = \mathbf{D}^y \boldsymbol{b}$, the variables $\mathbf{A}$ and $\mathbf{B}$ only have sparse constraints. Taking the coefficient $\mathbf{A}$ for example, the representation vectors $\boldsymbol{\alpha}_i$ for each example $\boldsymbol{x}_i$ in $\mathbf{X}$ is the $i$-th column of $\mathbf{A}$, which is attained by solving the following problem,
\begin{equation}
    \begin{aligned}
    \min_{\boldsymbol{\alpha}_i} {\lvert\lvert \boldsymbol{\alpha}_i \rvert\rvert}_0, \quad  s.t. \quad \boldsymbol{x}_i = \mathbf{D}^x\boldsymbol{\alpha}_i. 
    \end{aligned}
    \label{eq11}
\end{equation}
As aforementioned, the sparse coding problem of Eq. \ref{eq11} can be solved by many existing methods such as  matching pursuit (MP) or orthogonal matching pursuit (OMP) algorithms~\cite{aharon2006k, chen1989orthogonal}. We here use the OMP method for ease of implementation. Once the coefficients $\mathbf{A}$ and $\mathbf{B}$ are obtained, we have the row sums of the coefficients --- that is, $\boldsymbol{a}=\mathbf{A}\boldsymbol{1}_M$, $ \boldsymbol{b}=\mathbf{B}\boldsymbol{1}_N$ and they provide a discrete probability measure for  dictionary atoms. It is worth mentioning here that the probability distributions $\boldsymbol{a}$ and $\boldsymbol{b}$ must be positive, while the learned coefficients may be negative here. 
It is possible to remedy $\boldsymbol{d}^x_i=-\boldsymbol{d}^x_i$ and $\mathbf{A}(i,:) =-\mathbf{A}(i,:)$ without affecting the sparse encoding process when the $i$-th row sum $\boldsymbol{a}_i$ is negative. The non-negative sparse coding is also an alternative way for remedy in practice.  Without the ambiguity $\boldsymbol{a}$ and $\boldsymbol{b}$ are also denoted as the normalized counterparts. 

\textbf{Learning style dictionaries:}  By analogy, we then fix the coefficients  $\mathbf{A}, \mathbf{B}$ and transport map $\mathbf{T}$ and update the style dictionaries $\mathbf{D}^x$ and $\mathbf{D}^y$, respectively. Considering the relaxed constraints $\mathbf{X}\boldsymbol{1}_M= \mathbf{D}^x \boldsymbol{a}, \mathbf{Y}\boldsymbol{1}_N = \mathbf{D}^y \boldsymbol{b}$, we rewrite the Tikhonov regularization form of the sub-problem of Eq. \ref{eq10} with respect to $\mathbf{D}^x$ (or, $\mathbf{D}^y$) as, 
\begin{equation}
    \begin{aligned}
    \begin{split}
    \operatorname*{argmin}_{\{\boldsymbol{d}^x_i\}} &{\Vert \mathbf{D}^x\mathbf{A} \!-\! \mathbf{X} \Vert}_{F}^2 + \lambda_x {\Vert \mathbf{D}^x \boldsymbol{a} - \mathbf{X}\boldsymbol{1}_M \Vert}_{2}^2 \\+& \gamma \sum_{i,j} \mathbf{T}_{i,j} {\Vert\boldsymbol{d}^x_i\!-\!\boldsymbol{d}^y_j \Vert}^2_2  
    \end{split}	
    \end{aligned}
\label{eq12}
\end{equation}
where $\lambda_x$ and $\tau_x$ are the positive weights. Accordingly, Eq. \ref{eq10} can be viewed as a regularized dictionary learning problem. To learn the redundant style dictionaries, we use the famous K-SVD algorithm~\cite{aharon2006k} to update each element $\boldsymbol{d}^x_i$ of dictionary $\mathbf{D}^x$ sequentially. Notice that the original K-SVD algorithm is not directly applicable due to the regularization terms in Eq. \ref{eq12}. We instead introduce an extended K-SVD algorithm for updating dictionaries. For clarity, we first review the original K-SVD algorithm~\cite{aharon2006k} and show how to extend it to the proposed model.

\begin{figure*}[!t]
	\centering
	\begin{subfigure}{0.135\linewidth}
		\includegraphics[width=\textwidth, height = 0.08\textheight]{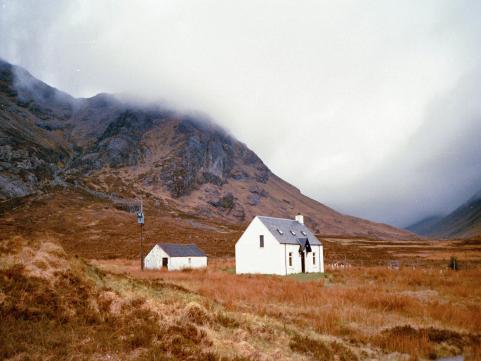}
		\includegraphics[width=\textwidth, height = 0.09\textheight]{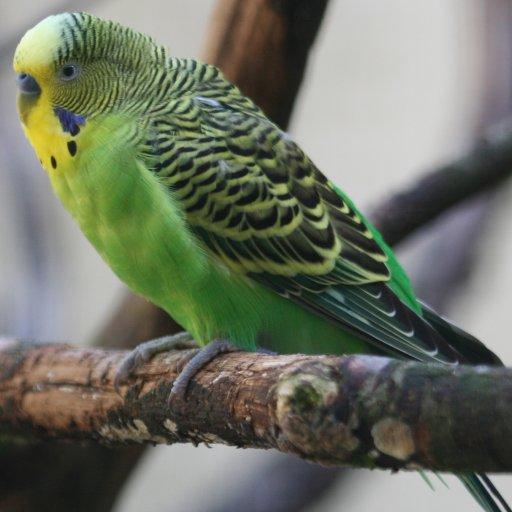}
		\includegraphics[width=\textwidth, height = 0.13\textheight]{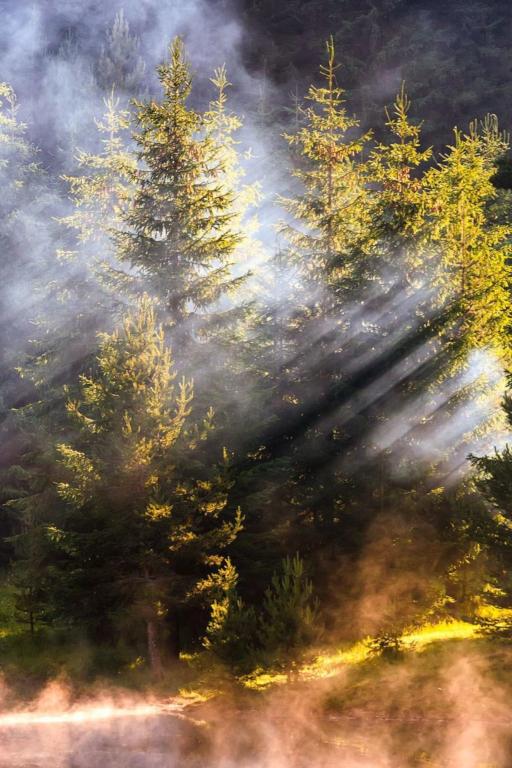}
		\includegraphics[width=\textwidth, height = 0.08\textheight]{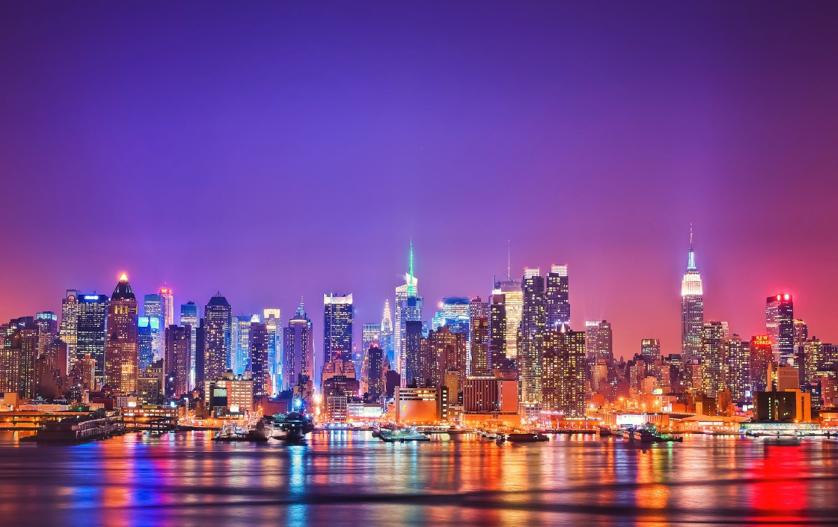}
		\caption{Input}
		\label{fig4:short-a}
	\end{subfigure}
	\begin{subfigure}{0.135\linewidth}
		\includegraphics[width=\textwidth, height = 0.08\textheight]{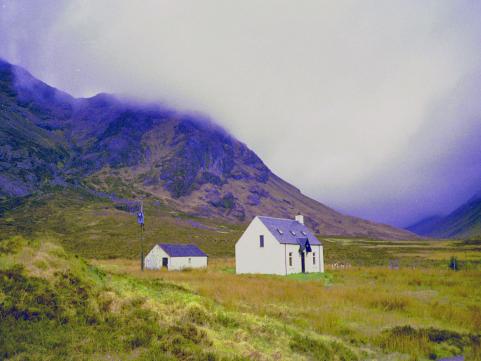}		
		\includegraphics[width=\textwidth, height = 0.09\textheight]{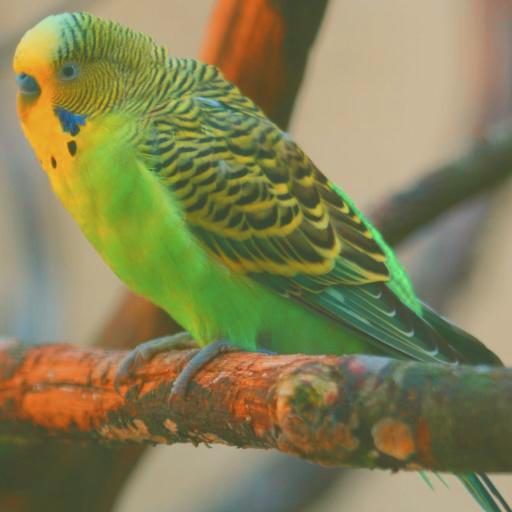}
		\includegraphics[width=\textwidth, height = 0.13\textheight]{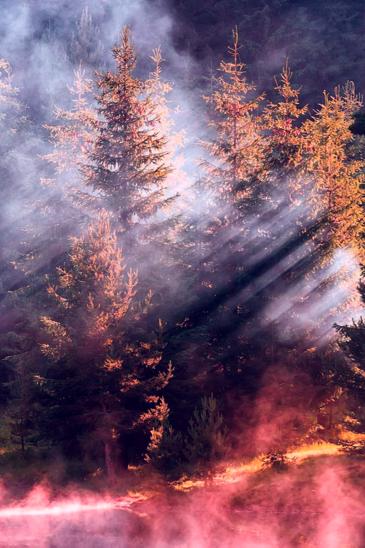}		
		\includegraphics[width=\textwidth, height = 0.08\textheight]{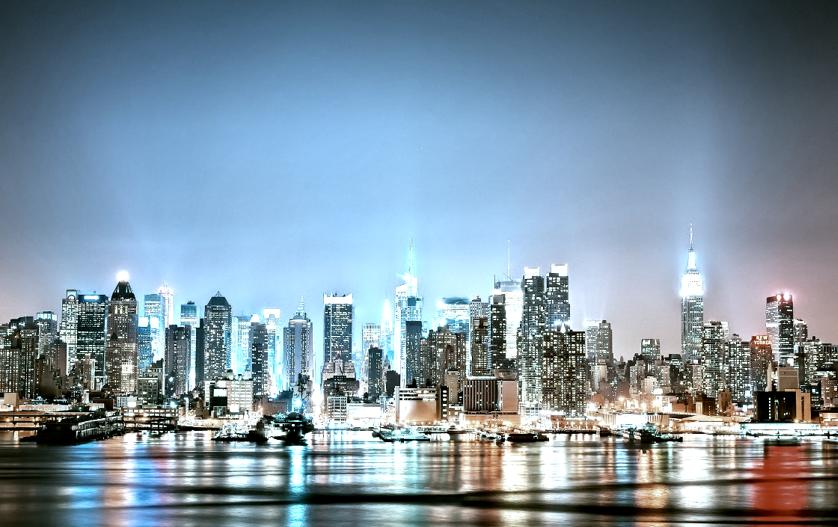}
		\caption{MKL}
		\label{fig4:short-b}
	\end{subfigure}
	\begin{subfigure}{0.135\linewidth}
		\includegraphics[width=\textwidth, height = 0.08\textheight]{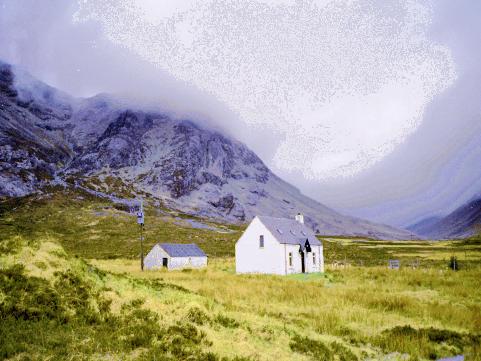}
		\includegraphics[width=\textwidth, height = 0.09\textheight]{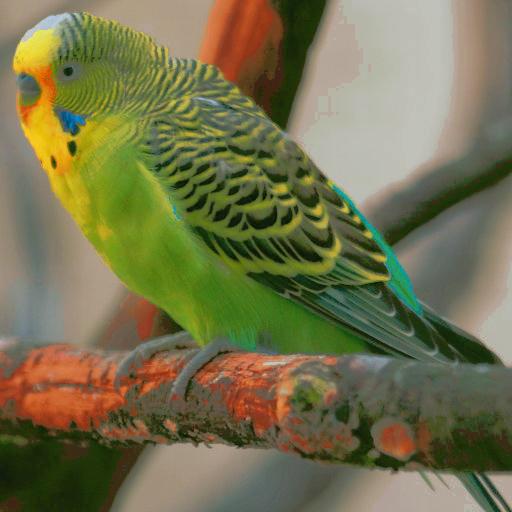}
		\includegraphics[width=\textwidth, height = 0.13\textheight]{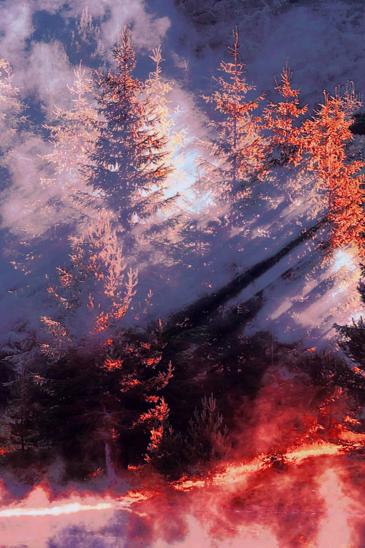}
		\includegraphics[width=\textwidth, height = 0.08\textheight]{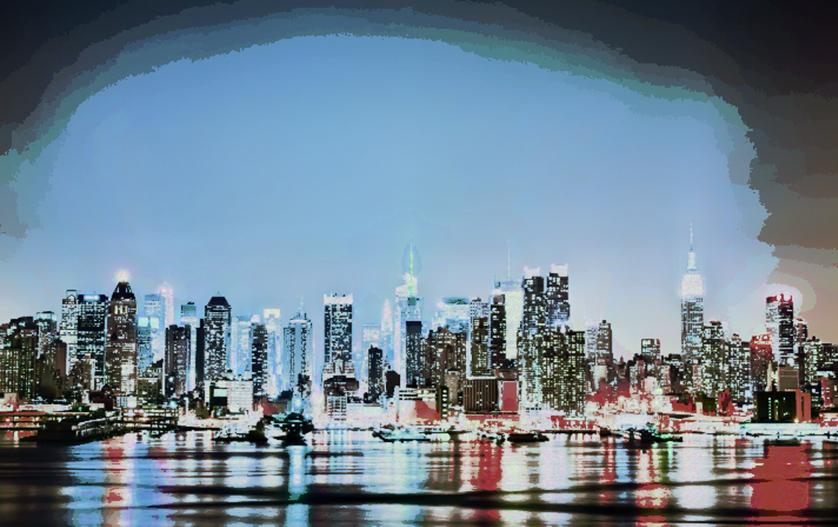}
		\caption{ROT}
		\label{fig4:short-c}
	\end{subfigure}
	\begin{subfigure}{0.135\linewidth}
		\includegraphics[width=\textwidth, height = 0.08\textheight]{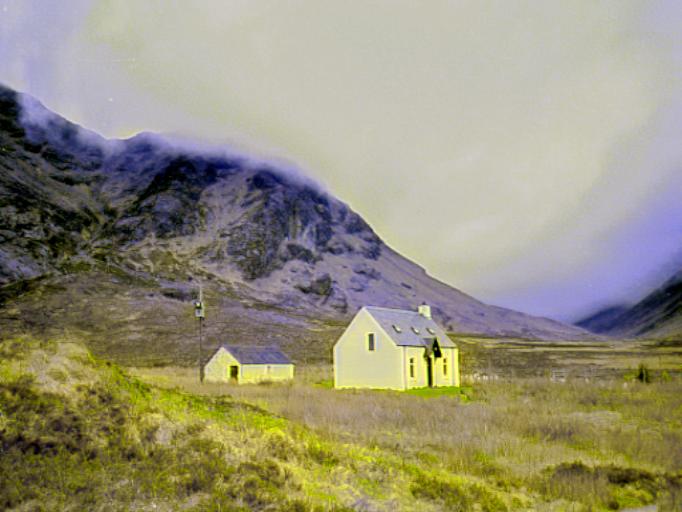}
		\includegraphics[width=\textwidth, height = 0.09\textheight]{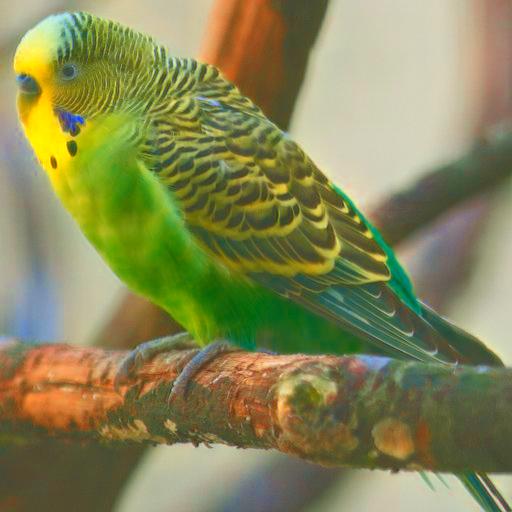}
		\includegraphics[width=\textwidth, height = 0.13\textheight]{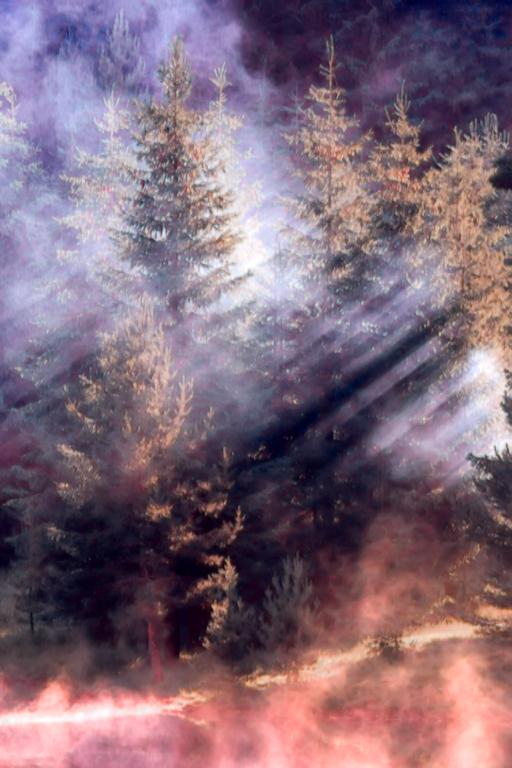}
		\includegraphics[width=\textwidth, height = 0.08\textheight]{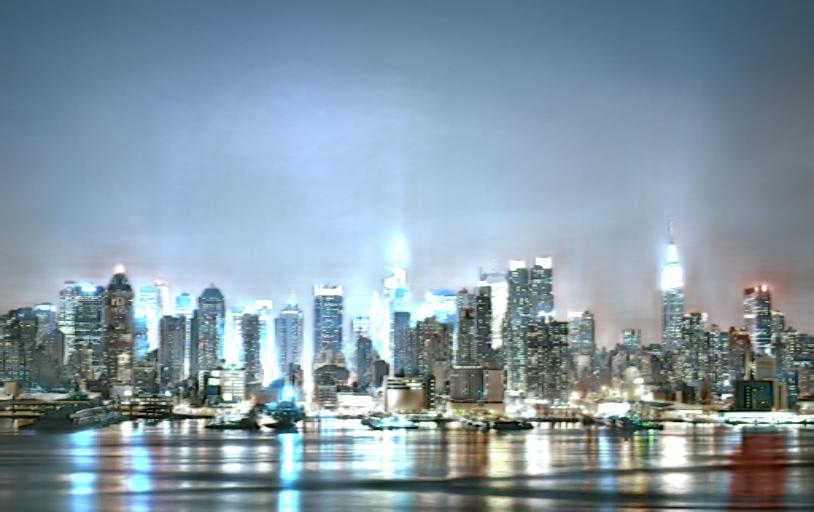}
		\caption{PhotoNAS}
		\label{fig4:short-d}
	\end{subfigure}
	\begin{subfigure}{0.135\linewidth}
		\includegraphics[width=\textwidth, height = 0.08\textheight]{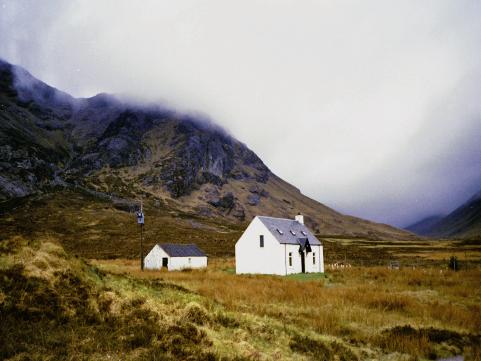}
		\includegraphics[width=\textwidth, height = 0.09\textheight]{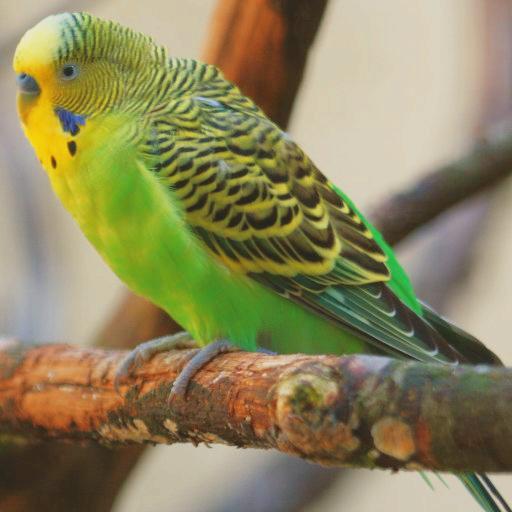}
		\includegraphics[width=\textwidth, height = 0.13\textheight]{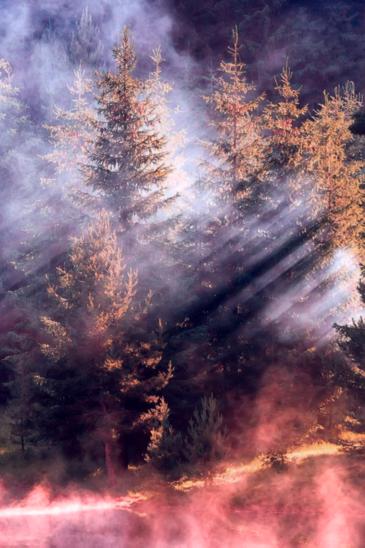}
		\includegraphics[width=\textwidth, height = 0.08\textheight]{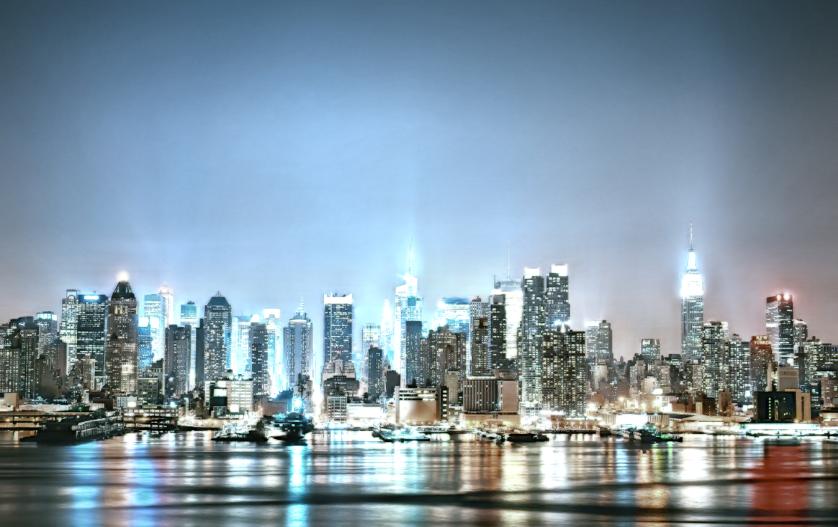}
		\caption{NeuralPreset}
		\label{fig4:short-e}
	\end{subfigure}
	\begin{subfigure}{0.135\linewidth}
		\includegraphics[width=\textwidth, height = 0.08\textheight]{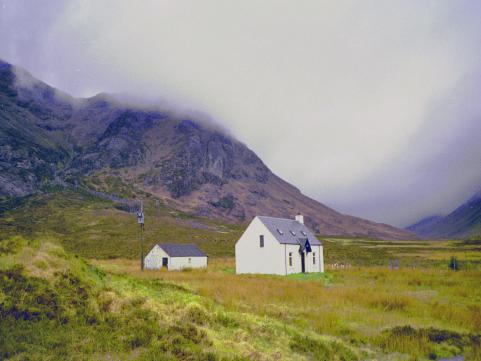}
		\includegraphics[width=\textwidth, height = 0.09\textheight]{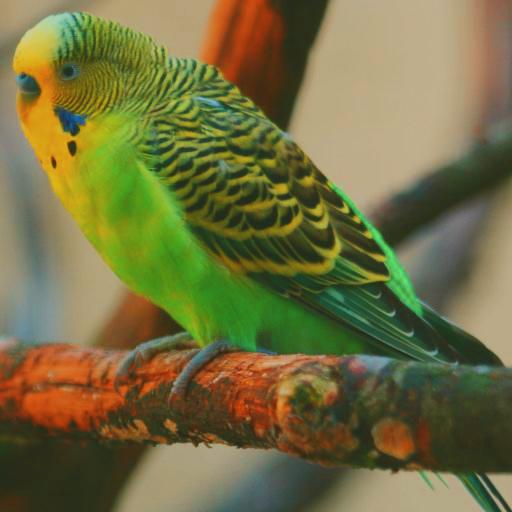}
		\includegraphics[width=\textwidth, height = 0.13\textheight]{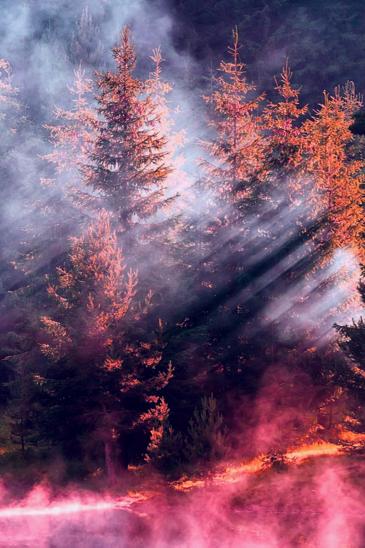}
		\includegraphics[width=\textwidth, height = 0.08\textheight]{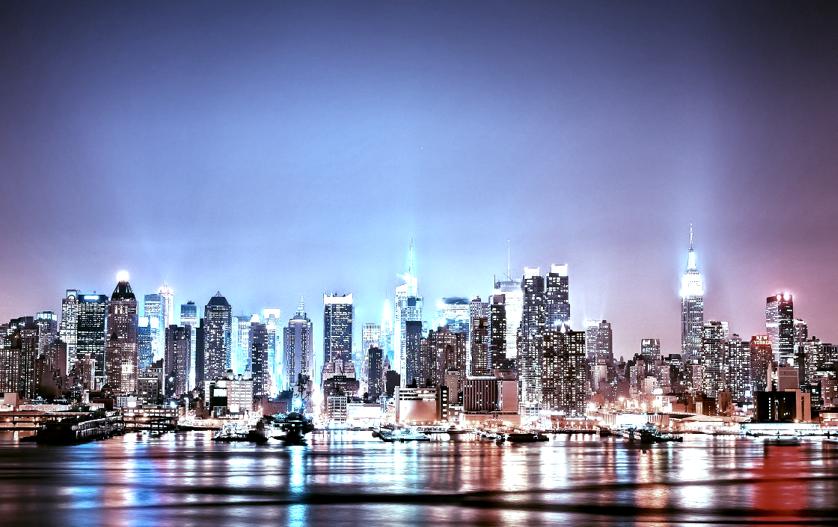}
		\caption{Ours}
		\label{fig4:short-f}
	\end{subfigure}
	\begin{subfigure}{0.135\linewidth}
		\includegraphics[width=\textwidth, height = 0.08\textheight]{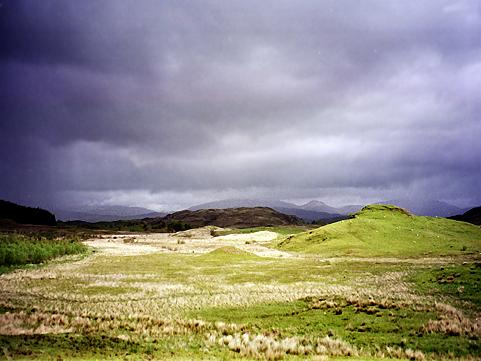}
		\includegraphics[width=\textwidth, height = 0.09\textheight]{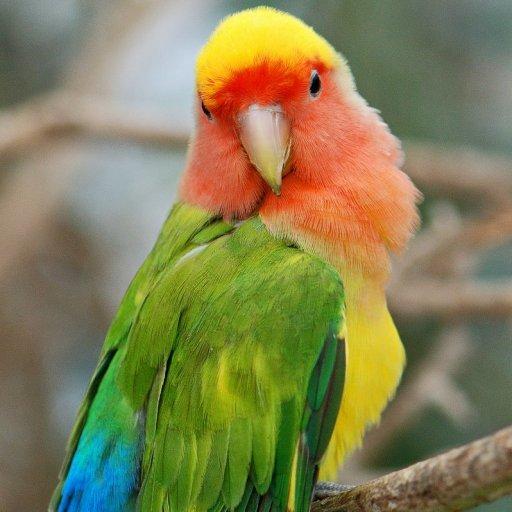}
		\includegraphics[width=\textwidth, height = 0.13\textheight]{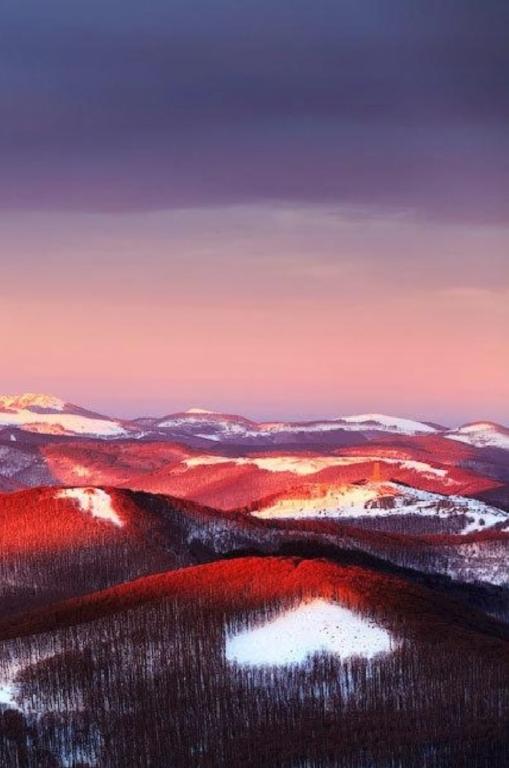}
		\includegraphics[width=\textwidth, height = 0.08\textheight]{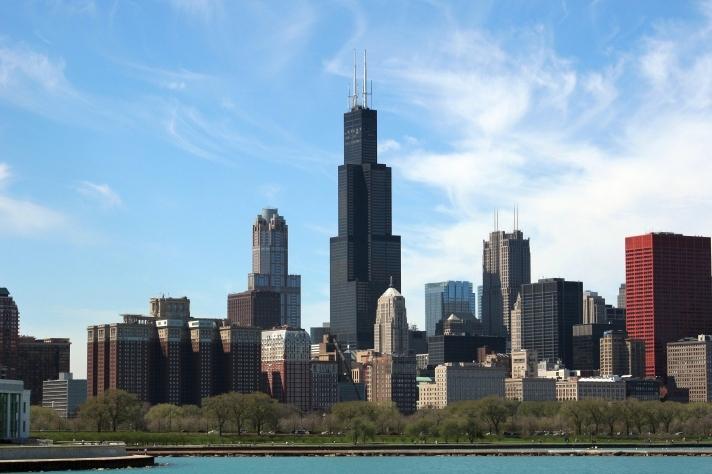}
		\caption{Reference}
		\label{fig4:short-g}
	\end{subfigure}
	\hfill
	\caption{Visual comparison of color transfer on natural images.  From left to right: (a) content images, (b) MKL~\cite{pitie2007linear}, (c) Regularized OT~\cite{ferradans2014regularized}, (d) PhotoNAS~\cite{an2020ultrafast}, (e) NeuralPreset~\cite{ke2023neural}, (f) our results, and (g) reference images.  (Zoom in for better view).}
	\label{fig4:short}
\end{figure*}

\textbf{The Extended K-SVD Algorithm:} In the K-SVD algorithm\cite{aharon2006k}, sparse representation is to factorized an image $\mathbf{X}$ into a multiple form of the dictionary $\mathbf{D}^x$ and coefficient $\mathbf{A}$, that is, $\mathbf{X}\!=\!\mathbf{D}^x\mathbf{A} $. We now focus on the style dictionaries-learning process by fixing the sparse coefficient $\mathbf{A}$. The dictionary $\mathbf{D}^x$ can be updated by minimizing the objective function,
\begin{equation}
\small
\begin{aligned}
\begin{split}
   {\Vert \mathbf{X}\!-\!\mathbf{D}^x\mathbf{A} \Vert}_{F}^2  \!=\! &  \left\| \mathbf{X}\!-\! \sum_{j=1}^n \boldsymbol{d}^x_j \boldsymbol{\alpha}_{T}^j \right\|_{F}^2 
  \!=\!  \left\| \left( \mathbf{X}\!-\! \sum_{j \neq k} \boldsymbol{d}^x_j \boldsymbol{\alpha}_{T}^j \right)  \!-\! \boldsymbol{d}^x_k \boldsymbol{\alpha}_{T}^k  \right\|_{F}^2 \\
  =  & \left\| \mathbf{E}^k\!-\! \boldsymbol{d}^x_k \boldsymbol{\alpha}_{T}^k  \right\|_{F}^2
\end{split}	
\end{aligned}
\label{eq13}
\end{equation}
where $\mathbf{E}^k=\mathbf{X}\!-\! \sum_{j \neq k} \boldsymbol{d}^x_j \boldsymbol{\alpha}_{T}^j$ is the residual part that not involves the $k$-th dictionary element $\boldsymbol{d}^x_k$, and $\alpha_{T}^k$ is the $k$-th row of the coefficient matrix $\mathbf{A}$.  With the above form, $\boldsymbol{d}_i^x$ is updated as the first left eigenvector given by SVD algorithm based on the K-SVD algorithm~\cite{aharon2006k}. The basic idea here is to decompose the term $\mathbf{D}^x \mathbf{A}$ into the sum of $n$ rank-1 matrices, where only one dictionary element $\boldsymbol{d}^x_k$ is involved and can be updated independently in each time when fixing the remainder $\mathbf{E}^k$. This strategy helps to learn the redundant dictionaries more efficiently. 

The process can be analogically extended to the regularization case. Let $\mathbf{E}^k \!=\!\mathbf{X}\!-\!\sum_{k\neq i} \boldsymbol{d}_k^x \boldsymbol{\alpha}_{k}^T$ and $\mathbf{F}^k\!=\! \mathbf{X}\mathbf{1}_{M}\!-\!\sum_{k\neq i} \boldsymbol{d}_k^x \boldsymbol{a}_{k}^T$ be the residual of ${\Vert \mathbf{D}^x\mathbf{A} \!-\! \mathbf{X} \Vert}_{F}^2$ and ${\Vert \mathbf{D}^x \boldsymbol{a} - \mathbf{X}\boldsymbol{1}_M \Vert}_{2}^2$ without using the element $\boldsymbol{d}_i^x$, where $\boldsymbol{\alpha}_{k}$ and $\boldsymbol{a}_{k}$ are the $k$-th row vector of $\mathbf{A}$ and  $\boldsymbol{a}$. With this notation, the first two terms in Eq. \ref{eq12} can be rewritten as ${\Vert \mathbf{E}^k \!-\!\boldsymbol{d}_i^x \boldsymbol{\alpha}_{i}^T \Vert}_{F}^2$ and $ {\Vert \mathbf{F}^k \!-\! \boldsymbol{d}_i^x \boldsymbol{a}_{i}^T \Vert}_{F}^2$, respectively. The sub-problem Eq. \ref{eq12} with respect to the $k$-th dictionary $\boldsymbol{d}^x_k$ can be rewritten into the form,
\begin{equation}
\begin{aligned}
\begin{split}
\operatorname*{argmin}_{\boldsymbol{d}^x_k} & \left\| \mathbf{E}^k\!-\! \boldsymbol{d}^x_k \boldsymbol{\alpha}_{T}^k  \right\|_{F}^2 + \lambda_x \left\| \mathbf{F}^k\!-\! \boldsymbol{d}^x_k \boldsymbol{a}_{T}^k  \right\|_{F}^2\\
+ & \gamma \sum_{j} \mathbf{T}_{k,j} {\Vert\boldsymbol{d}^x_k\!-\!\boldsymbol{d}^y_j \Vert}^2_2.  
\end{split}	
\end{aligned}
\label{eq14}
\end{equation}
It is easy to verify that $\boldsymbol{d}_i^x$ has a closed-form solution because the objective function in Eq. \ref{eq14} is quadratic with respect to $\boldsymbol{d}^x_k$. The last two terms in Eq. \ref{eq14} can be treated as the regularization terms compared with Eq. \ref{eq13} and such a regularization also helps to strengthen a more stable numerical solution. The reader is referred to the K-SVD algorithm\cite{aharon2006k} for more details.

\textbf{Optimal transport:} The transport mapping $\mathbf{T}$ over the learned style dictionaries is a standard optimal transport problem when fixing the style dictionaries $\mathbf{D}^x, \mathbf{D}^y$ and coefficients $ \mathbf{A}, \mathbf{B}$, that is,  
\begin{equation}
\begin{aligned}
    \begin{split}
        &\operatorname*{argmin}_{\mathbf{T}} \ 
        \sum_{i,j} \mathbf{T}_{i,j} {\Vert\boldsymbol{d}^x_i\!-\!\boldsymbol{d}^y_j \Vert}^2_2,  \\ & s.t.  \quad \mathbf{T}\boldsymbol{1}_n = \boldsymbol{a}, \  \mathbf{T}^\top\boldsymbol{1}_m = \boldsymbol{b}.	
    \end{split}	
\end{aligned}
\label{eq15}
\end{equation}	

It is worth noting that a discrete optimal transport can be solved by linear programming, while the computational cost increases significantly over the large-scale samplings~\cite{peyre2019computational}. It is empirically solvable in our case due to the small size of style dictionaries, which forms one of the cores of our method. Notice that $\boldsymbol{a}$ and $\boldsymbol{b}$ in Eq. \ref{eq12} represent the normalized counterparts.

\begin{algorithm}[!t]
	\caption{Optimal Transport using Sinkhorn algorithm.}
	\label{alg1}
	\begin{algorithmic}[0]
		\STATE \textbf{Input:} Cost function $\mathbf{C}$, discrete distributions $\boldsymbol{a}$ and $\boldsymbol{b}$, parameter $\eta$, and maximum iterations $K$;
		\STATE \textbf{Initialization:} Let $\mathbf{M}=e^{-\mathbf{C} / \eta}, \boldsymbol{v} \leftarrow \boldsymbol{1}, k \leftarrow 0$
		\WHILE{$k\leq K$}
		\STATE $\quad  \boldsymbol{u}^{k+1}= \boldsymbol{a} \oslash (\mathbf{M} \boldsymbol{v}^{k})$ 
		\STATE $\quad  \boldsymbol{v}^{k+1}= \boldsymbol{b} \oslash \left(\mathbf{M}^{\top}  \boldsymbol{u}^{k+1} \right)$ 
		\ENDWHILE
		\STATE \textbf{Output:} $\mathbf{T} =\operatorname{diag}(\boldsymbol{u}^{k+1}) \mathbf{M} \operatorname{diag} (\boldsymbol{v}^{k+1}).$
	\end{algorithmic}
\end{algorithm}

\textbf{Entropy-Regularized Optimal Transport:} It is well-known that an exact solution to optimal transport based on the network flow method has computational complexity $O\left(n^3\right)$  for the $n$ samplings\cite{peyre2019computational}. The solution may be unavailable when $n$ exceeds thousands of samplings in a general PC platform. Instead, we resort to a more efficient entropy-regularized optimal transport,
\begin{equation}
\begin{aligned}
\begin{split}
    \operatorname*{argmin}_{\mathbf{T}}& \ \sum_{i,j}  \mathbf{C}_{i,j}  \mathbf{T}_{i,j} + \eta H(\mathbf{T}) \\ 
     s.t. \quad &\mathbf{T}\boldsymbol{1}_m = \boldsymbol{a}, \quad  \mathbf{T}^\top\boldsymbol{1}_n = \boldsymbol{b}.
\end{split}	
\end{aligned}
\label{eq16}
\end{equation}
where $H(\mathbf{T}) =  \sum_{i,j}\mathbf{T}_{i,j} (log (\mathbf{T}_{i,j})-1)$ is the negative entropic regularization, and $\eta$ is the positive regularization parameter. As interpreted in~\cite{cuturi2013sinkhorn}, the regularized model of Eq. \ref{eq16} is a convex optimization problem and can be solved with the Sinkhorn-Knopp algorithm. A detailed solution is also presented in Alg. \ref{alg1}. Note that the sub-problems in Alg. \ref{alg1} involve component-wise divide operators $\oslash$ that can be computed efficiently.

\subsection{Image Synthesis}

Once the dictionaries $\mathbf{D}^{x}$ and $\mathbf{D}^{y}$ and transport map $\mathbf{T}$ are obtained, given an image $\mathbf{x}$ in style $s_{x}$, it is then easy to reconstruct an image $\hat{\mathbf{y}}$ with the style $s_{y}$ by swapping the corresponding style dictionaries but keeping the sparse representing coefficients invariant, that is,
\begin{equation}
	\small
\begin{split}
\hat{\mathbf{y}}_i= \hat{\mathbf{D}}^{x} \boldsymbol{\alpha}_i
=\mathbf{T}({\mathbf{D}}^{y}) \boldsymbol{\alpha}_i,
\end{split}	
\label{eq17}
\end{equation}
where the $k$-th column of $\hat{\mathbf{D}}^{x}$ is $\hat{\boldsymbol{d}}^{x}_i\!=\!\mathbf{T}(\boldsymbol{d}^{y}_j)\!=\!\frac{\sum_{j=1}^n \mathbf{T}_{i,j} \boldsymbol{d}^y_j}{\sum_{j=1}^n \mathbf{T}_{i,j}}$, which can be viewed as a posterior mean estimate to define a one-to-one of the transfer function~\cite{rabin2014adaptive}, and $\boldsymbol{\alpha}_{i}$ is the sparse coefficients of patch $\mathbf{x}_{i}$. In most cases, the image $\mathbf{x}$ is not exactly the same as training data --- but is sampled from an identical distribution, the coefficients $\boldsymbol{\alpha}_i$ for each patch $\mathbf{x}_i$ can be learned based on the sparse coding Eq. \ref{eq14}. In the cases of photo-realistic image transfer, it may be preferable to add some constraints for more consistent local textures, for example, using a simple gradient regularization for image synthesis,
\begin{equation}
	\small
	\begin{split}
		\operatorname*{argmin}_{\hat{\mathbf{y}}} \    & {\Vert \hat{\mathbf{y}} -\hat{\mathbf{D}}^{x} \boldsymbol{\alpha} \Vert}_{2}^2 + \rho {\Vert \nabla \hat{\mathbf{y}} - \nabla \mathbf{x} \Vert}_{2}^2\\
	\end{split}	
	\label{eq18}
\end{equation}
where $\nabla$ is the gradient operator and $\rho$ is the parameter to weight the gradient regularization term. It is easy to verify that Eq. \ref{eq18} can be easily computed due to its closed-form solution.

\section{Experiment Results}

In this section, we extensively illustrate the performance of optimal style transfer and show the empirical evidence on two fundamental image-to-image translation tasks: color transform and artistic style transfer. In each scenario, 
the sparse coefficients and individual dictionaries are firstly learned using sparse representation and then an optimal transport map is derived on the learned dictionaries (See Fig. \ref{fig1}). The process is updated iteratively until it converges to a given stop criteria. The learned dictionaries are treated as individual feature styles for image reconstruction.

\subsection{Configurations}

For simplicity, we only show the training and reconstruction process on a pair of content and reference images, while it is easy to immigrate the procedure to the case of large-scale datasets. Let ${\{\boldsymbol{x}_i\}}_{i=1}^{M}$, ${\{\boldsymbol{y}_j\}}_{j=1}^{N}$ be two patches data and $\mathbf{D}^x \!=\! [\boldsymbol{d}_1^x, \cdots, \boldsymbol{d}_m^x]$,  $\mathbf{D}^y \!=\! [\boldsymbol{d}_1^y, \cdots, \boldsymbol{d}_n^y]$ be dictionaries as defined before, we randomly select $M(N)\!=\! 10K \!\sim\! 100K$ patches depending on the size of images. The dictionary size $m(n)\!=\! 256$ in most cases for computational efficiency. In general, the larger size of dictionaries helps to produce better performance, which however is computationally expensive, especially for the large-scale optimal transport between dictionaries. The patch size is $16 \!\times\! 16$ pixels ($d=256$) and it is sequentially concatenated by channels for color images. As illustrated, we use an extended K-SVD algorithm to update the dictionaries. 

\begin{figure}[!t]
	\begin{center}
		\includegraphics[width=\linewidth]{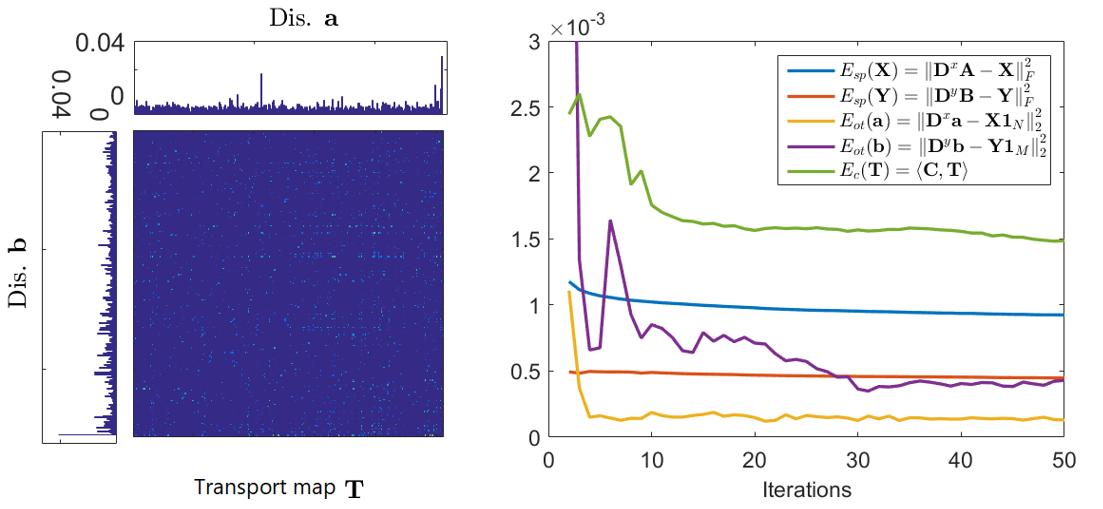}
	\end{center}
	\caption{The normalized distributions $\boldsymbol{a}$, $\boldsymbol{b}$, and the corresponding optimal map $\mathbf{T}$ (left), and the loss curves of sparse representation, OT constraints and transport plan with iterations (right). }
	\label{fig4}
\end{figure}

The parameters are configured as follows. In the sparse coding step, we update the sparse coefficients based on orthogonal matching pursuit (OMP) algorithm~\cite{chen1989orthogonal}, and specify the representation error ($\kappa = 10^{-5}$) as the stop criteria for each patch data $\boldsymbol{x}_i$ ($\boldsymbol{y}_j$). It is necessary to check the row sum of coefficient matrices $\mathbf{A}$ and $\mathbf{B}$ to be positive in each step. In the dictionary update step, each element $\boldsymbol{d}_i^x$ ($\boldsymbol{d}_j^y$) is sequentially updated by solving \ref{eq10}, where $\lambda_x(\lambda_y)\!=\!1.0$ and $\tau_x(\tau_y)\!=\!10.0$. Similar to the K-SVD algorithm~\cite{aharon2006k}, we replace the correlated dictionary atoms by the randomly-selected data samples, which helps to learn the individual styles more faithfully. In the optimal transport step, a linear program solver~\cite{peyre2019computational} is employed for the small-size cases. We set $\rho \!=\! 0.01$ in \ref{eq13} if necessary. In large-size dictionaries, for example, $m (n)\!\ge\! 512 $, one can resort to the entropy-regularization for efficiency~\cite{cuturi2013sinkhorn}. The process is updated iteratively until it converges to the stop criteria.  

\begin{figure*}[!t]
	\centering
	\begin{subfigure}{0.16\linewidth}
		\includegraphics[width=\textwidth, height=0.15\textheight]{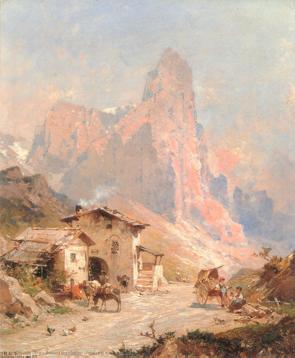}
		\includegraphics[width=\textwidth, height=0.15\textheight]{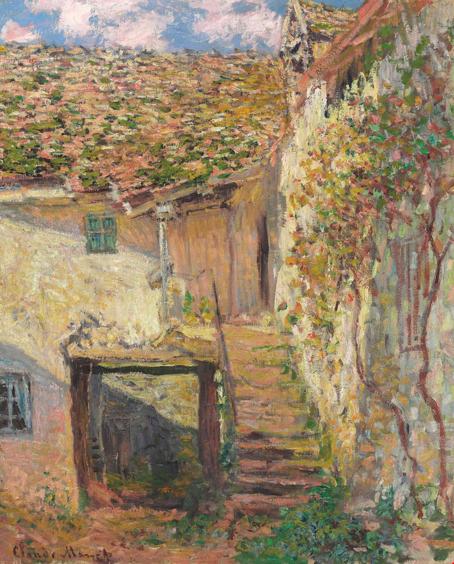}
		\caption{Content}
		\label{fig6-a}
	\end{subfigure}
	\begin{subfigure}{0.16\linewidth}
		\includegraphics[width=\textwidth, height=0.15\textheight]{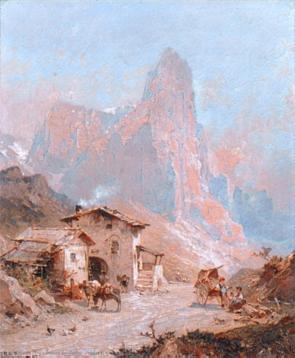}
		\includegraphics[width=\textwidth, height=0.15\textheight]{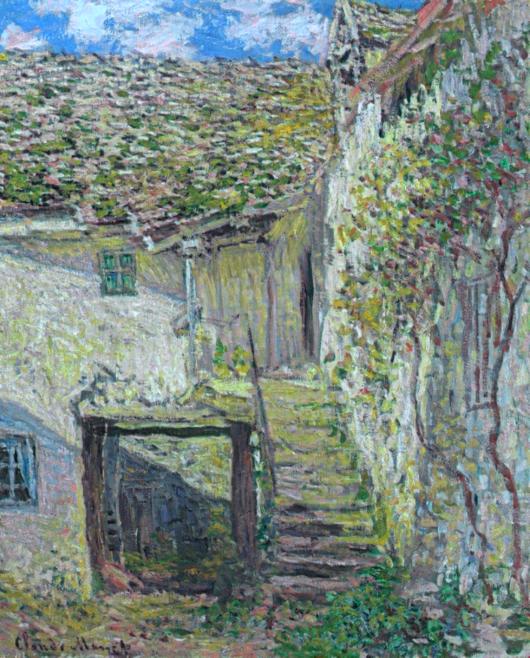}
		\caption{Ours}
		\label{fig6-b}
	\end{subfigure}
	\begin{subfigure}{0.16\linewidth}
		\includegraphics[width=\textwidth, height=0.15\textheight]{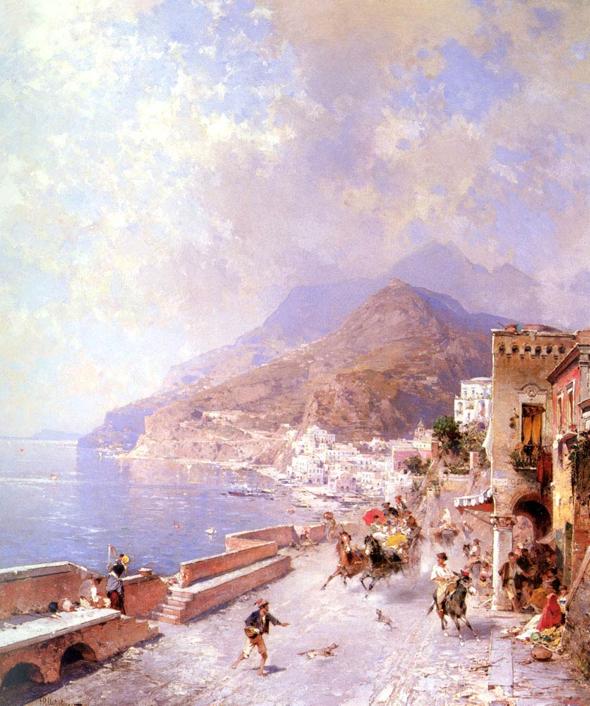}	
		\includegraphics[width=\textwidth, height=0.15\textheight]{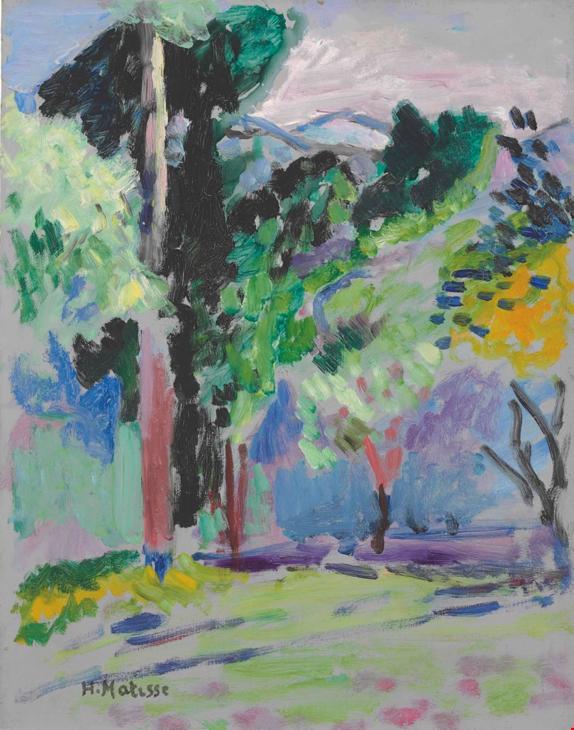}
		\caption{Reference}
		\label{fig6-c}
	\end{subfigure}
	\begin{subfigure}{0.155\linewidth}
		\includegraphics[width=\textwidth, height=0.15\textheight]{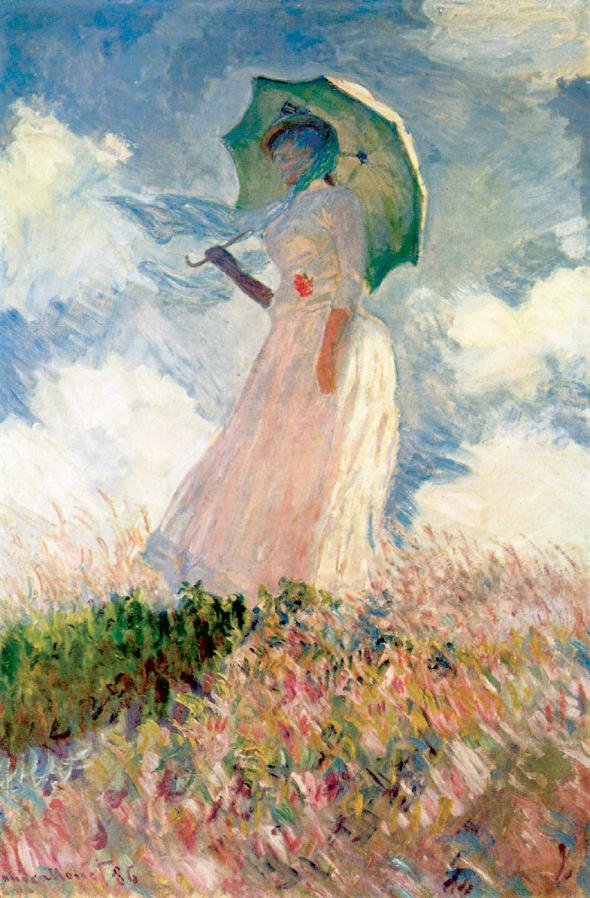}
		\includegraphics[width=\textwidth, height=0.15\textheight]{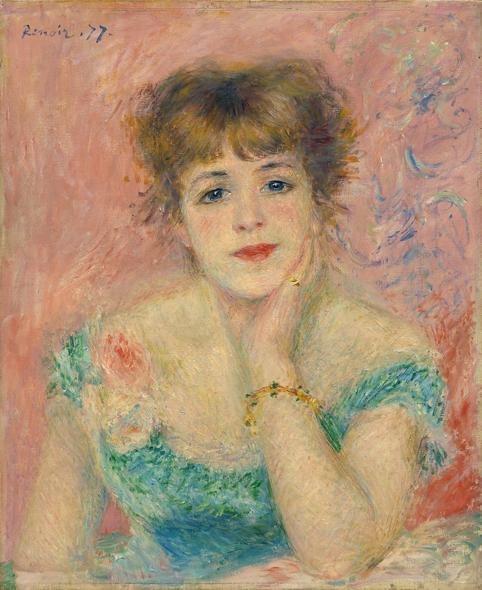}
		\caption{Content}
		\label{fig6-d}
	\end{subfigure}
	\begin{subfigure}{0.16\linewidth}
		\includegraphics[width=\textwidth, height=0.15\textheight]{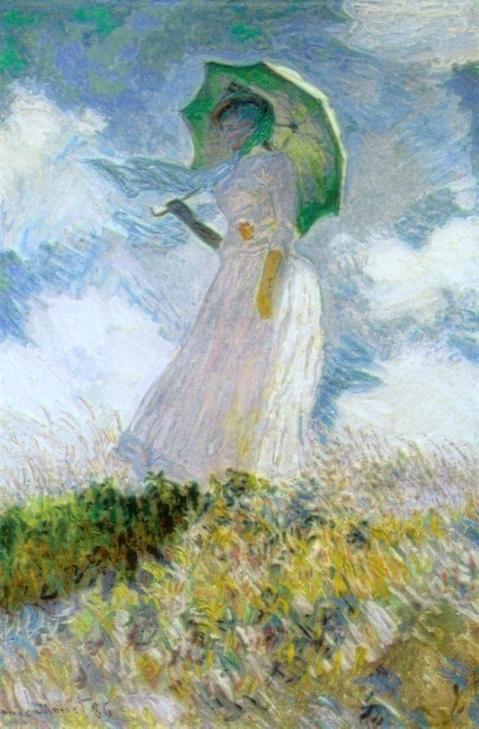}
		\includegraphics[width=\textwidth, height=0.15\textheight]{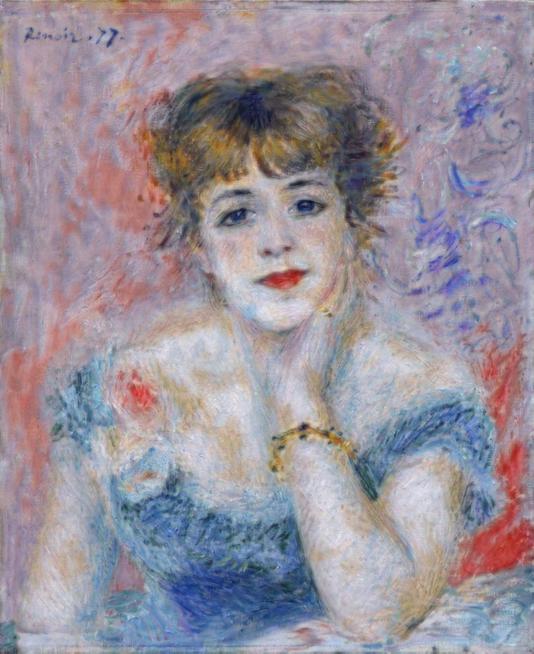}
		\caption{Ours}
		\label{fig6-e}
	\end{subfigure}
	\begin{subfigure}{0.16\linewidth}
		\includegraphics[width=\textwidth, height=0.15\textheight]{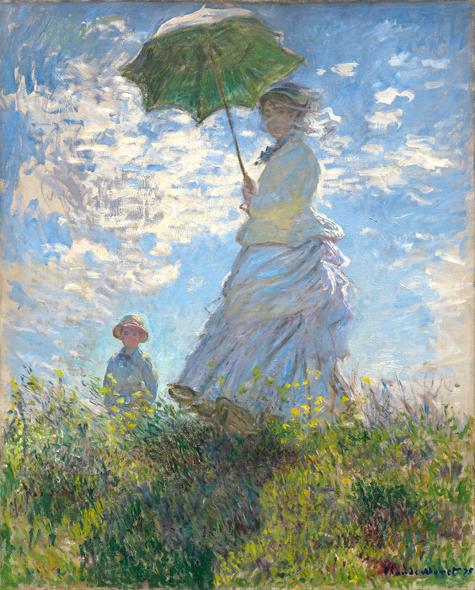}
		\includegraphics[width=\textwidth, height=0.15\textheight]{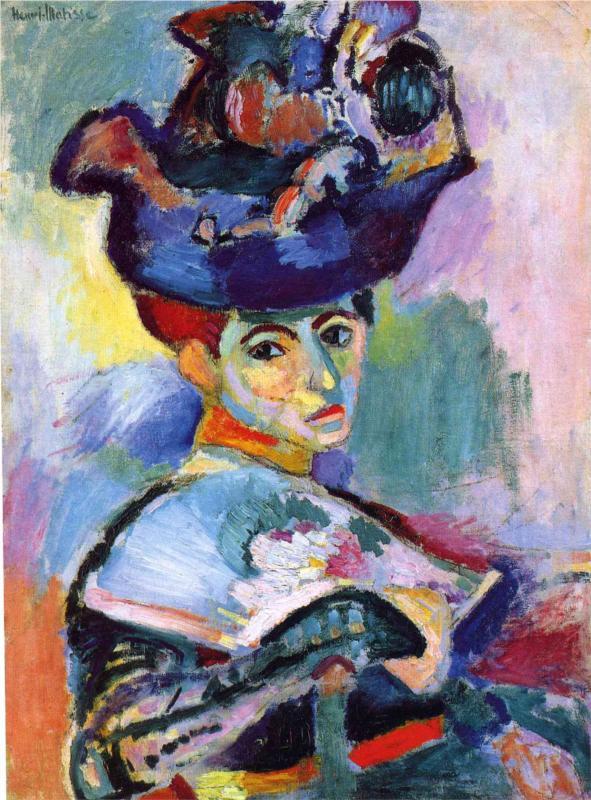}	
		\caption{Reference}
		\label{fig6-f}
	\end{subfigure}
	\caption{Artistic style transfer effects. Given input images (a) and (d),  and reference images (c) and (f), the proposed model gives rise to the stylized results (b) and (e) with consistent textures and structures (Zoom in for better view).}
	\label{fig6}
\end{figure*}

We interpret the training process by taking the example in \ref{fig1} into account. Let $E_{sp}(\mathbf{X}) \!=\! {\Vert \mathbf{D}^x\mathbf{A} \!-\! \mathbf{X} \Vert}_{F}^2$ and $E_{sp}(\mathbf{Y}) \!=\! {\Vert \mathbf{D}^y\mathbf{B} \!-\! \mathbf{Y} \Vert}_{F}^2$ be sparse representation errors, $E_{ot}(\mathbf{a}) \!=\! {\Vert \mathbf{D}^x\mathbf{a} \!-\!\mathbf{X}\mathbf{1}_N \Vert}_{2}^2$ and $E_{ot}(\mathbf{b}) \!=\! {\Vert \mathbf{D}^y\mathbf{b} \!-\!\mathbf{Y}\mathbf{1}_M \Vert}_{2}^2$ be errors of OT constrains of distributions $\boldsymbol{a}$ and $\boldsymbol{b}$, and the transport cost $E_{c}(\mathbf{T}) \!=\! \langle \mathbf{C}, \mathbf{T}\rangle$, the normalized distributions $\boldsymbol{a}$ and $\boldsymbol{b}$ of two dictionaries $\mathbf{D}^x$ and $\mathbf{D}^y$, and the optimal transport plan $\mathbf{T}$ are illustrated in \ref{fig6} (left), and the loss curves are plotted with iterations in \ref{fig4} (right). It takes around 20$\sim$50 iterations to converge the stable solution. The configurations enable us to produce acceptable results in most cases.  It takes around 2s for reconstructing a pair of $512 \times 512$ resolution color images, while it takes 5s to update coupled dictionaries and the transport in each iteration.The implementation is based on our Matlab 2015b with a desktop PC, Intel i7-9800X CPU 3.80GHz and 64G RAM. 

\subsection{Color Transform}

We first show the color transform performance against two OT-based methods: Monge-Kantorovitch linear (MKL) mapping~\cite{pitie2007linear} and regularized discrete optimal transfer (ROT)~\cite{ferradans2014regularized} respectively. Due to the high computational cost of large-scale OT problems, the transformation maps in both cases are firstly derived on sub-samplings, and post-processing such as interpolation and filtering method is then applied for image reconstruction. We also compare the results with neural color transfer methods: PhotoNAS~\cite{an2020ultrafast} and neural preset~\cite{ke2023neural}, respectively. As stated therein, both of them employ specially-designed network architectures and are trained on huge amount datasets to avoid artifacts with cutting-edge color mapping effects.

\begin{figure*}[!t]
	\centering
	\begin{subfigure}{0.192\linewidth}
            \small
		\begin{minipage}{\textwidth}
			\centering
			\begin{tabular}{c}
				\diaghead{\theadfont}{Reference \\Images}{ Content \\Images}\\
				\vspace{10mm}
			\end{tabular}
		\end{minipage}
		\hfill
		\includegraphics[width=\textwidth]{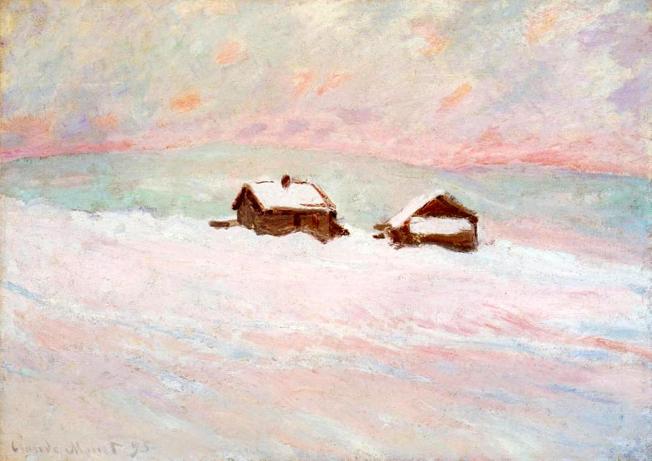}
		\includegraphics[width=\textwidth]{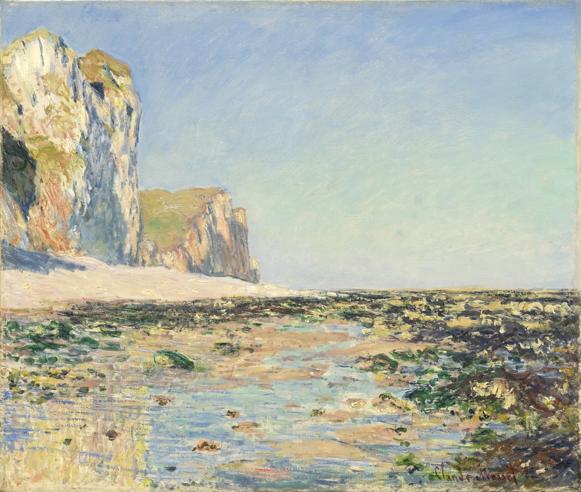}
		\includegraphics[width=\textwidth]{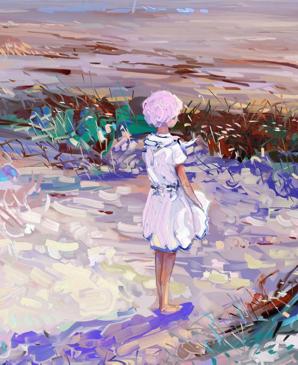}
		\includegraphics[width=\textwidth]{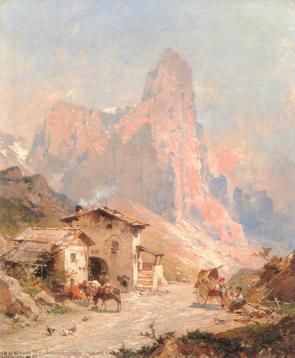} 
		\includegraphics[width=\textwidth]{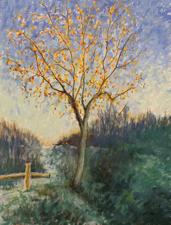} 
		\caption{Input}
		\label{fig7-a}
	\end{subfigure}
	\hfill
	\begin{subfigure}{0.192\linewidth}
		\includegraphics[width=\textwidth, height=0.11\textheight]{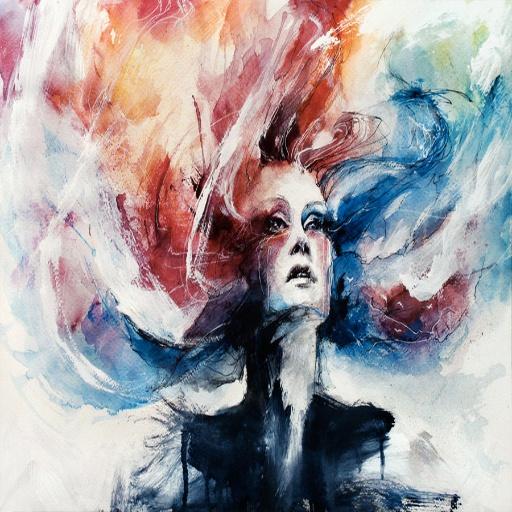}
		\includegraphics[width=\textwidth]{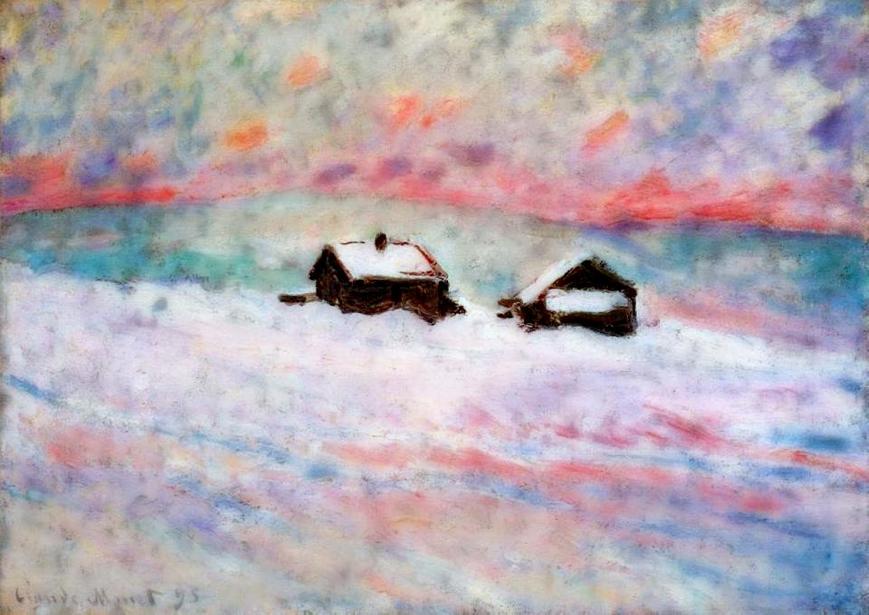}
		\includegraphics[width=\textwidth]{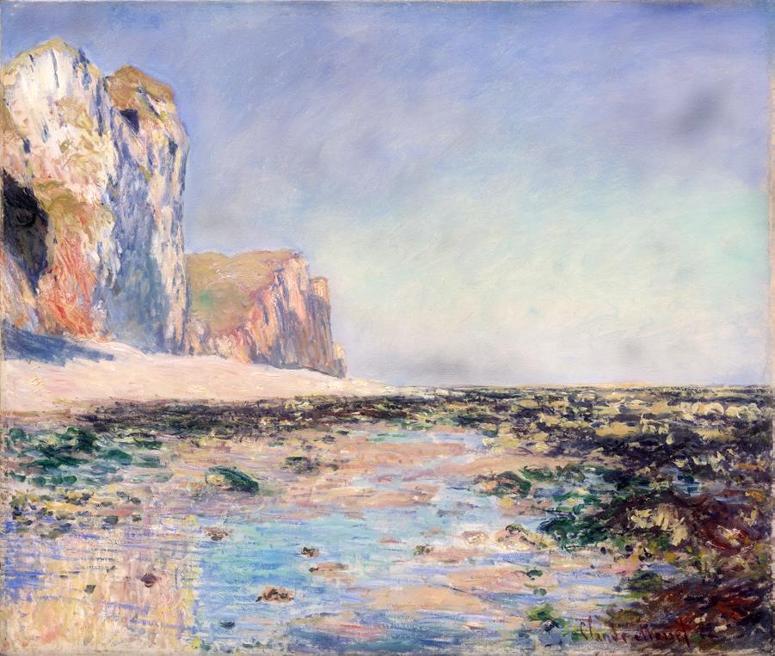}
		\includegraphics[width=\textwidth]{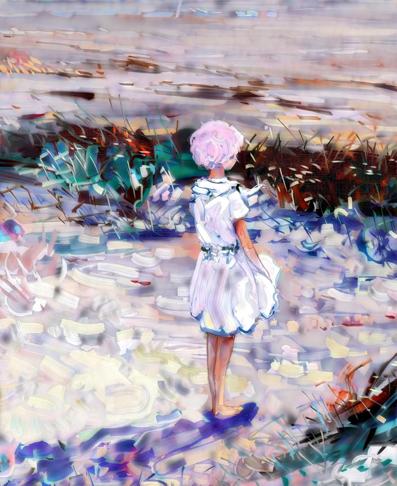}
		\includegraphics[width=\textwidth]{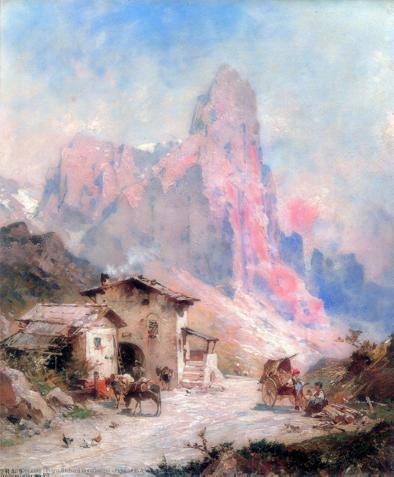}
		\includegraphics[width=\textwidth]{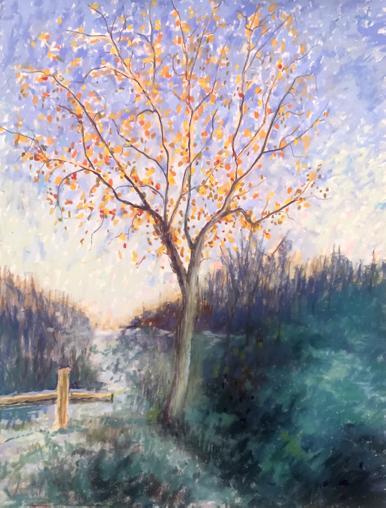}
		\caption{Style 1}
		\label{fig7-b}
	\end{subfigure}
	\hfill
	\begin{subfigure}{0.192\linewidth}
		\includegraphics[width=\textwidth, height=0.11\textheight]{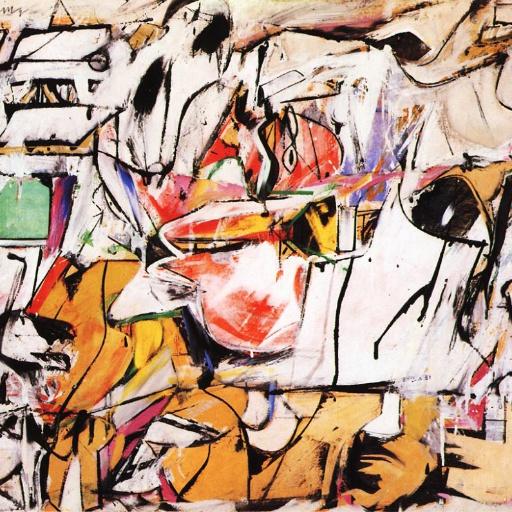}
		\includegraphics[width=\textwidth]{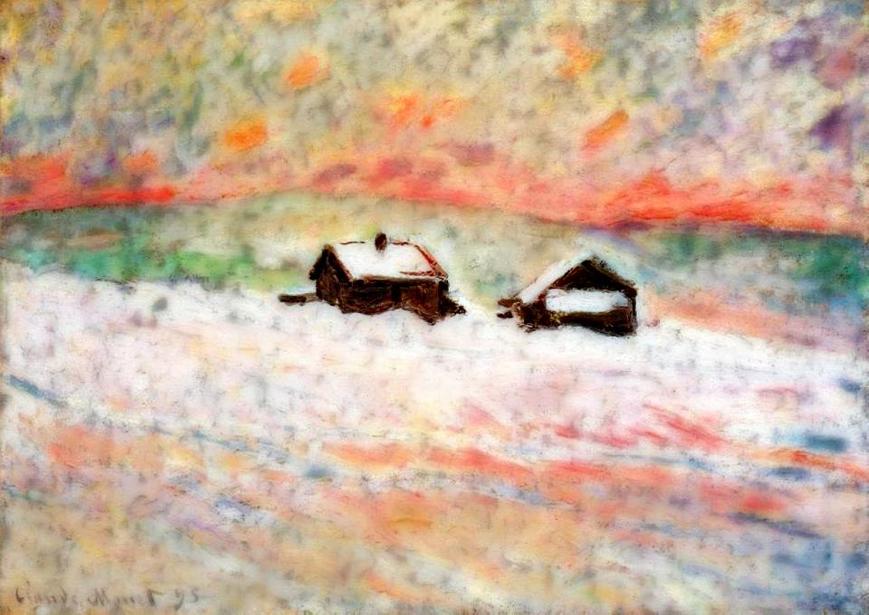}
		\includegraphics[width=\textwidth]{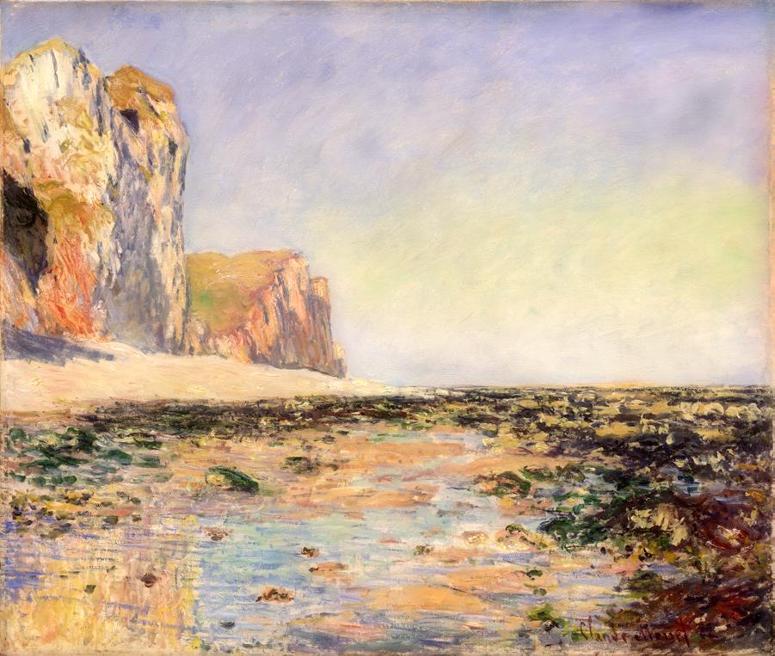}
		\includegraphics[width=\textwidth]{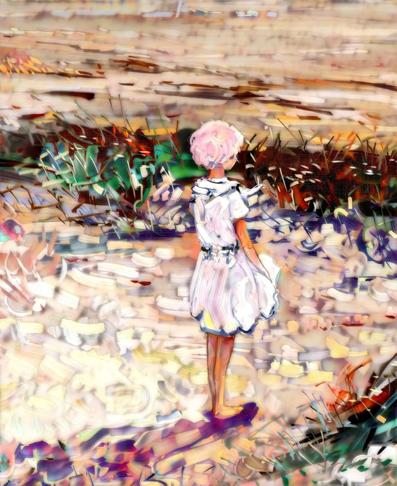}
		\includegraphics[width=\textwidth]{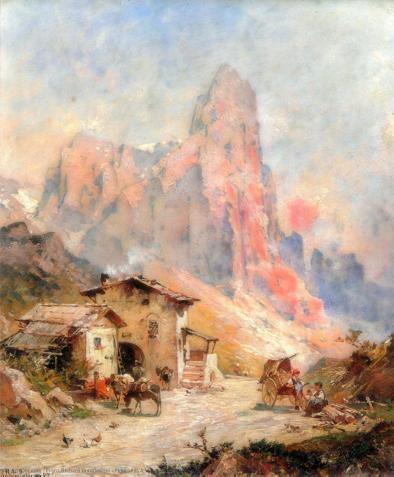}
		\includegraphics[width=\textwidth]{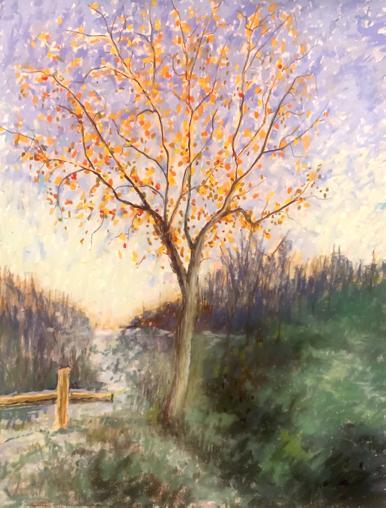}
		\caption{Style 2}
		\label{fig7-c}
	\end{subfigure}
	\hfill
	\begin{subfigure}{0.192\linewidth}
		\includegraphics[width=\textwidth, height=0.11\textheight]{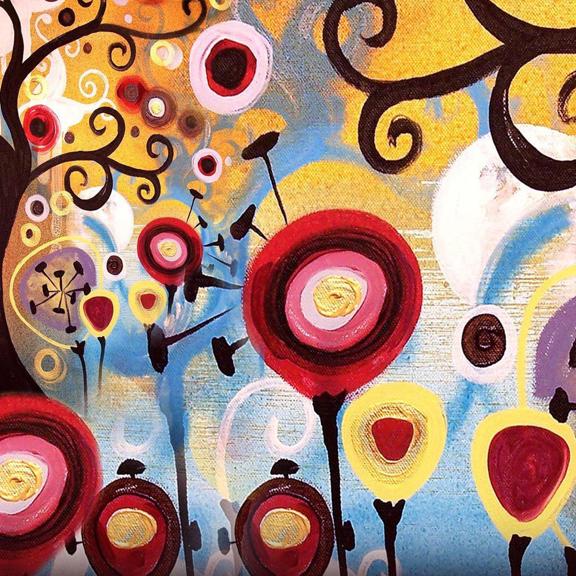}
		\includegraphics[width=\textwidth]{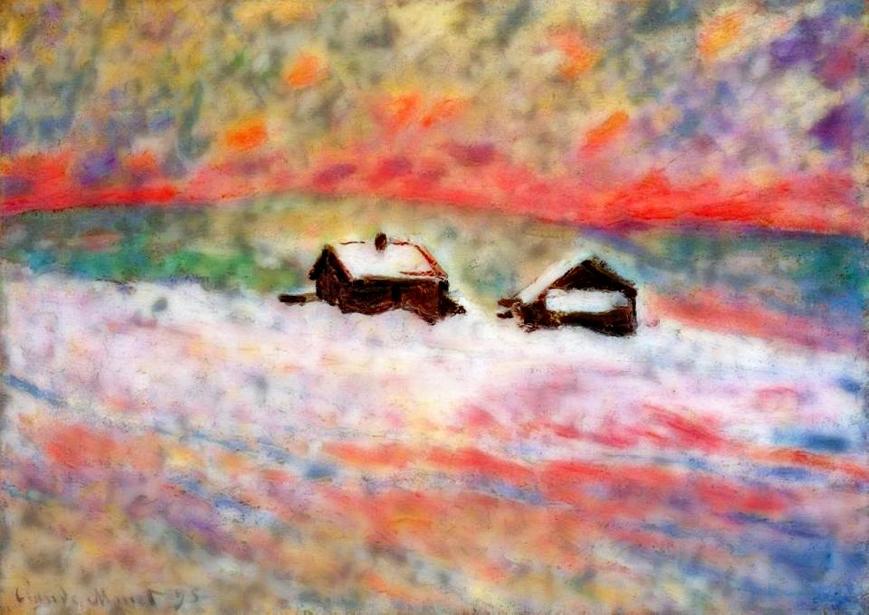}
		\includegraphics[width=\textwidth]{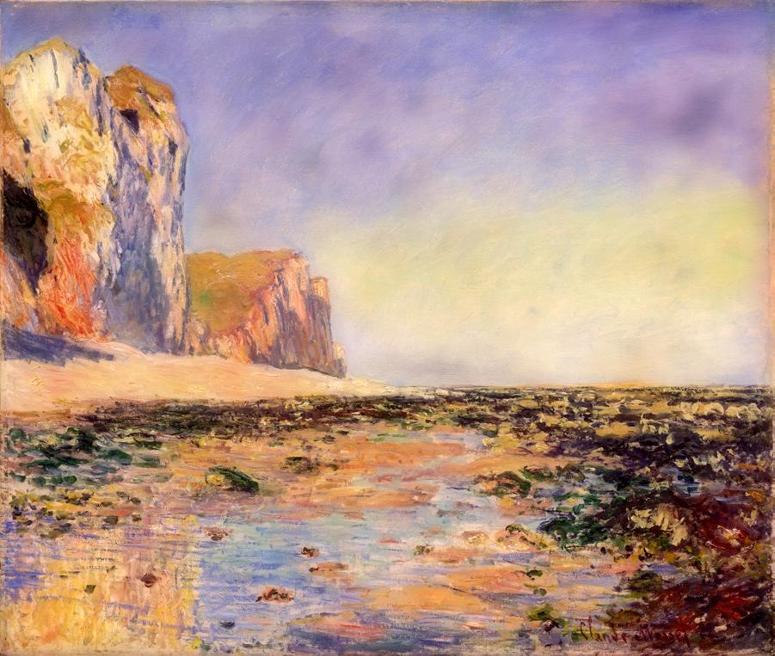}
		\includegraphics[width=\textwidth]{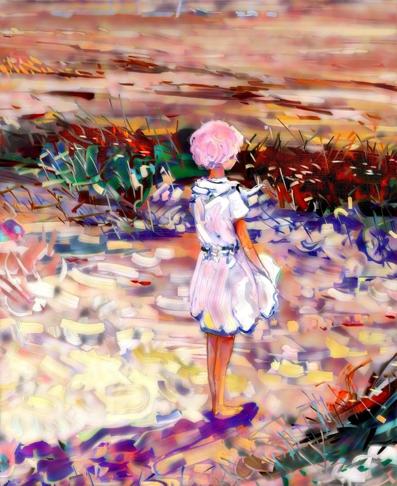}
		\includegraphics[width=\textwidth]{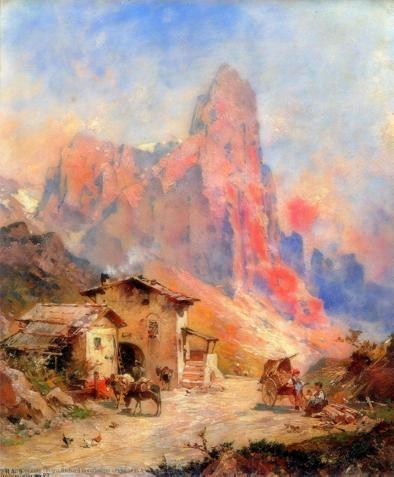}
		\includegraphics[width=\textwidth]{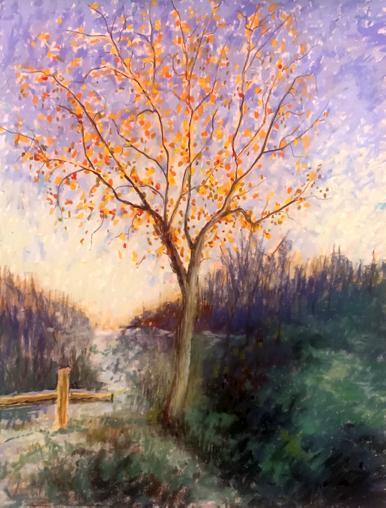}
		\caption{Style 3}
		\label{fig7-d}
	\end{subfigure}
	\hfill
	\begin{subfigure}{0.192\linewidth}
		\includegraphics[width=\textwidth, height=0.11\textheight]{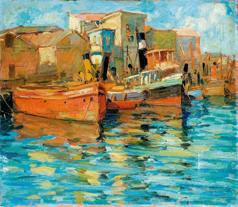}
		\includegraphics[width=\textwidth]{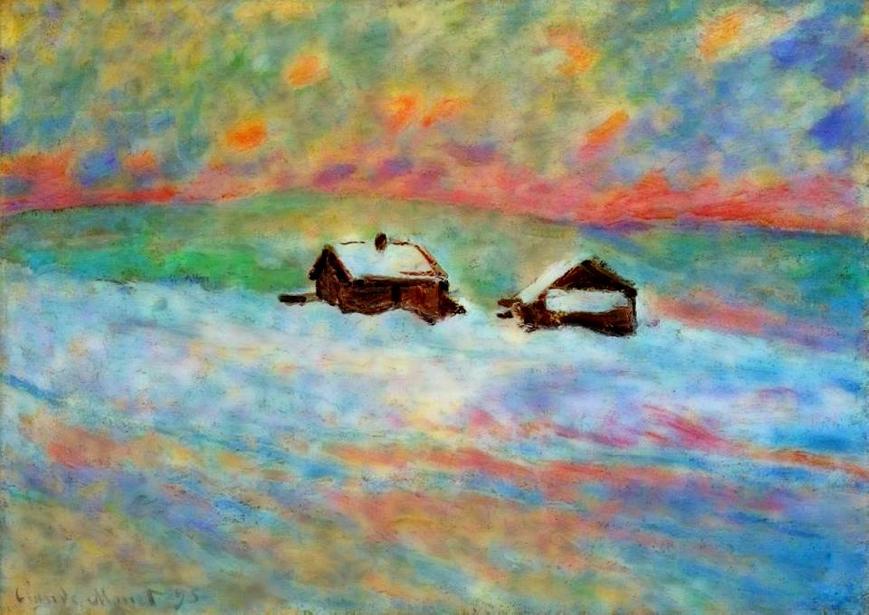}
		\includegraphics[width=\textwidth]{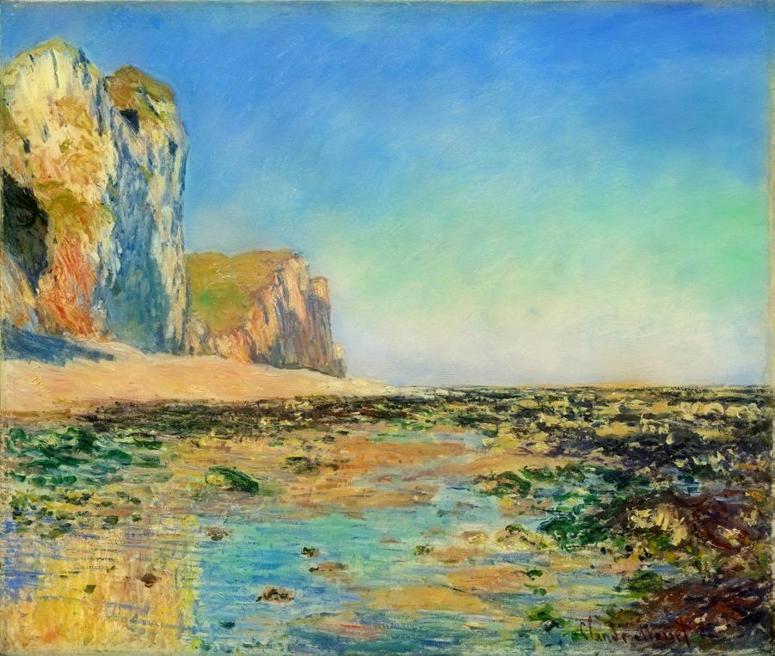}
		\includegraphics[width=\textwidth]{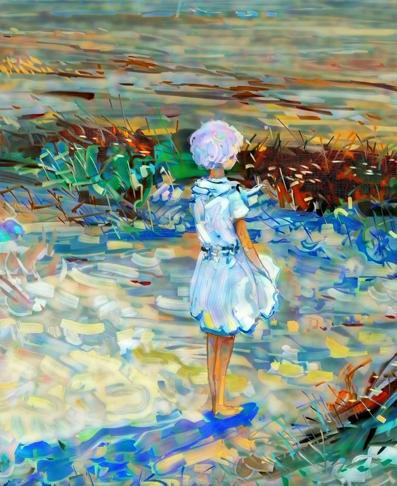}
		\includegraphics[width=\textwidth]{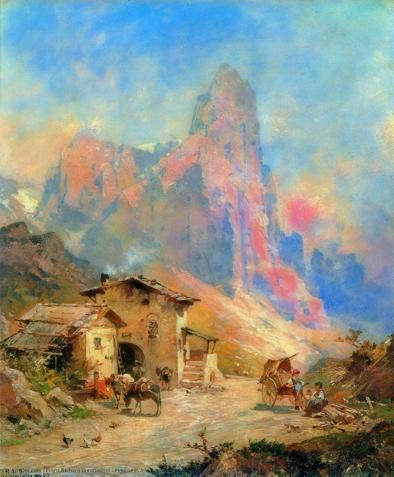}
		\includegraphics[width=\textwidth]{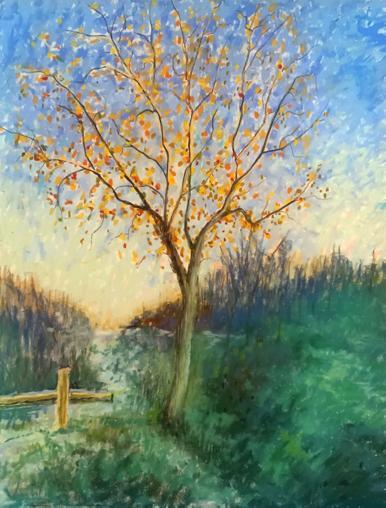}
		\caption{Style 4}
		\label{fig7-e}
	\end{subfigure}
	\hfill
	\caption{Visual results of artistic style transfer of the proposed method. The content images in the first column are transferred according to the given styles in the first row.  (Zoom in for better view).}
	\label{fig7}
\end{figure*}

As shown in Fig.\ref{fig6}, the OT-based MKL mapping~\cite{pitie2007linear} and ROT model~\cite{ferradans2014regularized} faithfully convert the content color in both cases according to the guidance of the reference images. The former reveals color degradation with over-smoothing details (Fig. \ref{fig6-b}), while the latter exhibits block artifacts (Fig. \ref{fig6-c}) due to the sub-sampling strategy, although they can be partially rectified by some filtering methods. The neural transfer methods~\cite{an2020ultrafast,  ke2023neural} produce fine details but they may suffer from inappropriate color mapping, leading to over-saturated or under-saturated effects. In contrast, the proposed model (Fig. \ref{fig6-f})  greatly alleviates the drawbacks for better results due to the strategy of simultaneous representation and transformation, while the improvement is achieved at the expense of high computational cost of dictionary training and image reconstruction compared with the sub-sampling strategy.  

\begin{table*}[!t]
	\small
	\caption{Quantitative comparison of style transfer methods. The best two results are highlighted in \textbf{bold} and \underline{underlined}, respectively.}
	\label{Tab:tab1}
	\vspace{-2mm}
	\begin{center}
		\setlength{\tabcolsep}{1.25mm}{
			\begin{tabular}{|c|c|c|c|c|c|c|c|c|} 
				\hline
				Methods &\makecell{Metrics}  & \makecell{AdaIN~\cite{huang2017arbitrary}} & \makecell{WCT~\cite{li2017universal}} &  \makecell{Photo WCT~\cite{li2018closed}}  & \makecell{WCT$^2$~\cite{yoo2019photorealistic}} & \makecell{StyTr$^2$~\cite{deng2022stytr2}} & \makecell{QuantArt~\cite{huang2023quantart}} & Ours \\ 
				\hline
				\hline
				\multirow{2}{*}{\makecell{Pixel \\level}} & \makecell{SSIM (edge)} $\uparrow$   & 0.5785    & 0.6174   & 0.6773    &0.6160   & 0.4944   & \underline{0.7590}   & \textbf{0.7934} \\
				&\makecell{IIT loss~\cite{huang2022intrinsic}} $\downarrow$   & 61.35     & 44.82    & 35.34   & \underline{34.51}   & 50.02   & 41.38   & \textbf{32.87}\\
				\hline   
				\multirow{3}{*}{\makecell{Feature \\level}} & \makecell{Gram loss~\cite{huang2017arbitrary}} $\downarrow$  	& 1.64      &1.87     &1.45   & \textbf{1.28}    &1.51    & 2.17 	&\underline{1.37} \\ 
				&\makecell{LPIPS loss~\cite{zhang2018unreasonable}} $\downarrow$       & 0.5260    & 0.5645   &\underline{0.2284}   &0.2885   &0.4532    &0.4257      &\textbf{0.2146} \\
				&\makecell{FID metric~\cite{heusel2017gans}} $\downarrow$         & 278.45    &272.57    &153.81   &152.49    &246.75    & \textbf{146.50} 	&\underline{152.20} \\ 
				\hline
		\end{tabular}}
	\end{center}
	\vspace{-2mm}
\end{table*}

\begin{figure*}[!t]
	\centering
	\begin{subfigure}{0.12\linewidth}
		\includegraphics[width=\textwidth, height=0.08\textheight]{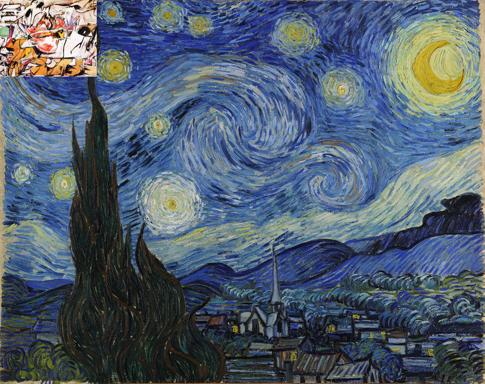}\hfill
		\includegraphics[width=\textwidth, height=0.08\textheight]{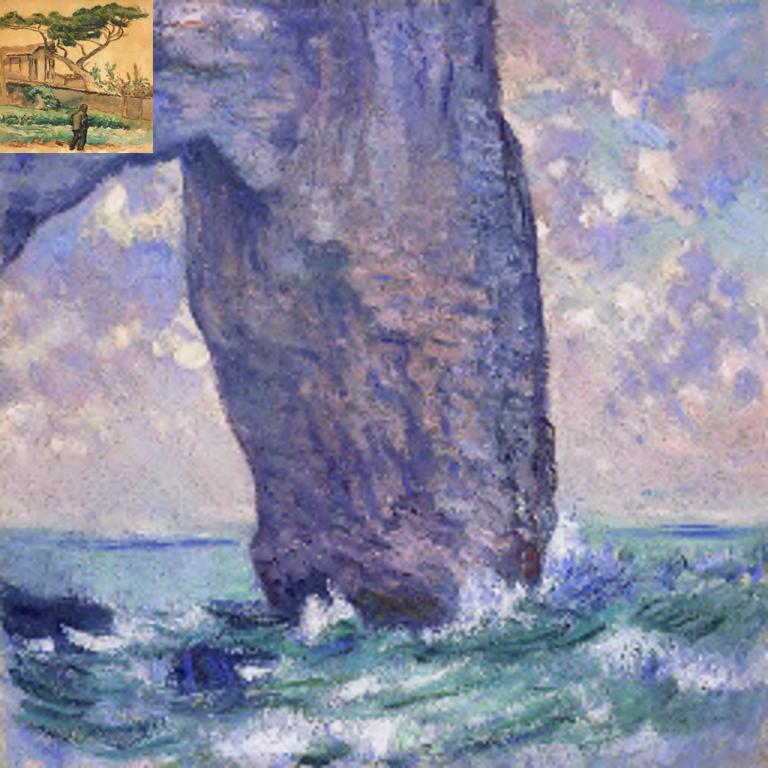}\hfill
		\includegraphics[width=\textwidth, height=0.08\textheight]{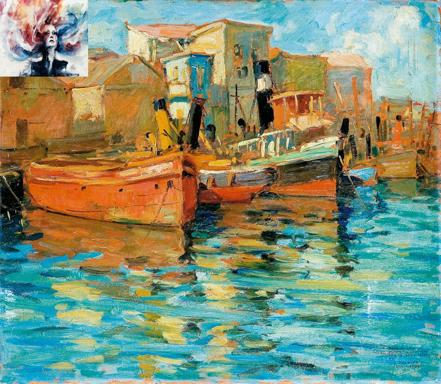}\hfill
		\includegraphics[width=\textwidth, height=0.11\textheight]{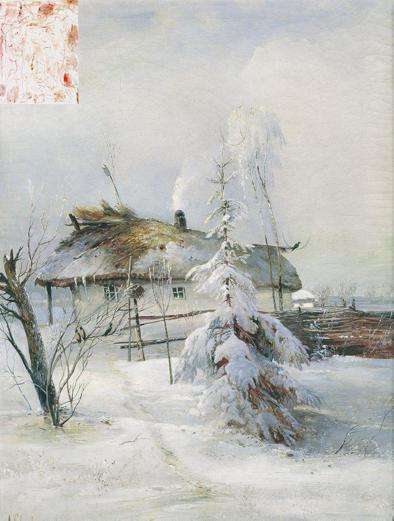}\hfill
  		\includegraphics[width=\textwidth, height=0.11\textheight]{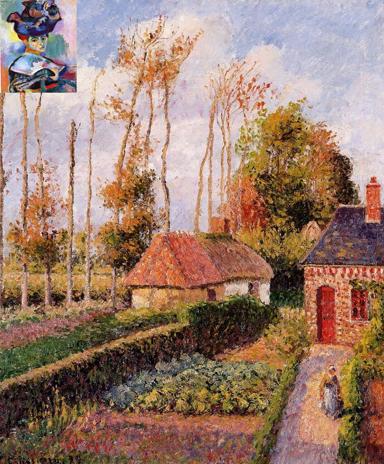}\hfill
		\includegraphics[width=\textwidth, height=0.07\textheight]{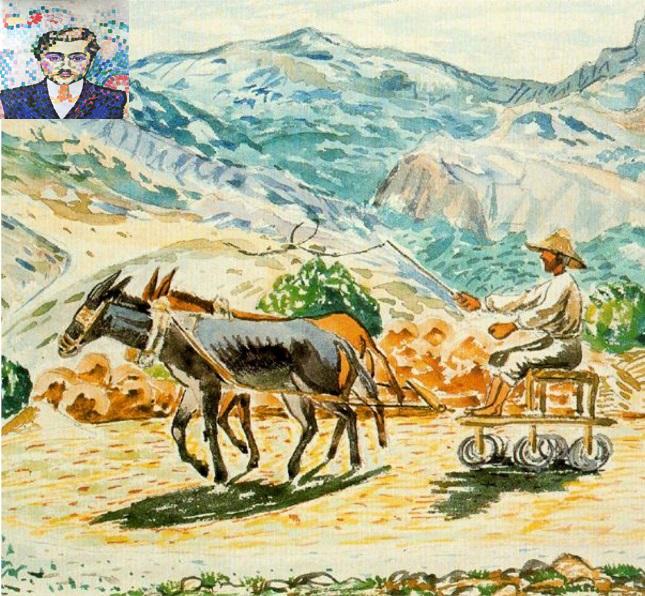}\hfill
		\includegraphics[width=\textwidth, height=0.07\textheight]{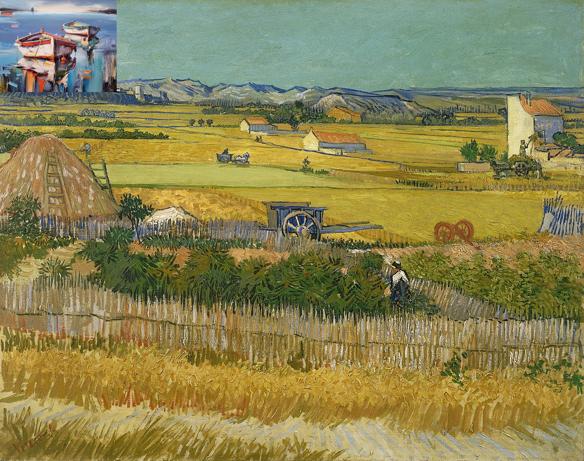}\hfill
		\includegraphics[width=\textwidth, height=0.07\textheight]{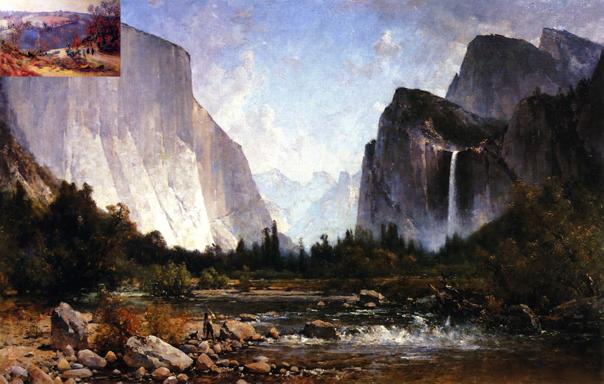}\hfill
		\caption{}
		\label{fig8-a}
	\end{subfigure}
	\begin{subfigure}{0.12\linewidth}
		\includegraphics[width=\textwidth, height=0.08\textheight]{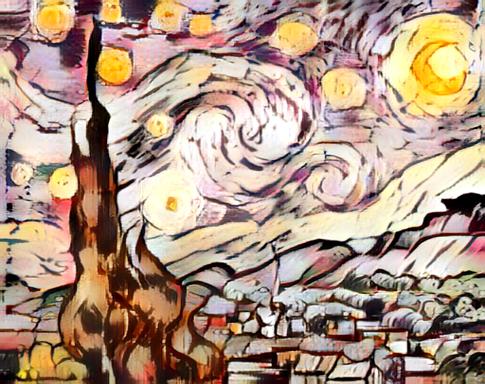}\hfill
		\includegraphics[width=\textwidth, height=0.08\textheight]{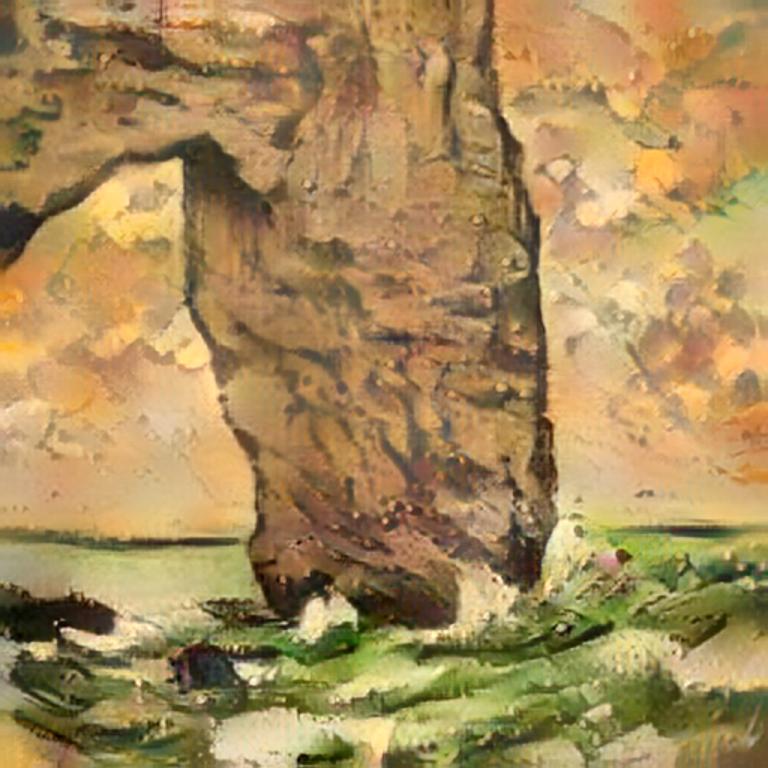}\hfill
		\includegraphics[width=\textwidth, height=0.08\textheight]{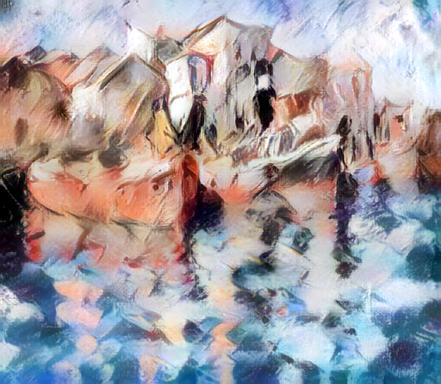}\hfill
		\includegraphics[width=\textwidth, height=0.11\textheight]{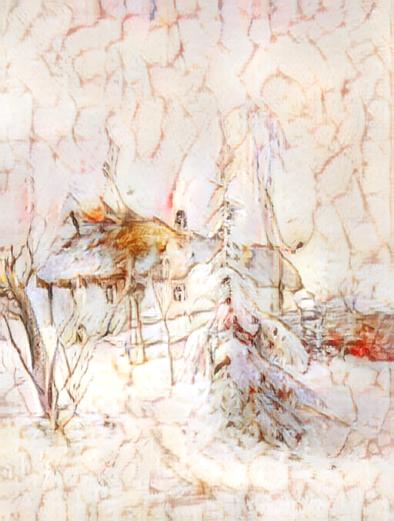}\hfill
		\includegraphics[width=\textwidth, height=0.11\textheight]{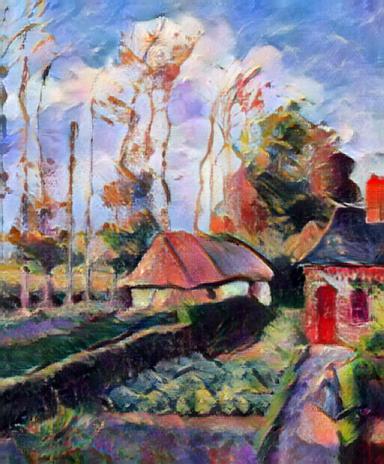}\hfill
		\includegraphics[width=\textwidth, height=0.07\textheight]{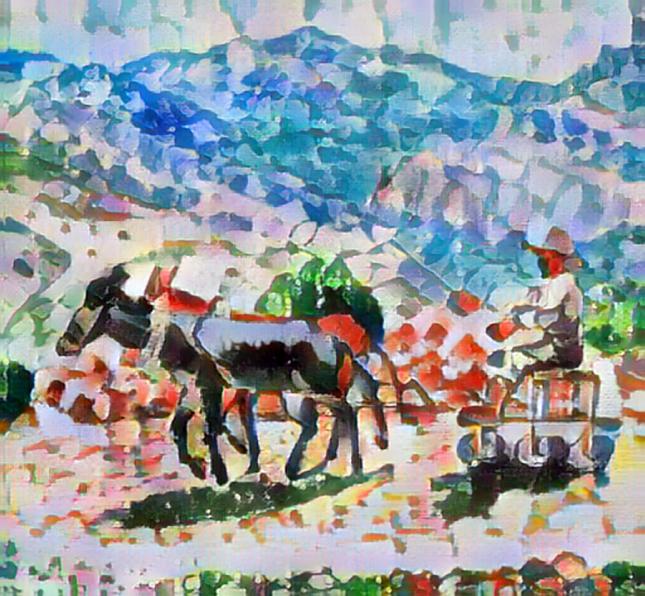}\hfill
		\includegraphics[width=\textwidth, height=0.07\textheight]{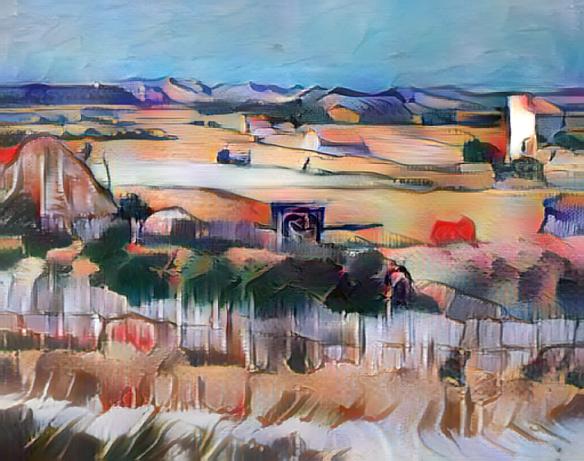}\hfill
		\includegraphics[width=\textwidth, height=0.07\textheight]{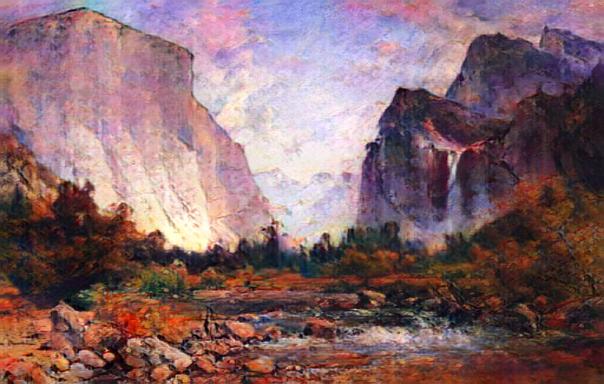}\hfill
		\caption{}
		\label{fig8-b}
	\end{subfigure}
	\begin{subfigure}{0.12\linewidth}
		\includegraphics[width=\textwidth, height=0.08\textheight]{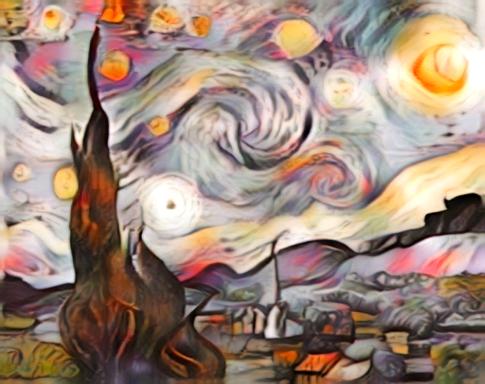}\hfill
		\includegraphics[width=\textwidth, height=0.08\textheight]{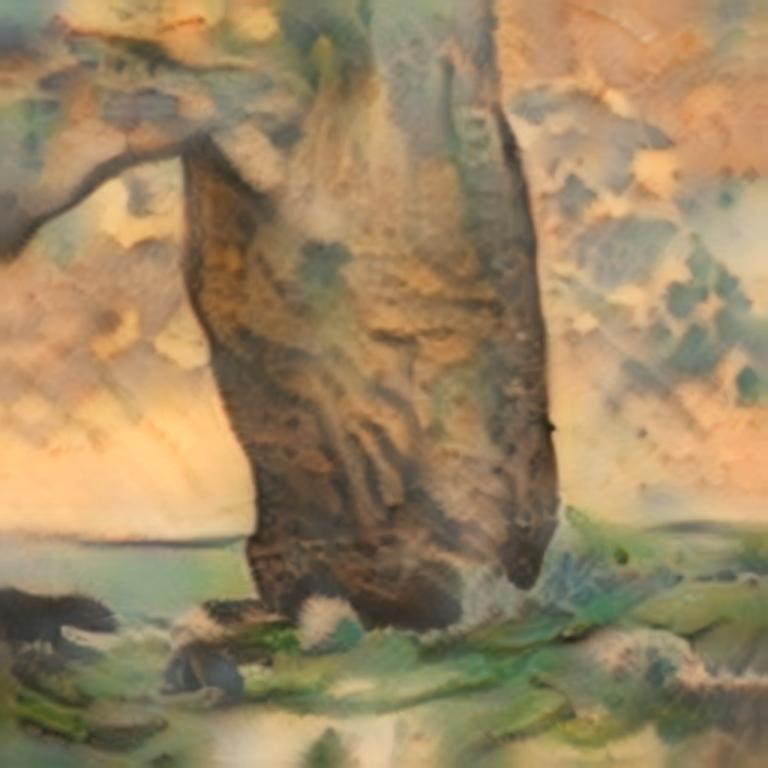}\hfill
		\includegraphics[width=\textwidth, height=0.08\textheight]{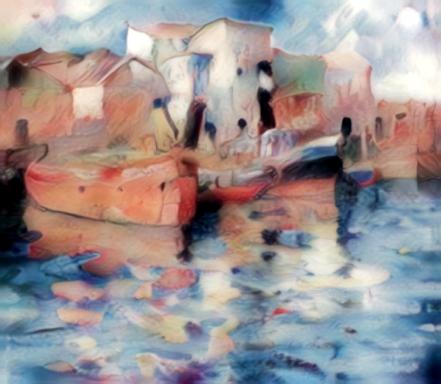}\hfill
		\includegraphics[width=\textwidth, height=0.11\textheight]{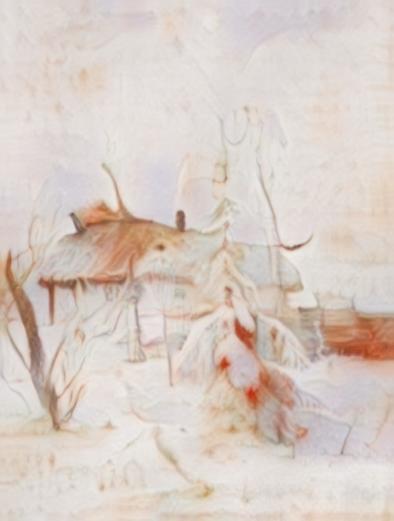}\hfill
		\includegraphics[width=\textwidth, height=0.11\textheight]{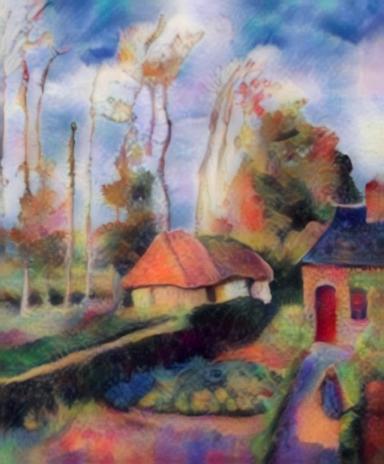}\hfill
		\includegraphics[width=\textwidth, height=0.07\textheight]{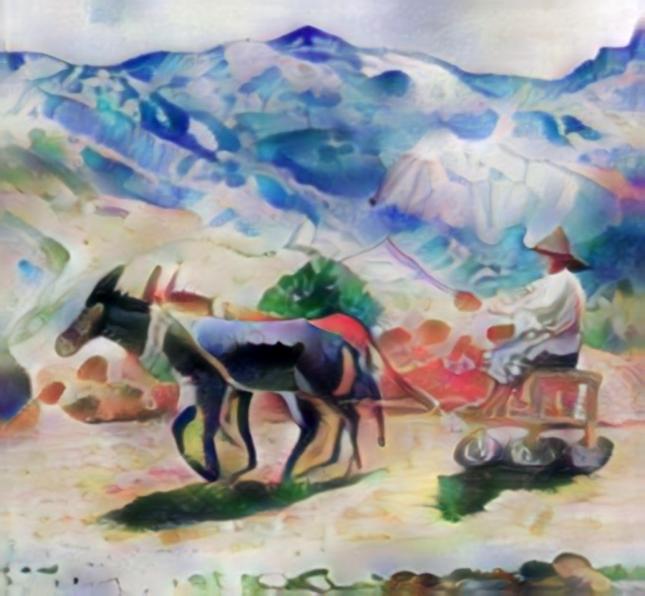}\hfill
		\includegraphics[width=\textwidth, height=0.07\textheight]{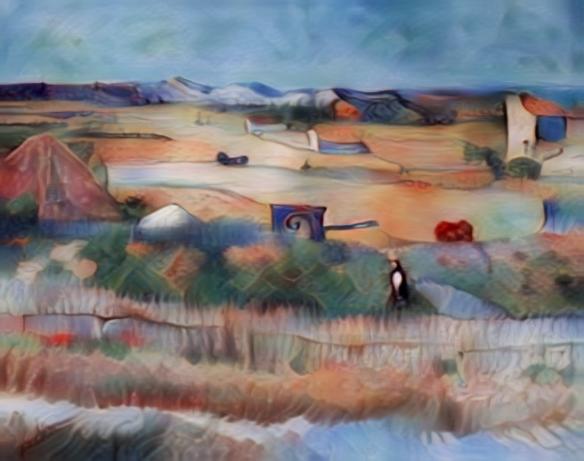}\hfill
		\includegraphics[width=\textwidth, height=0.07\textheight]{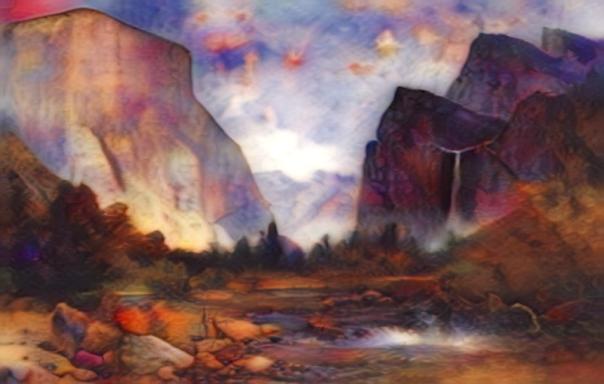}\hfill
		\caption{}
		\label{fig8-c}
	\end{subfigure}
	\begin{subfigure}{0.12\linewidth}
		\includegraphics[width=\textwidth, height=0.08\textheight]{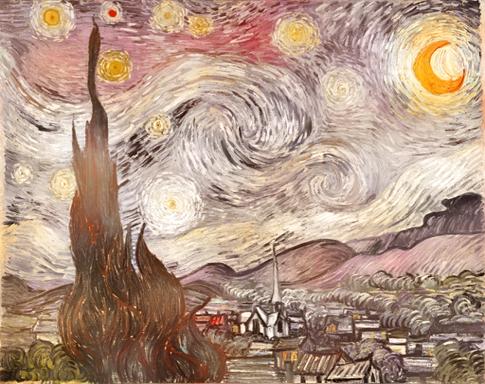}\hfill
		\includegraphics[width=\textwidth, height=0.08\textheight]{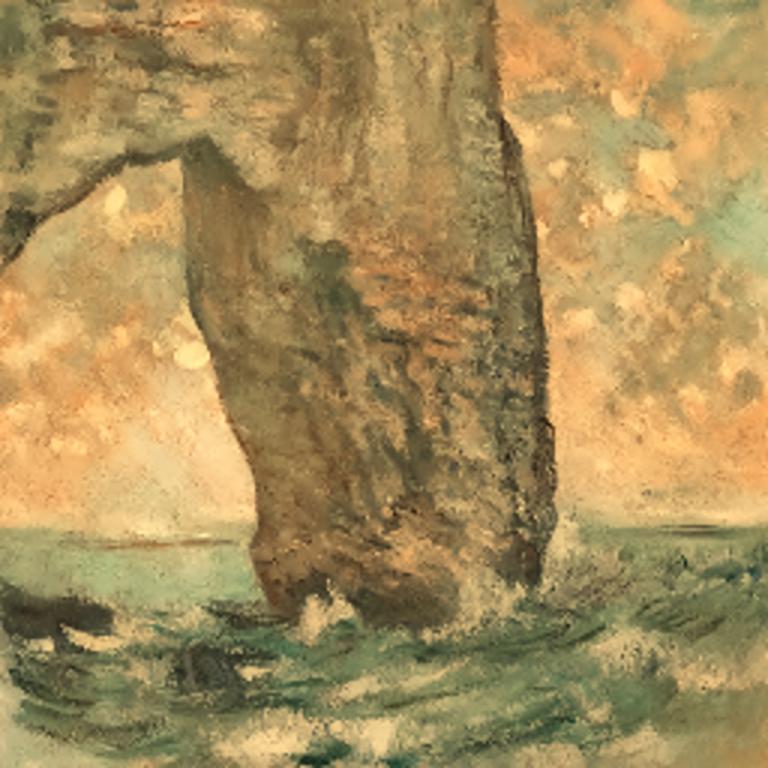}\hfill
		\includegraphics[width=\textwidth, height=0.08\textheight]{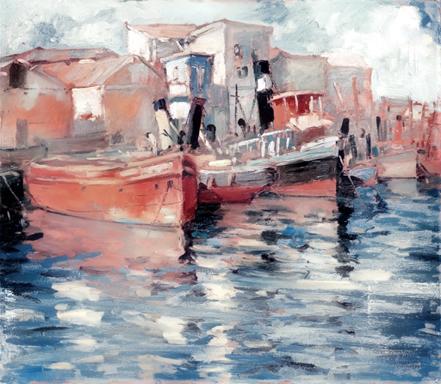}\hfill
		\includegraphics[width=\textwidth, height=0.11\textheight]{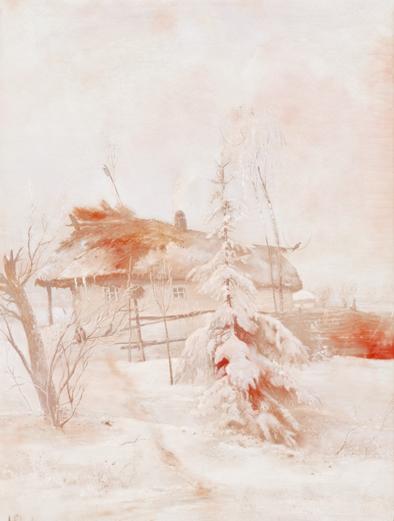}\hfill
		\includegraphics[width=\textwidth, height=0.11\textheight]{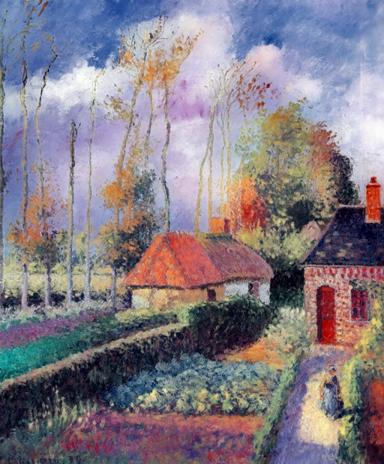}\hfill
		\includegraphics[width=\textwidth, height=0.07\textheight]{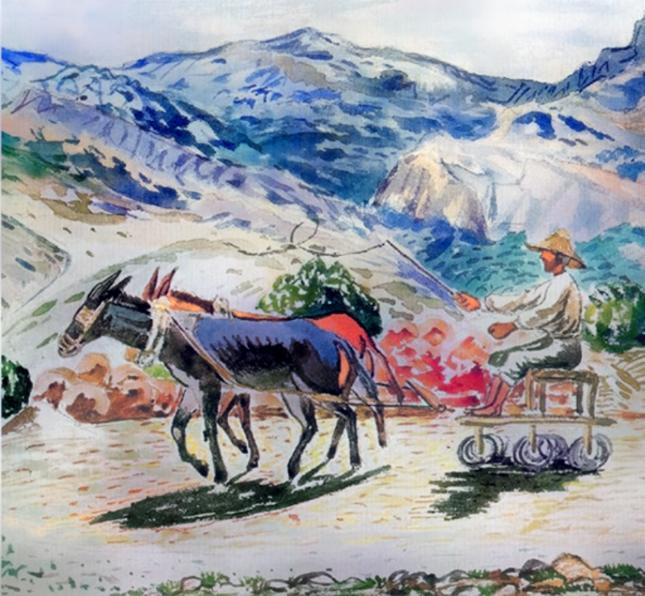}\hfill
		\includegraphics[width=\textwidth, height=0.07\textheight]{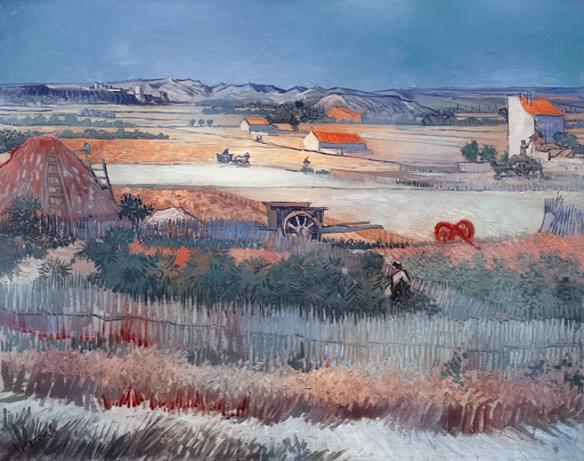}\hfill
		\includegraphics[width=\textwidth, height=0.07\textheight]{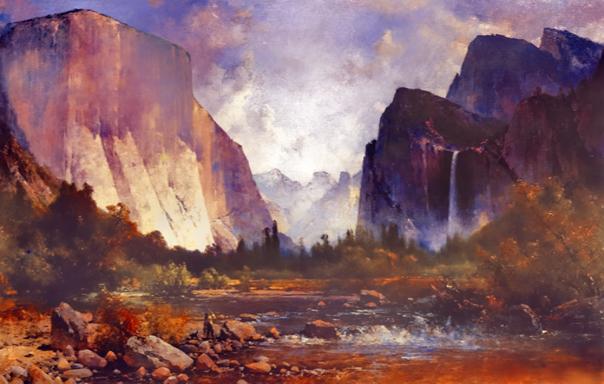}\hfill
		\caption{}
		\label{fig8-d}
	\end{subfigure}
	\begin{subfigure}{0.12\linewidth}
		\includegraphics[width=\textwidth, height=0.08\textheight]{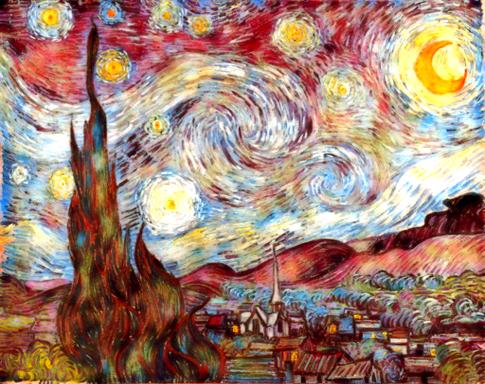}\hfill
		\includegraphics[width=\textwidth, height=0.08\textheight]{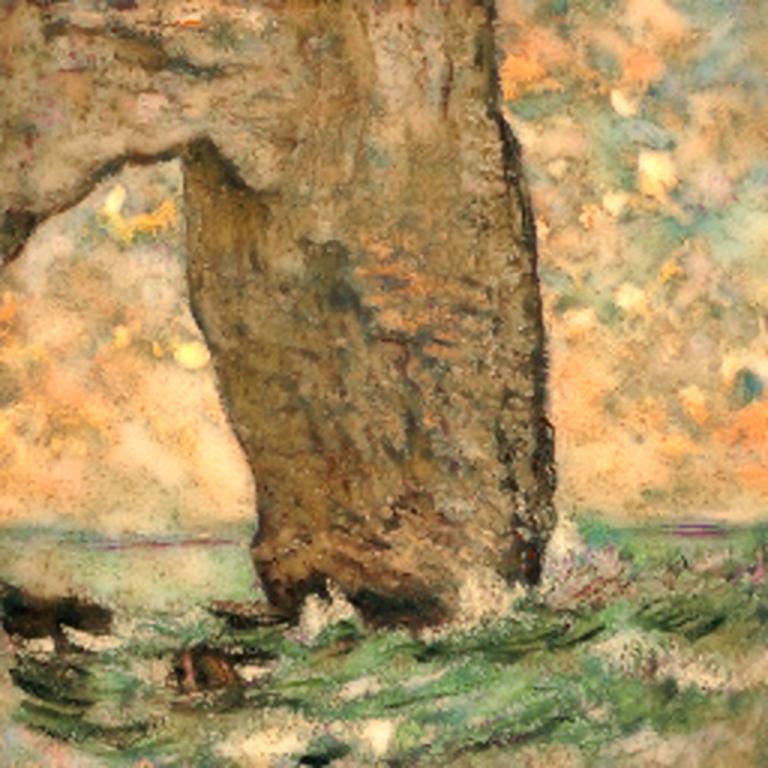}\hfill
		\includegraphics[width=\textwidth, height=0.08\textheight]{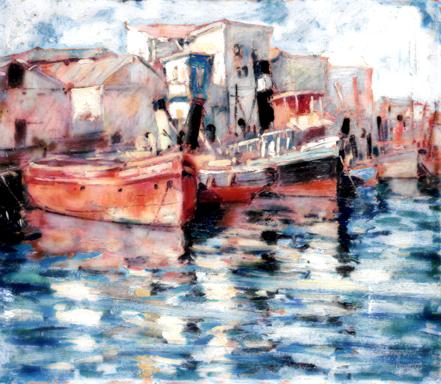}\hfill
		\includegraphics[width=\textwidth, height=0.11\textheight]{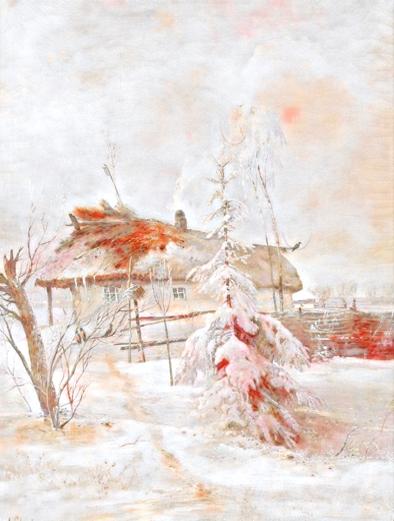}\hfill
		\includegraphics[width=\textwidth, height=0.11\textheight]{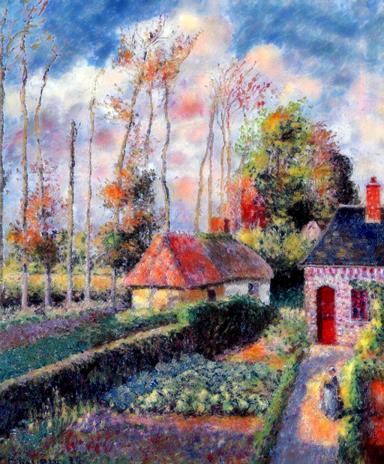}\hfill
		\includegraphics[width=\textwidth, height=0.07\textheight]{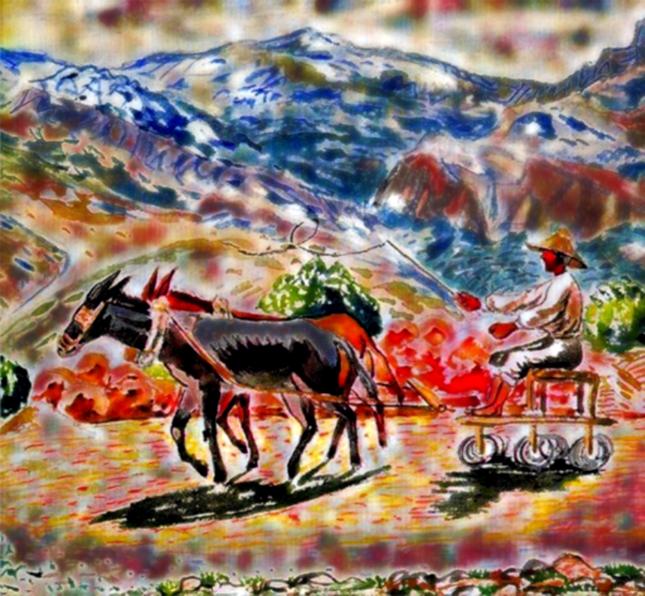}\hfill
		\includegraphics[width=\textwidth, height=0.07\textheight]{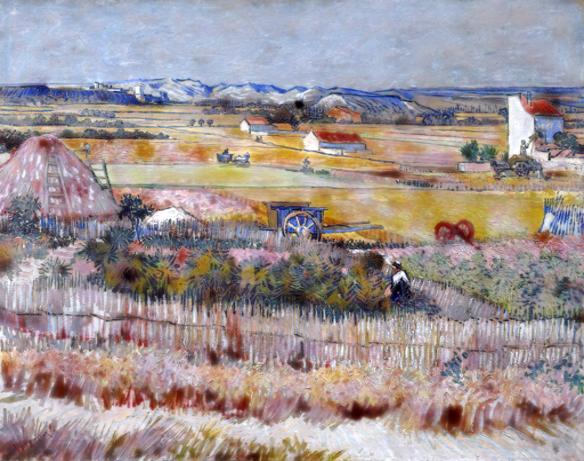}\hfill
		\includegraphics[width=\textwidth, height=0.07\textheight]{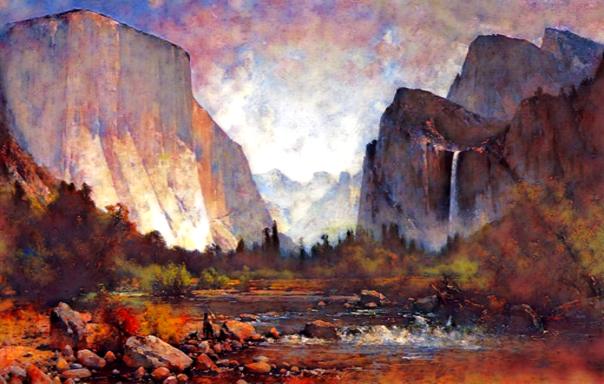}\hfill
		\caption{}
		\label{fig8-e}
	\end{subfigure}
	\begin{subfigure}{0.12\linewidth}
		\includegraphics[width=\textwidth, height=0.08\textheight]{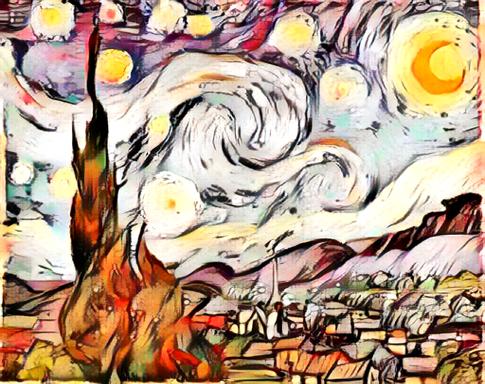}\hfill
		\includegraphics[width=\textwidth, height=0.08\textheight]{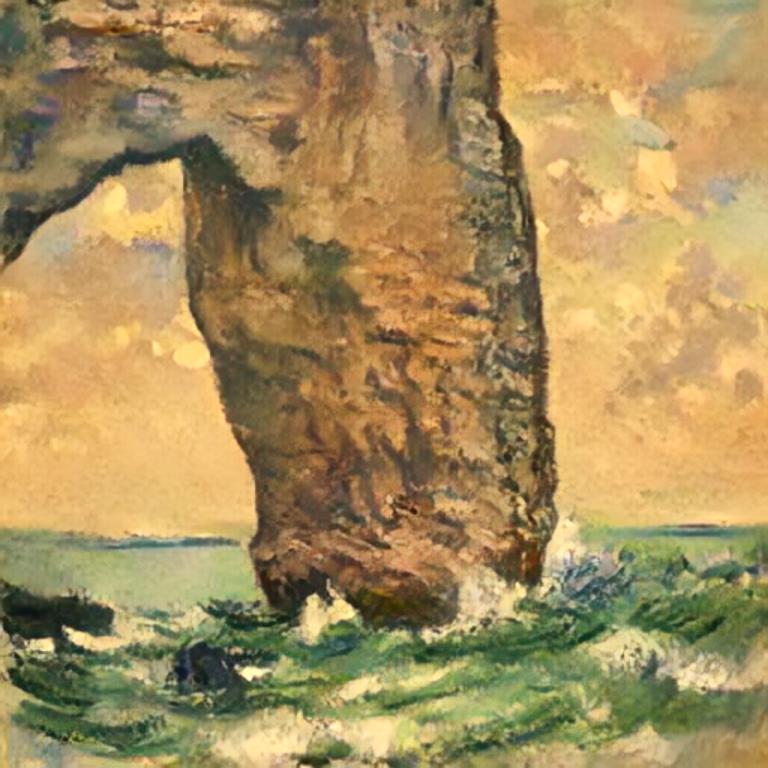}\hfill
		\includegraphics[width=\textwidth, height=0.08\textheight]{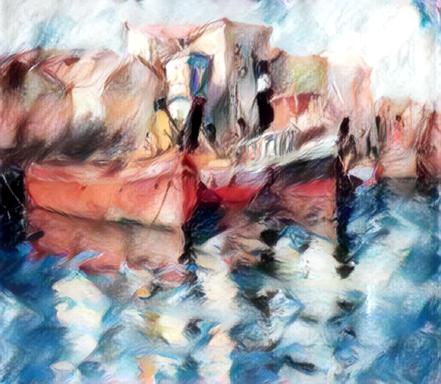}\hfill
		\includegraphics[width=\textwidth, height=0.11\textheight]{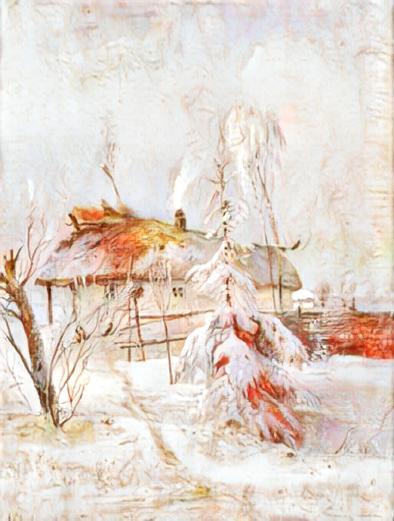}\hfill
		\includegraphics[width=\textwidth, height=0.11\textheight]{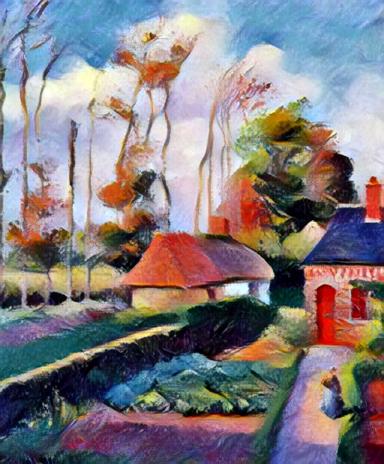}\hfill
		\includegraphics[width=\textwidth, height=0.07\textheight]{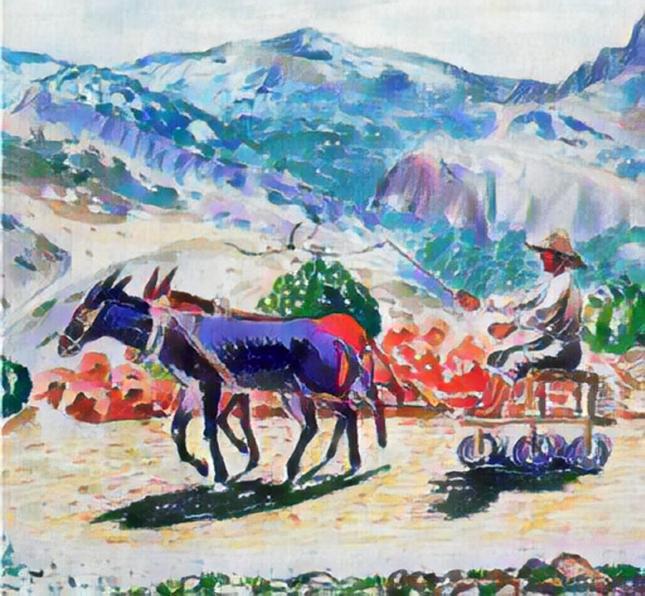}\hfill
		\includegraphics[width=\textwidth, height=0.07\textheight]{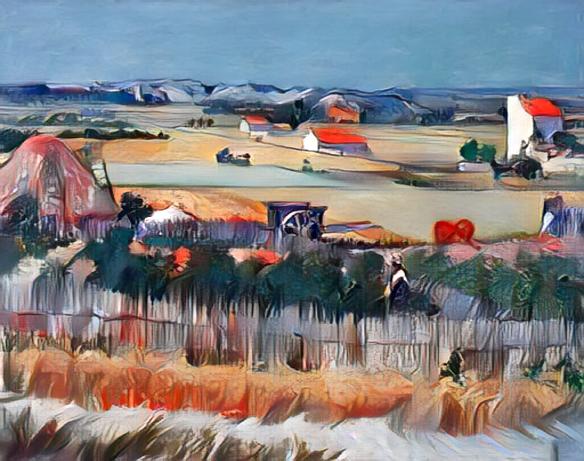}\hfill
		\includegraphics[width=\textwidth, height=0.07\textheight]{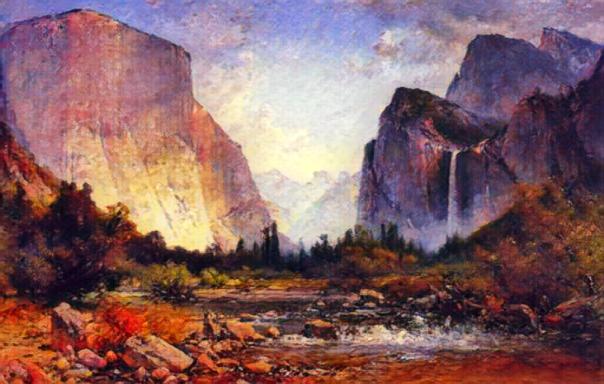}\hfill
		\caption{}
		\label{fig8-f}
	\end{subfigure}
	\begin{subfigure}{0.12\linewidth}
		\includegraphics[width=\textwidth, height=0.08\textheight]{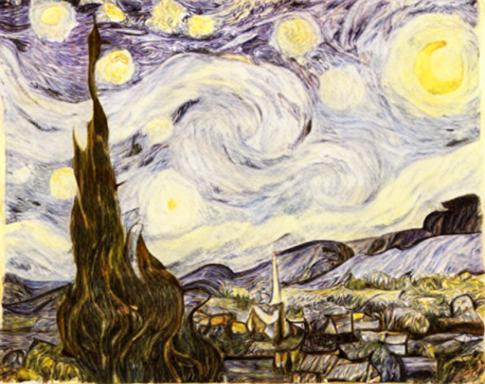}\hfill
		\includegraphics[width=\textwidth, height=0.08\textheight]{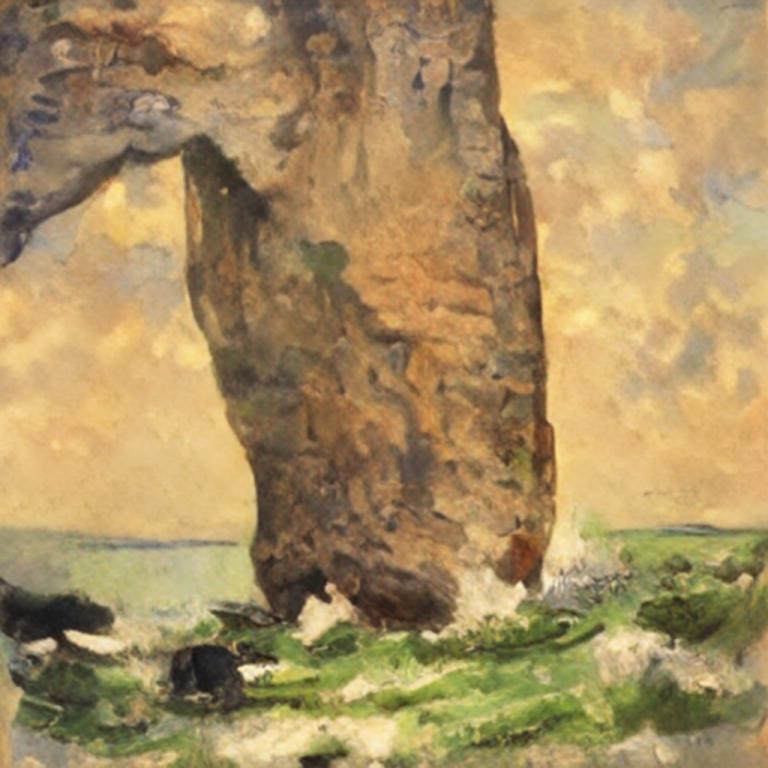}\hfill
		\includegraphics[width=\textwidth, height=0.08\textheight]{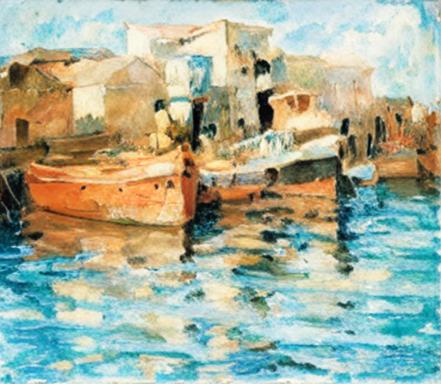}\hfill
		\includegraphics[width=\textwidth, height=0.11\textheight]{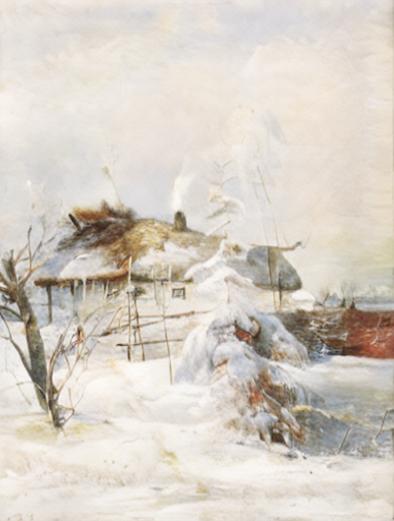}\hfill
		\includegraphics[width=\textwidth, height=0.11\textheight]{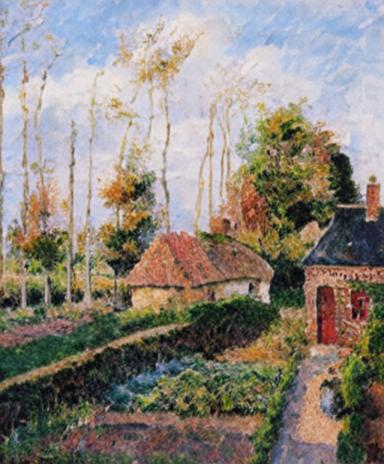}\hfill
		\includegraphics[width=\textwidth, height=0.07\textheight]{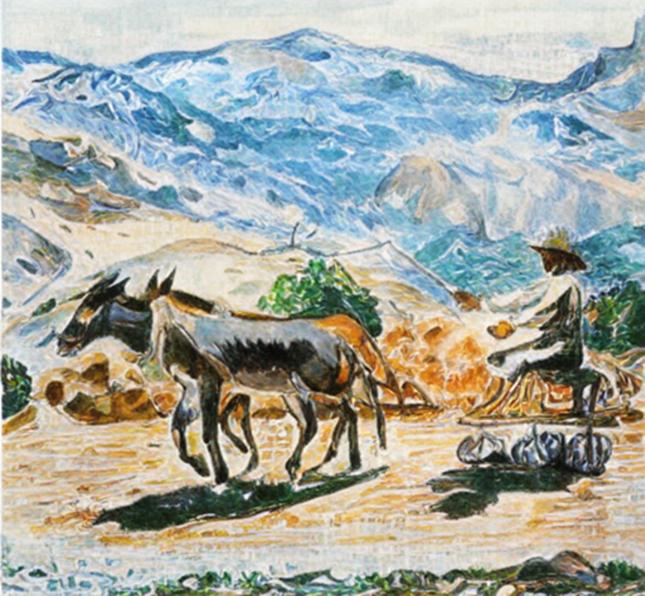}\hfill
		\includegraphics[width=\textwidth, height=0.07\textheight]{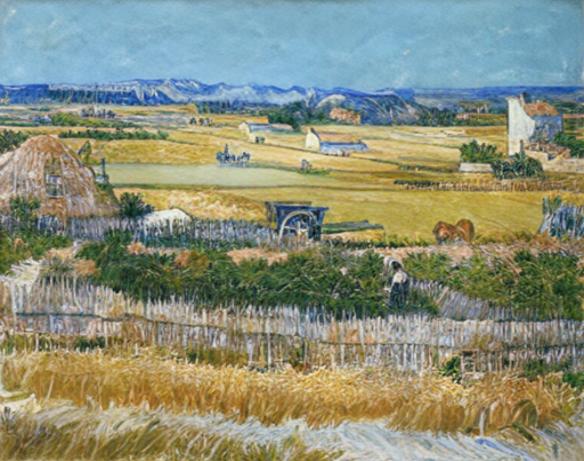}\hfill
		\includegraphics[width=\textwidth, height=0.07\textheight]{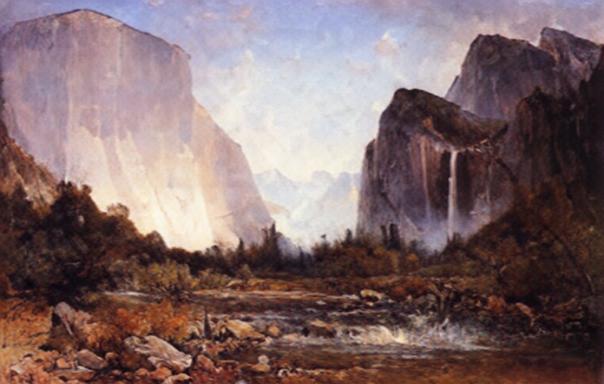}\hfill
		\caption{}
		\label{fig8-g}
	\end{subfigure}
	\begin{subfigure}{0.12\linewidth}
		\includegraphics[width=\textwidth, height=0.08\textheight]{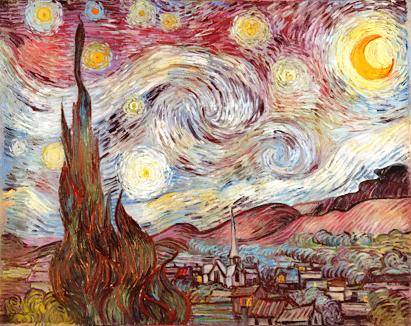}\hfill
		\includegraphics[width=\textwidth, height=0.08\textheight]{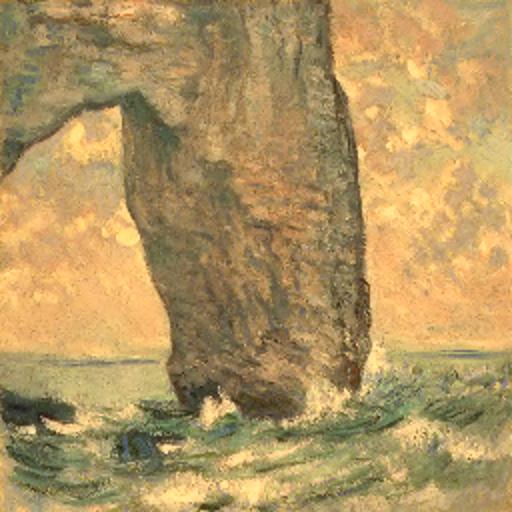}\hfill
		\includegraphics[width=\textwidth, height=0.08\textheight]{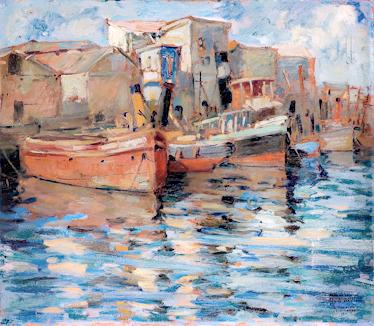}\hfill
		\includegraphics[width=\textwidth, height=0.11\textheight]{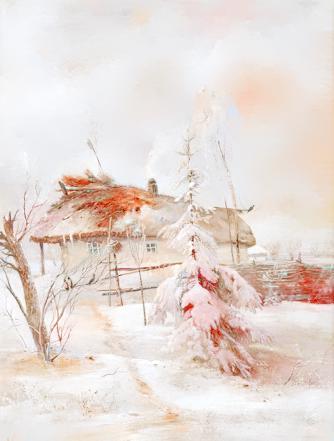}\hfill
		\includegraphics[width=\textwidth, height=0.11\textheight]{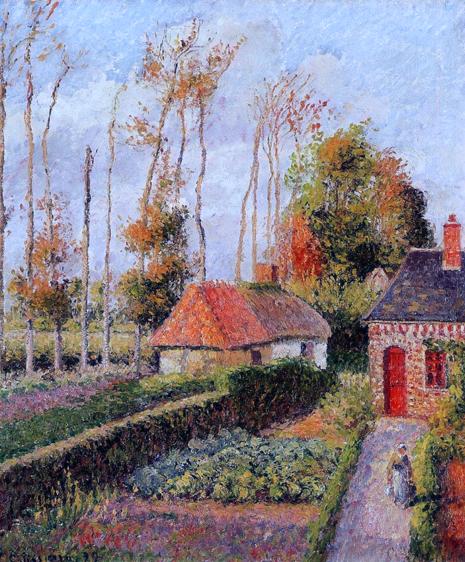}\hfill
		\includegraphics[width=\textwidth, height=0.07\textheight]{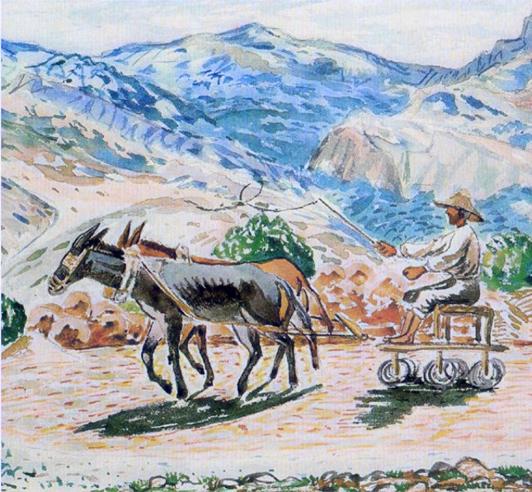}\hfill
		\includegraphics[width=\textwidth, height=0.07\textheight]{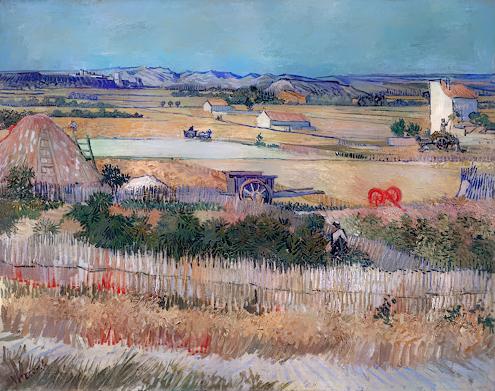}\hfill
		\includegraphics[width=\textwidth, height=0.07\textheight]{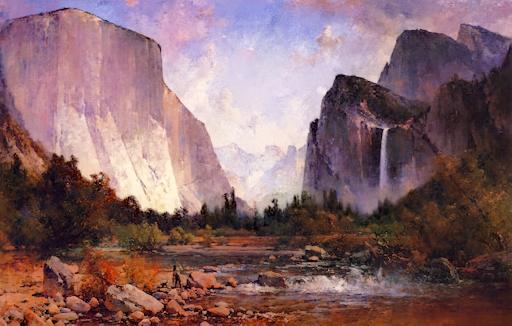}\hfill
		\caption{}
		\label{fig8-h}
	\end{subfigure}
	\caption{Visual comparison of artistic style transfer with deep learning methods: (a) Input content and style images, (b) AdaIN~\cite{huang2017arbitrary}, (c) WCT~\cite{li2017universal}, (d) PhotoWCT~\cite{li2018closed}, (e) WCT$^2$~\cite{yoo2019photorealistic} (f) StyTr$^2$~\cite{deng2022stytr2}, (g) QuantArt~\cite{huang2023quantart} and (h) our results.  (Zoom in for better view).}
    \label{fig8}
\end{figure*}

\subsection{Photo-realistic Style Transfer}

We further show that the proposed model is applicable to artistic style transfer, especially the photo-realistic effects. As shown in Fig. \ref{fig6}, we present the transferred results with the artworks created by famous artists. It is clear that all cases have very complex content information composed of sophisticated painting strokes and multiple colors, and textures for aesthetic effects. To receive visual-pleasant results, we randomly select $100K$ patches from content and reference images, and use $1024$ dictionary elements for training. The optimal transport is solved by the entropy-regularized method~\cite{cuturi2013sinkhorn} with regularization parameter $\gamma = 0.05$ and $200$ iterations for each optimal transport step. The new model is possible to produce visual-pleasant transferred results with consistently aesthetic styles and well-preserved details. In Fig. \ref{fig7}, we additionally show the high-quality model performance in the scenarios of different reference styles, which demonstrate the benefits of sparse representation and optimal transport.

In Fig. \ref{fig8}, we compare the transferred results with deep learning-based methods. The AdaIN model~\cite{huang2017arbitrary} and WCT method~\cite{huang2017arbitrary} are two well-known arbitrary image style transfer methods. As shown in Fig. \ref{fig8-b} and Fig. \ref{fig8-c}, both of them are capable of transforming the global features such as image colors and main structures for impressive results, however, they may have limitations in suppressing non-consistent local details. Recently, the transformer-based architecture shows great attention in many vision-based tasks. We here also include the StyTr$^2$ model~\cite{deng2022stytr2} for comparison, however, it also suffers slight non-consistent local details in Fig. \ref{fig8-d} despite the powerful transformer architecture. In addition, we also compare the transferred results with the photo-realistic methods: PhotoWCT model~\cite{li2018closed}, WCT$^2$ method~\cite{yoo2019photorealistic} and QuantArt~\cite{huang2023quantart}, in which the non-consistent local details can be significantly suppressed as verified in Fig. \ref{fig8-e} $\sim$ Fig. \ref{fig8-g}. Similarly, the proposed method also gives arise to photo-realistic results with more consistent local textures and details in \ref{fig8-h}. In summary, the proposed method is applicable for both natural and artistic images, and the results further demonstrate its ability in retrieving consistent details.

Additionally, we present the quantitative performance against recent deep learning-based methods. Notice that an objective assessment is often difficult due to the lack of ground truth in aesthetic significance. As suggested in recent work~\cite{an2020ultrafast, deng2022stytr2, huang2023quantart}, we adopt the structural similarity (SSIM) of edge maps between content and transferred images to indicate detail preservation ability. We also take the structure fidelity into account based on the intrinsic image transfer (IIT) algorithm~\cite{huang2022intrinsic} in consideration of the robust structure-preserving property in varying illumination (color and brightness) conditions. Meanwhile, we introduce three deep learning-based evaluation metrics: LPIPS loss~\cite{zhang2018unreasonable}, Gram loss (VGG style features)~\cite{huang2017arbitrary, deng2022stytr2, huang2023quantart} and FID metric~\cite{heusel2017gans}, which measure the perceptual similarity between the content and generated images from the aspects of image content, style and visual fidelity, respectively.

The evaluation is conducted on a small subset with 21 paired content and reference images sampled from of the WikiArt dataset~\cite{wikiart}. For fairness, all images are rescaled into $512\times512$ resolution and the statistic results are listed in \ref{tab1}. As we can see, the results are consistent with the visual effects in Fig. \ref{fig8}. The AdaIN~\cite{huang2017arbitrary} method is efficient but has low performance of structural fidelity on both SSIM-edge and IIT indexes. The WCT~\cite{li2017universal} receives obvious improvements in structural similarity, but the perceptual similarity is decreased due to the over-smoothing local structures. The increased trend of structural similarity is also observed in both PhotoWCT~\cite{li2018closed} and WCT$^2$~\cite{yoo2019photorealistic} methods. Moreover, they have much better perception-based LPIPS, Gram loss, and FID metric, which can be also demonstrated from their photo-realistic transferred effects. The StyTr$^2$~\cite{deng2022stytr2} shows very similar effects as AdaIN method and the QuantArt~\cite{huang2023quantart} produce high-quality consistent structures, but has limited perceptual similarity in Gram loss and FID metric, which may be caused by the limited color transform in some cases. In contrast, the proposed method gives a fine balance between structural fidelity and perceptual similarity in content, style, and visual fidelity, which is observed in the visual effects in Fig. \ref{fig8}. The benefits mainly underpin the fact that sparse representation provides a simple but effective tool that is especially suitable for low-level or middle-level feature extraction, especially in the case of pursuing photo-realistic image transfer effects.

\section{Conclusions}

In this paper, we propose a novel optimal transport over sparse dictionaries to explore the two-fold benefits of sparse representation and optimal transport. We have illustrated that sparse representation provides an easy-grasped tool to encode abstract features such as color, textures, and optimal transport over a small size of learned dictionaries is also computationally efficient in practice. As a result, it helps to simplify the procedure of many image-to-image translation problems significantly. Experimental results show that the proposed model is empirically solvable on several image-to-image translation tasks with plausible transferred results. It is worth noting that the proposed method can be further extended from different aspects, for example, using shared dictionaries, extending to multi-scale cases  and learning more confident individual styles with regularization techniques, which are leaving for further work.

\ifCLASSOPTIONcompsoc
  \section*{Acknowledgments}
   This work was in part supported by the National Science and Technology Major Project, China (Nos. 2019-I-0001-0001 and 2019-I-0019-0018) and Shandong MSTI Project, China (No. 2019JZZY010122), and Foundation for Innovative Young Talents in Higher Education of Guangdong, China (No. 2021KQNCX213). This work was also partially supported by the FWO Odysseus Project, Leverhulme Grant RPG-2017-151 and EPSRC grant EP/R003025/1. We thank Prof. Andreas Wiersmann to offer his paintings for our research. We appreciate the anonymous reviewers for their careful reading of our manuscript and their insightful comments and suggestions, and also the related online resources, including images, codes, software, and so on. 
\else
  \section*{Acknowledgment}
\fi


\end{document}